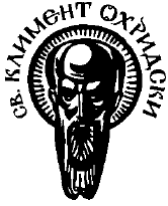

Софийски Университет "Св. Климент Охридски"

Факултет по Математика и Информатика

Катедра "Компютърна информатика"

Специализация "Изкуствен интелект"

# ДИПЛОМНА РАБОТА

Тема:

Проектиране, разработка и имплементация на средство за създаване на декларативни функционални описания на семантични мрежови услуги, базирани върху WSMO методология.

| | |
|---|---|
| Дипломант: | Петър Стефанов Кормушев, Ф№ M-21308 |
| Научен ръководител: | ст.н.с. д-р Геннадий Агре<br>Институт по информационни технологии – БАН |

юни 2005 г.
гр. София

„The Semantic Web is not a separate Web but an extension of the current one, in which information is given well-defined meaning, better enabling computers and people to work in cooperation."

<div align="right">Tim Berners-Lee</div>

———————————————

„Семантичната глобална мрежа не е самостоятелна мрежа, а по-скоро разширение на съществуващата, в което на информацията е даден добре дефиниран смисъл, позволяващ на хора и компютри да работя съвместно по-добре от преди."

<div align="right">Тим Бърнарс-Лии</div>

# **Резюме**

Мрежовите услуги (Web services) определят една нова парадигма в използването на глобалната мрежа (Web), при която една мрежа от компютърни програми става потребител на информация. За съжаление, съществуващите технологии за мрежови услуги описват единствено техния синтактичен аспект и по този начин се получава множество от статични услуги, които не могат да се адаптират към променящата се среда без човешко вмешателство.

Един начин за реализация на пълния потенциал на мрежовите услуги се търси чрез превръщането им в семантични мрежови услуги (Semantic Web Services) с прилагане на технологии от семантичната глобална мрежа (Semantic Web). Семантичните мрежови услуги са самостоятелни и семантично добре описани софтуерни ресурси, които могат да бъдат публикувани, намерени, композирани и изпълнени в глобалната мрежа по един сравнително автоматизиран начин.

Текущите усилия в тази насока са съсредоточени върху разработването на пълен набор от софтуерни средства за създаване и обслужване на семантични мрежови услуги. Важна част на една семантична мрежова услуга е декларативното онтологично описание на нейната функционалност. Формален синтаксис и семантика за това описание се описва на избрания език за моделиране на семантични услуги, който може да се базира на такива логически формализми като дескриптивна логика, логика от първи ред или фреймова логика. Функционалността на услугата се представя като определена съвкупност от логически изрази на този език.

Целта на настоящата дипломна работа е проектиране, разработка и имплементация на средство за създаване на такива семантични функционални описания на мрежови услуги. Ще наричаме това средство „Axiom Editor", като изборът на това име ще стане ясен в хода на изложението.

Изложението е разделено на четири големи глави. В първа глава се прави въведение в предметната област и се дефинира по-точно целта на дипломната работа. Във втора глава се прави анализ на съществуващите подходи и се проектира Axiom Editor. В трета глава се прави детайлна разработка на Axiom Editor. В четвърта глава се описва имплементацията на Axiom Editor и използваните технологии. Изложението завършва със заключение и насоки за бъдещо развитие.

# Благодарности

---

[1] ИИТ-БАН – Институт по информационни технологии на Българската Академия на Науките

# Съдържание

# 1. Въведение в предметната област

## *1.1. Семантични мрежови услуги*

Мрежовата услуга (web service) е самостоятелен, добре обособен и описан приложен модул, който предоставя някакъв вид бизнес функционалност на останалите приложения. Тя може да бъде имплементирана на който и да е език за програмиране и реализацията й не е видима за потребителя или приложението, което я извиква. Всяка друга програма, която има достъп до глобалната мрежа, може да отправя извиквания към мрежови услуги.

През последните години мрежовите услуги се превърнаха в преобладаващ метод за комуникация по Интернет на разпределени компютърни приложения. Мрежовите услуги значително подобряват взаимосвързаността на програмите, тъй като използват само стандартизирани технологии и формати като XML [XML, 2005], WSDL [Chinnici et al., 2003] и SOAP [SOAP, 2005]. Днес съществува огромно количество средства, които улесняват тяхното създаване, внедряване, използване и поддръжка.

Мрежовите услуги предлагат възможности за автоматизиране на определени процеси. Една от тях е възможността за автоматично извикване на мрежова услуга. Едно приложение може без намеса на човек да се свърже и извика отдалечена услуга, като в най-простия случай за това е достатъчен само адреса на мрежовата услуга. По дадения адрес, приложението може да получи документ, който описва мрежовата услуга, и от това описание програмата може автоматично да реализира извикване на мрежовата услуга.

Въпреки че този механизъм осигурява по-голяма автономност и взаимосвързаност на разпределените приложения, все пак има един голям проблем, който не може напълно да бъде преодолян от съществуващите технологии.

Проблемът, който възниква, е как да се определи коя мрежова услуга да се използва? Веднъж след като е определена мрежовата услуга, тя може да се извика по описания по-горе начин и да свърши работата, за която е предназначена. Но какво може да се направи, ако например потребителят често променя изискванията към мрежовата услуга? Или ако в динамична среда (като Интернет) доставчикът на услугата, която сме използвали по-рано, вече спре да я предоставя?

Във всеки от изброените случаи се налага ръчна намеса от страна на човек, за да се разреши съответният проблем. С текущите технологии за мрежови услуги потребителите ще трябва да намерят подходяща мрежова услуга ръчно, като сами установят дали дадена мрежова услуга обезпечава желаната функционалност или не. Това в повечето случаи се прави чрез ръчно преглеждане на каталози за мрежови услуги (като UDDI [UDDI, 2005]). За да се минимизира времето и работата, която трябва да се влага за поддържане на разпределени компютърни приложения, трябва да се създаде някакъв нов механизъм, който да позволи преодоляването на тези проблеми.

Именно тук се намесва семантичната глобална мрежа (Semantic Web). Това е инициатива на W3C консорциума и една от основните й цели е да улесни откриването (discovery) на мрежови ресурси [Semantic Web, 2005]. Семантичната глобална мрежа има за цел да направи възможно откриването на ресурси по смисловото им съдържание, а не просто чрез търсене по ключови думи, както се прави най-често в наши дни. Това се постига чрез добавяне на семантична информация към мрежовите ресурси. Тази информация обикновено се нарича семантична анотация (semantic markup). За да бъде разбираема както за хората, така и за машините, тази семантична информация трябва да бъде описана чрез споделени/общоприети/ концептуализации (shared conceptualizations). За тази цел се използват така наречените онтологии.

Терминът „онтология" е заимстван от философията, където с онтология се означава учението/науката за това, което съществува, т.е. за битието. Когато става дума за семантичната глобална мрежа, под онтология се разбира описание на понятията и релациите, които съществуват в дадена конкретна предметна област. Следователно, онтологията не е нищо повече от терминология за дадена област. Онтологиите позволяват на компютърните програми да интерпретират недвусмислено значението на мрежовите ресурси.

Един интересен факт за онтологиите е, че те могат да реферират други онтологии. Това в крайна сметка ще доведе до създаването на малко на брой независими от конкретна област (общи) терминологии и много на брой конкретни онтологии за конкретни области. Общите онтологии описват най-основни понятия и релации, докато конкретните онтологии се градят над тях и описват с повече подробности конкретни понятия и релации на дадена област.

През последните години бяха разработени множество програми за създаване на онтологии: Ontolingua, OntoSaurus, WebOnto, Protégé, OilEd, OntoEdit и др. [Gomez et al., 2002]. Ето защо въпросът за получаване на онтологии се счита за решен и настоящата дипломна работа цели създаване на средство, което ще използва готови онтологии, без да включва тяхното създаване. Тези готови онтологии ще се използват в процеса на създаване на семантична анотация на мрежовите услуги.

Съществуват две основни инициативи, целящи разработването на общ стандарт за семантична анотация на мрежовите услуги:

Първата е OWL-S – съвместна разработка на BBN Technologies, Carnegie Mellon University, Nokia, Stanford University, SRI International и Yale University [OWL-S, 2004]. OWL-S осигурява описване на три аспекта на една семантична мрежова услуга: какво изисква като входни данни, какво осигурява като резултат и как го прави.

Втората е Web Service Modeling Ontology (WSMO) – Европейска инициатива за създаване на онтология за описание на различни аспекти на семантичните мрежови услуги и за решаване на проблема с тяхната интеграция [WSMO, 2005]. WSMO консорциумът включва повече от 50 академични и индустриални партньори от Европа.

Настоящата дипломна работа е ориентирана изцяло към WSMO методологията, въпреки че основните принципи в нея биха могли да се използват и за произволна друга методология за семантично описание на мрежови услуги.

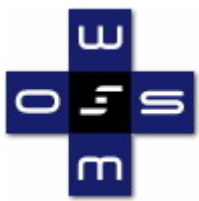

## *1.2. WSMO методологията*

WSMO (Web Service Modeling Ontology) или „Онтология за моделиране на мрежови услуги" е концептуален модел за описание на различни аспекти, свързани със семантичните мрежови услуги. Целта на WSMO и всички свързани с него инициативи е да се реши проблемът с интеграцията на мрежовите услуги чрез дефиниране на стандартизирана технология за семантични мрежови услуги.

WSMO използва за отправна точка платформата Web Service Modeling Framework (WSMF [Fensel & Bussler, 2002]), като я развива и разширява. WSMO продължава да се разработва и в момента от работната група „WSMO working group" (http://www.wsmo.org/). Ето защо някои части на методологията все още не са напълно дефинирани.

WSMF се състои от четири главни елемента (фигура 1.1):

- *Онтологии*: осигуряват формална семантика на информацията, използвана от всички останали елементи.
- *Цели*: задават целите, които един клиент поставя при използване на мрежова услуга.
- *Мрежови услуги*: представят функционалността и поведението по семантичен начин, за да може да се автоматизира използването им от компютърни програми.
- *Медиатори*: използват се като свързващи елементи, осигуряват връзки между останалите елементи и помагат за разрешаване на проблеми със съвместимостта. Тази част от спецификацията е в процес на активно разработване, така че все още няма завършена дефиниция.

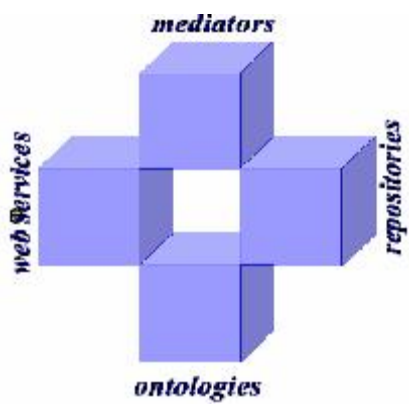

*Фигура 1.1. Главните елементи на WSMF*

WSMO наследява тези четири основни елемента от WSMF и допълнително ги подобрява и разширява. Следва по-детайлно описание на всеки от тези основни компонента на WSMO.

*Онтологиите* са ключов елемент в WSMO, защото осигуряват терминология за описание на останалите елементи в рамките на конкретна проблемна област. Те изпълняват двойна роля: от една страна, дефинират формална семантика за информацията, а от друга, свързват терминологиите на хората и машините. WSMO специфицира следните съставни части на онтологиите: *понятия (concepts)*, *релации (relations)*, *функции (functions)*, *аксиоми (axioms)*, *екземпляри (instances)*, *нефункционални свойства (non-functional properties)*, *импортирани онтологии (imported ontologies)* и *използвани медиатори (used mediators)*. Последните позволяват връзки между различни отнологии чрез използване на медиатори, които разрешават конфликти в терминологията.

*Мрежовите услуги* свързват компютри и устройства помежду им като използват стандартни Интернет-базирани протоколи за обмяна на информация. Възможността те да бъдат разпръснати в Интернет им дава предимството на платформената независимост. Всяка мрежова услуга представлява атомарно парче функционалност, която може да се използва многократно за изграждането на по-сложни услуги. За да се осигури тяхното намиране, изпълнение и композиране, мрежовите услуги се описват в WSMO от три различни гледни точки: не-функционални свойства, функционалност и поведение. Всяка една мрежова услуга може да се описа с множество интерфейси, но има точно едно описание на възможностите.

*Целите* описват желаните от потребителя функции, които се съпоставят с предлаганата от мрежовите услуги функционалност, описана в секцията „възможности".

*Медиаторите* описват елементи, които целят преодоляването на някакви структурни, семантични или концептуални различия, които може да се появят между отделни компоненти на едно WSMO опсиание. В момента спецификацията покрива четири вида медиатори: *OOMediators, GGMediators, WGMediators, WWMediators*.

Принципно е възможно създаването на един огромен, всеобхватен WSMO стандарт. Само че това не се прави, поради трудността от възприемане, която поражда. Вместо това се използва подход на изграждане на няколко слоя, като се започне с основния WSMO (наречен WSMO-Lite), продължи се с неговото разширяване (WSMO-Standard) и накрая се оформи пълна онтология (наречена WSMO-Full).

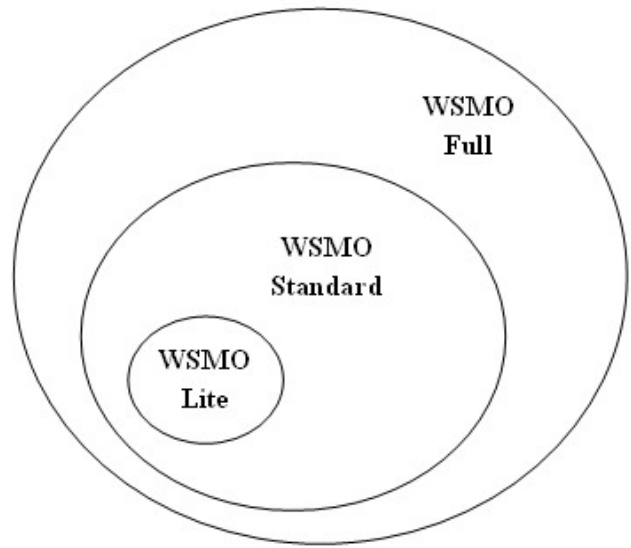

*Фигура 1.2. WSMO-Lite, WSMO-Standard и WSMO-Full.*

Както е показано на Фигура 1.2, различните варианти на WSMO се надграждат един друг, като всеки следващ наследява всичко от предходните.

За да се осъществи автоматизирано намиране, композиране и изпълнение на мрежови услуги само концептуален модел не е достатъчен. Трябва още и формален език, на който да се пишат семантичните описания на услугите според концептуалния модел. След това върху тези описания се прилагат стандартни механизми за логически извод.

За да се постигнат добри възможности за логически изводи на база на описанията на семантичните мрежови услуги, те трябва да са написани на език, който осигурява достатъчно добра изразителност и формализирана семантика. Фамилията от езици Web Services Modeling Language (WSML) служи точно за да формализира WSMO. Той се разработва от групата „WSML working group" (http://www.wsmo.org/wsml/), която е подгрупа на „WSMO working group".

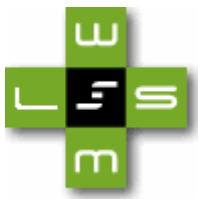

## *1.3. WSML – език за описание на семантични услуги*

Езикът WSML (Web Service Modeling Language) [Bruijn et al., 2005] осигурява формален синтаксис и семантика за WSMO. WSML се базира на различни логически формализми, а именно: дескриптивна логика, логика от първи ред, фреймова логика и логическо програмиране. Всички тези формализми допринасят за гъвкаво моделиране на семантични мрежови услуги.

WSML се състои от няколко разновидности, базирани на тези различни логически формализми, а именно: WSML-Core, WSML-DL, WSML-Flight, WSML-Rule и WSML-Full. Тези различни варианти на WSML отговарят на различно ниво на логическа изразителна сила и на използване на различни езикови парадигми. По точно, разглеждаме дескриптивната логика, логиката от първи ред и логическото програмиране като отправни точки за разработване на вариантите на езика WSML. Всички те са както синтактично, така и семантично наслоени (layered). Схематично това е илюстрирано на Фигура 1.3.

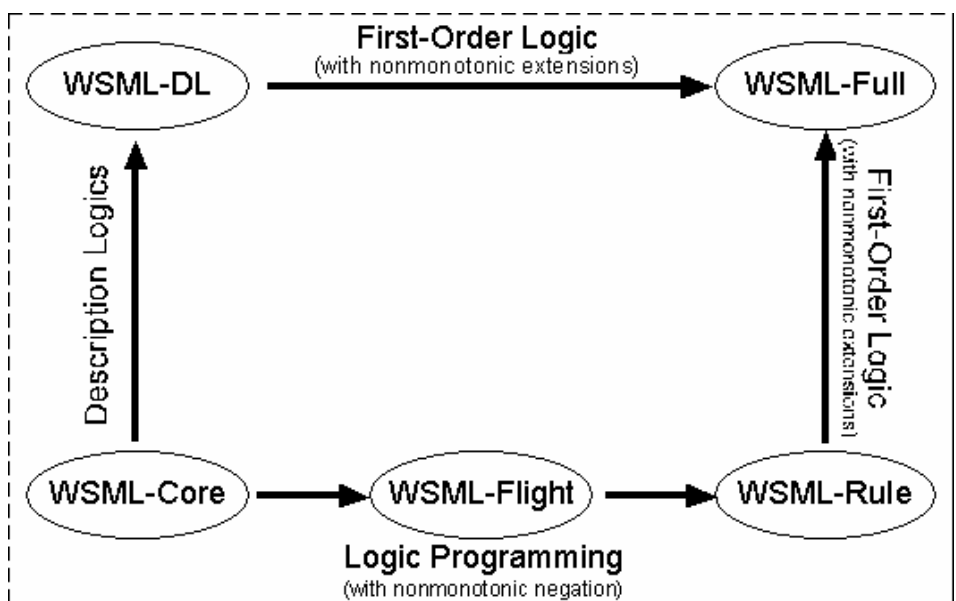

*Фигура 1.3. Връзките между разновидностите на WSML*

Всички разновидности на WSML са специфицирани чрез разбираем за човек синтаксис с ключови думи, подобни на елементите от концептуалния модел на WSMO. Тук няма да се спираме по-подробно на разновидностите на WSML. Детайлно описание на вариантите на WSML може да бъде намерено в [WSML, 2005]. WSML също е в процес на разработване. Дори в хода на разработването на настоящата дипломна работа синтаксисът на WSML претърпя значителни промени.

### 1.3.1. Елементи на WSML

Тук ще опишем само най-често използваните елементи на езика WSML, които са важни за правилно разбиране на изложението. Тези елементи се използват в описания на онтологии, цели, медиатори и описания на семантични мрежови услуги. За целите на настоящата разработка най-важно значение имат два от тези елементи: онтологии и описания на мрежови услуги.

**Онтологии в WSML**

Онтологии в WSML се описват с ключовата дума `ontology`, следвана от незадължителен адрес URI и дефиниции на елементите на онтологията.

Например:

```
ontology family
```

Основните елементи на онтологията, които WSML описва, са следните:
- Понятия
- Екземпляри
- Релации
- Аксиоми

**Понятия в WSML**

Дефиницията на понятие започва с ключовата дума *concept*, евентуално следвана от идентификатор на понятието. Следва незадължителна дефиниция за над-понятия (super-concepts) с ключовата дума *subConceptOf*, следвана от един или няколко идентификатори на понятия. Присъства още и незадължителен блок *nonFunctionalProperties*, след който следват нула или повече дефиниции на атрибути на понятието.

Важно е да се отбележи, че WSML разрешава наследяване на дефинициите на атрибутите от над-понятията. В случаите, когато две над-понятия имат дефиниция за един и същ атрибут, но с различни ограничения върху него, то наследеният атрибут има за ограничения конюнкцията от наследените ограничения от всички над-понятия.

Понятията се дефинират чрез йерархията на наследяването (дефинирано с над-понятия) и атрибутите, включително дефиниция на типовете им. Типовете на атрибутите могат да бъдат или от вградените типове, или пак понятия от онтологии. За указване, че даден атрибут може да има за стойност произволен обект, се използва като тип: *wsml#true*.

Пример за дефиниция на понятие в WSML:

```
concept Human subConceptOf {Primate, LegalAgent}
  nonFunctionalProperties
    dc#description hasValue "concept of a human being"
    dc#relation hasValue humanDefinition
  endNonFunctionalProperties
  hasName ofType foaf#name
  hasParent impliesType Human
  hasChild impliesType Human
  hasAncestor impliesType Human
  hasWeight ofType _float
  hasWeightInKG ofType _float
  hasBirthdate ofType _date
  hasObit ofType _date
  hasBirthplace ofType loc#location
  isMarriedTo impliesType Human
  hasCitizenship ofType oo#country
```

**Екземпляри в WSML**

Дефиницията на екземпляр започва с ключовата дума *instance*, евентуално следвана от идентификатор на екземпляра, ключовата дума *memberOf* и името на понятието, към което принадлежи този екземпляр.

Тази дефиниция е следвана от описание на стойностите, които са свързани с атрибутите на екземпляра. Всяка стойност се задава с ключовата дума *hasValue* и свързва име на атрибут с неговата стойност.

Пример за дефиниция на екземпляр в WSML:

```
instance Mary memberOf {Parent, Woman}
  nfp
    dc#description hasValue "Mary is parent of the twins Paul and Susan"
  endnfp
  hasName hasValue "Maria Smith"
  hasBirthdate hasValue _date(1949,9,12)
  hasChild hasValue {Paul, Susan}
```

**Релации в WSML**

Релацията описва взаимовръзка между множеството си от параметри. Дефиницията на всяка релация започва с ключовата дума *relation* евентуално следвана от идентификатор на релацията. WSML позволява дефинирането на релация с произволна арност (брой параметри). За всеки параметър допълнително се указва тип по същия начин както и за атрибутите на понятията. За релацията допълнително може да бъде зададена друга релация, която тя наследява, с ключовата дума *subRelationOf*.

Параметрите на релациите могат да имат имена, но това не е задължително. Ето защо е важно да се отбележи, че параметрите на релацията са строго подредени, за да е ясно тяхното използване.

Пример за дефиниция на релация в WSML:

```
    relation distance (ofType City, ofType City, impliesType _decimal)
subRelationOf measurement
```

**Аксиоми в WSML**

Дефиницията на аксиома започва с ключовата дума *axiom*, последвана от идентификатор на аксиомата и незадължителен блок *nonFunctionalProperties*. Логическият израз на аксиомата започва след ключовата дума *definedBy*.

Пример за дефиниция на аксиома в WSML:

```
  axiom humanBMIConstraint
      definedBy
        !- naf bodyMassIndex(bmi hasValue ?b, length hasValue ?l, weight
hasValue ?w)
           and ?x memberOf Human and
           ?x[length hasValue ?l,
             weight hasValue ?w,
             bmi hasValue ?b].
```

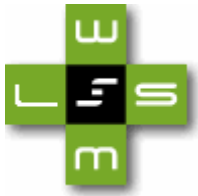

## 1.3.2. Синтаксис на WSML

Синтаксисът на WSML използва стил, наподобяващ фреймове. Информацията за едно понятие и неговите атрибути, за една релация и нейните параметри и за един екземпляр и неговите стойности на атрибути се задава в една голяма синтактична конструкция, вместо да се разделя на голям брой отделни парчета. В някои случаи това е възможно, но се практикува рядко.

Логически изрази се срещат в аксиомите, които се намират в описанията на целите (goals) и на възможностите (service capabilities) на семантичните мрежови услуги. Обичайният синтаксис за логически изрази включва всички разновидности на WSML и е еквивалентен на синтаксиса на WSML-Full. Той е вдъхновен от логиката от първи ред [Enderton, 2002] и т.нар. фреймова логика F-Logic [Kifer et al., 1995].

WSML разрешава използването на променливи на всички места, на които биха могли да се срещат понятия, атрибути, екземпляри, параметри на релации или стойности на атрибути. Една променлива обаче не може да стои на мястото на WSML ключова дума. Също така, променливи могат да се използват само във вътре в логически изрази.

WSML позволява използването на следните логически съединения: *and, or, implies, impliedBy, equivalent, neg, naf, forAll* и *exists* и на следните спомагателни символи: '(', ')', '[', ']', ',', '=', '!=', ':=:', *memberOf, hasValue, subConceptOf, ofType* и *impliesType*. Ето няколко примера за правилно построени аксиоми на синтаксиса на WSML [Bruijn et al., 2005]:

```
?x[gender hasValue {?y, ?z}] memberOf Human and ?y = Male and ?z = Female.
      (Човек не може да бъде едновременно мъж и жена)

?x[gender hasValue Woman] impliedBy neg ?x[gender hasValue Man].
      (Човек, който не е мъж, е жена)

?x[uncle hasValue ?z] impliedBy ?x[parent hasValue ?y] and
?y[brother hasValue ?z].
      (Брат на родител се пада чичо)

?x[distrust hasValue ?y] :- naf ?x[knows hasValue ?y].
      (Не вярвай на непознати)
```

Дори само тези примерчета показват, че синтаксисът на WSML е разбираем само за хора с добри познания в логика от първи ред. За неспециалист този синтаксис е почти неразбираем.

### 1.3.3. WSML описание на WSMO възможности на услуги

Описанието на възможности на услуги (service capabilities) започва с ключовата дума *capability*, която (незадължително) е следвана от име (идентификатор) на съответната възможност. Следва блок *nonFunctionalProperties* с нефункционални свойства, блок *importsOntology*, указващ използваните външни отологии и блок *usesMediator*, указващ необходимите медиатори. Тези блокове не са задължителни.

Блокът *sharedVariables* се използва за указване на променливите, които са споделени между секциите за пред-условия, пост-условия, предположения и ефекти на възможностите на услугата, които са дефинирани в секциите с ключови думи *precondition*, *postcondition*, *assumption,* и *effect* съответно. Броят на тези дефиниции не е ограничен, т.е. те може да се срещат по няколко пъти дори в рамките на едно и също описание на възможностите на мрежовата услуга.

Всяка от дефинициите се състои от ключовата дума, последвана от незадължителен идентификатор, незадължителен блок *nonFunctionalProperties* и задължителен логически израз, започващ с ключовата дума *definedBy* и имащ същото съдържание като на една аксиома.

Ето пример за описание на възможности на семантична мрежова услуга на езика WSML [Bruijn et al., 2005]. Аксиомите в него са логически изрази, започващи с ключовата дума *definedBy* и също написани на WSML:

```
capability
  sharedVariables ?child
  precondition
    nonFunctionalProperties
        dc#description hasValue "The input has to be boy or a girl
          with birthdate in the past and be born in Germany or has a parent,
          who is aGerman citizen"
    endNonFunctionalProperties
  definedBy
```

```
            ?child memberOf Child
                and ?child[hasBirthdate hasValue ?brithdate]
                and wsml#dateLessThan(?birthdate,wsml#currentDate())
                and ?child[hasBirthplace hasValue ?location]
                and ?location[locatedIn hasValue oo#de]
                or (?child[hasParent hasValue ?parent] and
                    ?parent[hasCitizenship hasValue oo#de] ) .
    assumption
        nonFunctionalProperties
            dc#description hasValue "The child is not dead"
        endNonFunctionalProperties
        definedBy
            ?child memberOf Child
                and naf ?child[hasObit hasValue ?x].
    effect
        nonFunctionalProperties
            dc#description hasValue "After the registration the child
                is a German citizen"
        endNonFunctionalProperties
        definedBy
            ?child memberOf Child
                and ?child[hasCitizenship hasValue oo#de].
```

Това формално описание на възможностите всъщност указва, че семантичната мрежова услуга предлага услуга за регистриране на дете като германски гражданин. В нея са описани необходимите пред-условия (детето да е родено в Германия), предположения (че детето не е мъртво) и ефекти (детето придобива германско гражданство.

## 1.4. Проектът INFRAWEBS

Следващата по-технологично ориентирана стъпка в процеса на развитие на семантичните мрежови услуги е предложена в текущия Европейски изследователски проект „INFRAWEBS" по 6-та рамкова програма, в който Институтът по информационни технологии на Българската Академия на Науките (ИИТ-БАН) е един от академичните партньори.

Основната цел на проекта е разработката на софтуерна среда, позволяваща доставчиците на мрежови услуги да генерират отворени и адаптиращи се платформи за използване на мрежови услуги. Една такава платформа се състои от взаимосвързани INFRAWEBS възли, всеки от които предоставя средства и адаптивни компоненти за анализи, създаване и поддръжка на семантични мрежови услуги. Една от по-конкретните цели на проекта се състои в разработването на пълен набор от софтуерни средства за създаване и обслужване на семантични мрежови услуги, базирани на WSMO методология.

Важна част от проекта INFRAWEBS е т.нар. Semantic Web Unit (SWU) – една обща платформа за семантично и базирано на онтологии създаване и поддържане на семантични мрежови услуги [Atanasova & Agre, 2004]. Основните модули в SWU са представени схематично на фигура 1.4.

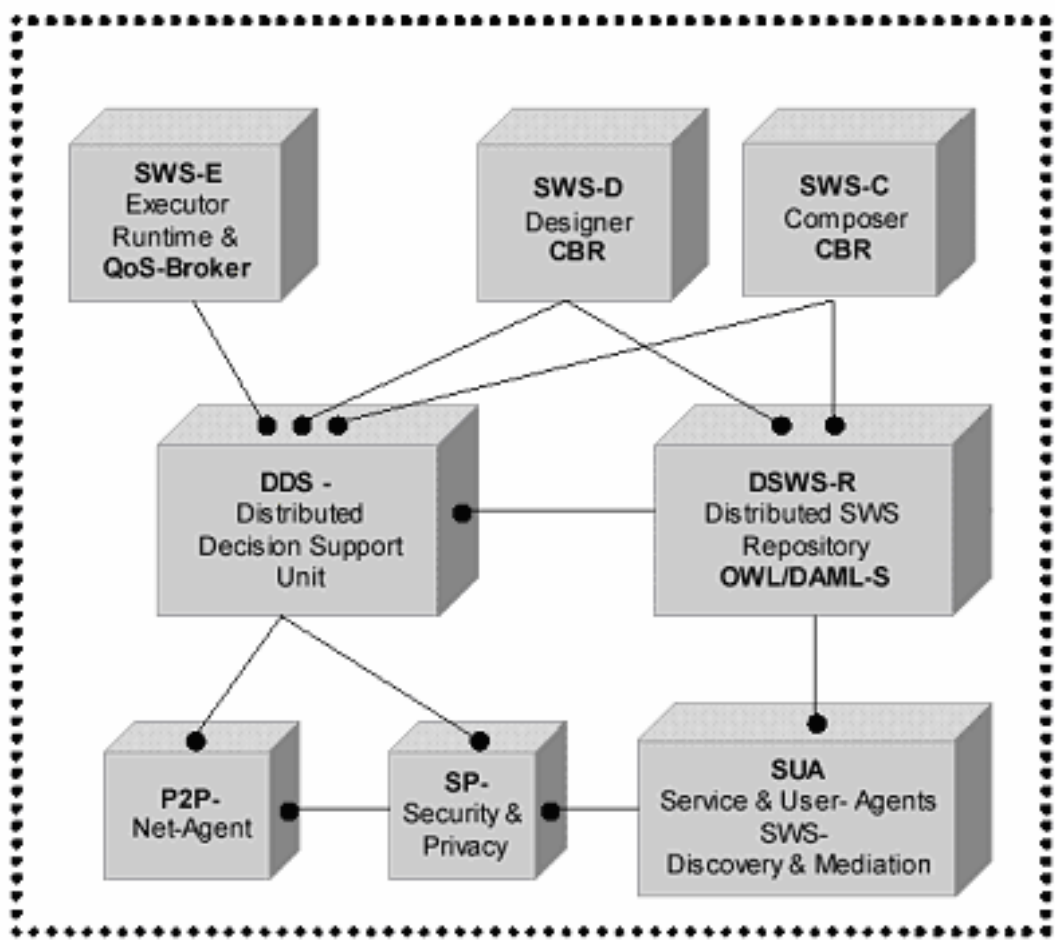

*Фигура 1.4. SWU - Semantic Web Unit*

Настоящата дипломна работа попада в тази част от SWU, наречена „Semantic Web Service Designer" (SWSD) или „Редактор за създаване на семантични мрежови услуги". SWSD цели *полу-автоматично конструиране на WSMO-базирани семантични мрежови услуги от налични WSDL описания на съществуващи мрежови услуги и знания за областта.*

Функционалната архитектура на SWSD се описва с множество от функции, необходими за създаване на семантична мрежова услуга, при използвана на избраната методология за описание на услугата WSMO.

Според WSMO [Roman et al., 2004], една семантична мрежова услуга се описва от:

- *Нефункционални свойства (Nonfunctional properties)*, които се състоят от главни свойства (по стандарта Dublin Core) и допълнителни свойства, свързани с аспекта качество на услугата (QoS) - бързодействие, надеждност, сигурност и т.н.

- *Импортирани онтологии (Imported ontologies),* описващи кои онтологии трябва да се импортират (или обработват). Такива онтологии съдържат понятия, релации и аксиоми, които са използвани в дефиницията на конкретната услуга.

- *Възможности (service capabilities)*, дефиниращи семантичната услуга откъм функционалност. На свой ред възможностите се описват чрез техните нефункционални свойства, пред-условия, пост-условия, предположения и последици (PPAE: pre-conditions, post-conditions, assumptions, effects). Добре обособеното описание на възможностите ги превръща в самостоятелно съществуващи обекти, които могат да се реферират или не от WSMO цели (goals). От своя страна, всяка част на PPAE е представена от своите нефункционални свойства и от *аксиоми,* които са логически изрази, написани на езика, дефиниран от WSMO. Към момента този език е WSML [WSML 2004].

- *Интерфейс,* описващ как може да бъде достъпена функционалността на услугата. До този момент все още WSMO консорциумът няма спецификация за описание на интерфейс на мрежова услуга.

Анализът на WSMO описанието на мрежови услуги показва, че създаването на такава услуга може да се реализира от следните главни функционални компонента [Atanasova & Agre, 2004]:

- *Structural Text Editor* – текстов редактор за създаване на дефиниции за нефункционалните свойства *(Nonfunctional properties)* и качество на услугата за описанието на семантичната мрежова услуга.

- *Axiom Editor* - редактор за създаване на PPAE частта от описанието на услугата, т.е. на възможностите ѝ *(Capabilities)*. Той трябва да бъде базиран на онтологии редактор на логически изрази, който създава правилно логическо описание на възможностите на услугата. Axiom Editor може да се разглежда като редактор на аксиоми на езика WSML, тъй като PPAE частта е съставена точно от такива логически изрази.

- *Control Flow Editor* – редактор за операционната семантика на услугата.
- *Data Flow Editor* – редактор за семантиката на данните. Този модул може да включва и валидираща част, която да извършва всички необходими проверки за съгласуваност.

Мястото на редактора Axiom Editor в общата архитектура на INFRAWEBS SWSD (Semantic Web Service Designer) е илюстрирано на фигура 1.5.

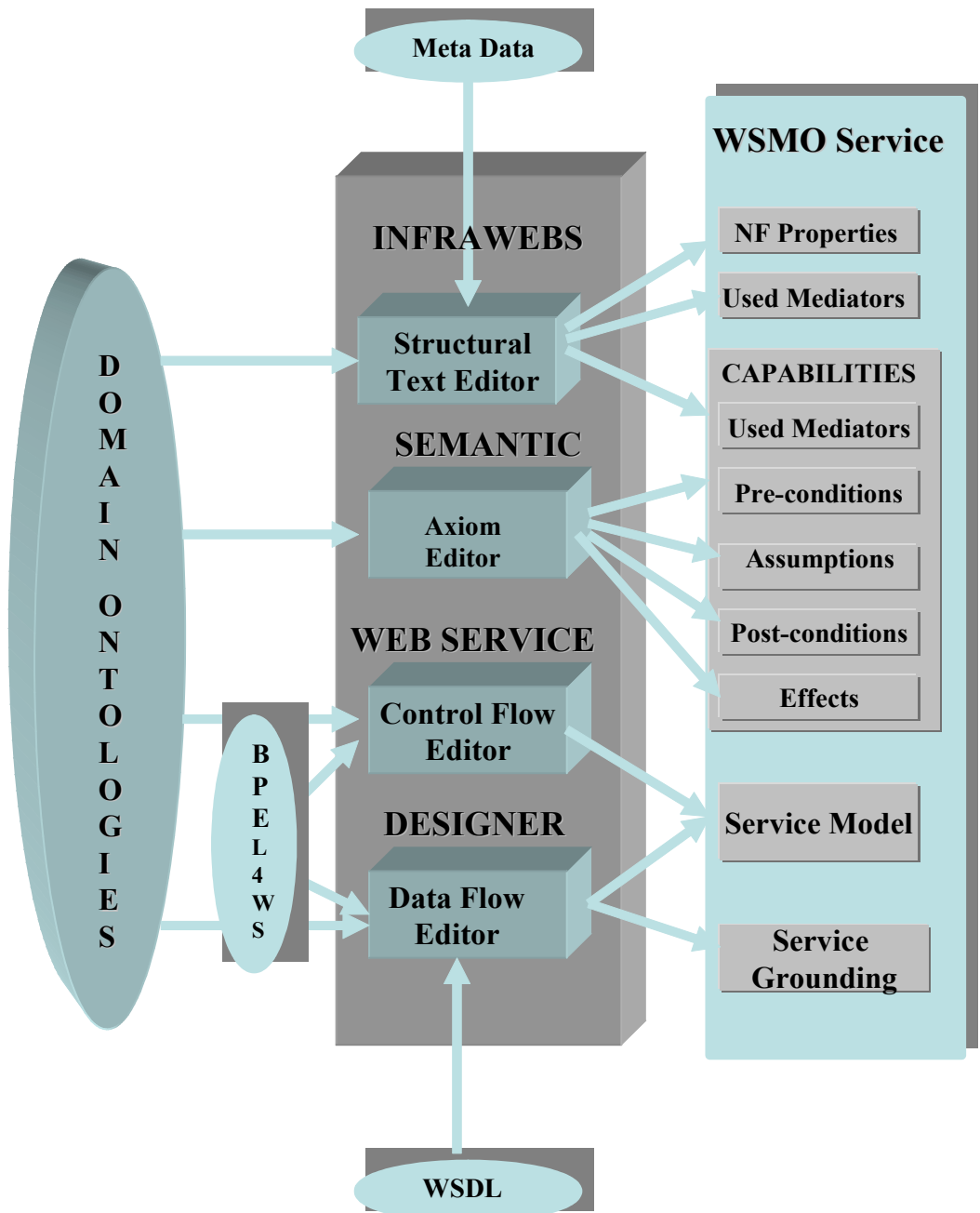

*Фигура 1.5. Функционална архитектура на INFRAWEBS SWS Designer*

## 1.5. Цел на дипломната работа

След като представихме семантичните мрежови услуги, методологията WSMO за тяхното описание и проекта INFRAWEBS, вече можем точно да дефинираме целта на настоящата дипломна работа:

*Проектиране, разработка и имплементация на редактора INFRAWEBS Axiom Editor, който е част от INFRAWEBS SWS Designer, служещ за създаване на логически изрази (аксиоми) за описание на възможностите на семантични мрежови услуги, конструирани съгласно WSMO методологията и написани на езика WSML.*

# 2. Проектиране на Axiom Editor

INFRAWEBS Axiom Editor е специализиран, удобен за използване инструмент за създаване и редактиране на сложни логически изрази, използващи множество от налични онтологии. В тази глава дискутираме изискванията към редактора INFRAWEBS Axiom Editor и проектираме процес на създаване на аксиоми и графичен интерфейс, който адекватно ги удовлетворява.

## *2.1. Анализ на изискванията към редактора*

В заданието за дипломната работа са дефинирани пет основни изисквания към INFRAWEBS Axiom Editor:

1. *Специализация:* инструментът е предназначен за конструиране на логически изрази, използвани в описание на възможностите на WSMO-базирани семантични мрежови услуги, а не за създаване на аксиоми в онтологии, които са много по-сложни. Анализът показва, че логическата структура на такива изрази е сравнително по-проста и в повечето случаи не изисква използването на такива сложни логически свързвания (оператори) като *Implies*, *ImpliedBy* или *Equivalent*.

2. *Леснота за използване:* потенциалният кръг от потребители на разработваното средство са доставчици и клиенти на мрежови услуги. И в двата случая потребителите не се очаква да бъдат специалисти в логиката от първи ред. По тази причина използването на какъвто и да е текстов редактор (дори и с много помощни функции) за конструиране на логическите изрази изглежда неприемливо като решение. Ето защо предлагаме да се разработи графичен начин за създаване и редактиране на сложните логически изрази, който позволява абстрахиране от конкретния синтаксис на WSML. В същото време е добре редакторът да позволява на по-напредналите потребители по-добър контрол над процеса на конструиране. Това би могло да се постигне по начин, подобен на използвания в HTML редакторите, в които има WYSIWYG интерфейс, като в същото време се позволява и директна модификация на изходния код.

3. *Интензивно използване на онтологии:* създаването на онтологии изисква тясно сътрудничество между високо квалифицирани логици и експерти в конкретни проблемни области. И двете категории потребители не принадлежат на множеството от потенциални потребители на предлаганото за разработка средство. За такива потребители има по-подходящи инструменти, като например редактори за онтологии от вида на Protégé2000 [Gennari et al., 2002] или Ontology Management Suit, който в момента се разработва в рамките на WSMO проекта. Заради тези съображения предполагаме, че нашите потребители ще бъдат главно *ползватели* на вече създадени онтологии, а не *създатели* на нови онтологии. Съответно, редакторът трябва да бъде по-скоро средство *за използване* на онтологии, отколко *средство за създаване* на такива.

4. *Семантична съгласуваност*: анализът показва, че основните трудности при създаване на сложни логически изрази са свързани не с изразяване на логиката, а с точното използване на имената на съществуващите в онтологии понятия, атрибути и релации, както и със съгласуване на техните типове. По тази причина процесът на създаване на логическите изрази в INFRAWEBS Axiom Editor трябва да бъде ръководен от онтологии, което означава, че на всяка фаза от този процес потребителят може да избира само тези елементи от съществуващи онтологии, които се съгласуват семантично с вече построената част от израза. С други думи, създаваната аксиома трябва винаги да бъде семантично съгласувана с онтологиите, използвани за нейното построяване.

5. *Разширяемост:* разработваното средство трябва да позволява лесно разширяване в посока потенциално неговото използване на създаване на описания на аксиоми в онтологии или други потребителски цели. Затова още по време на планирането се залага гъвкав процес, който по-късно лесно да бъде разширяван и допълван. От гледна точка на архитектурата, за постигане на разширяемост едно изцяло монолитно средство не е удачно, главно поради сложността на WSMO и заради различните изисквания на потенциалните потребители. По-скоро има нужда от компонентен модел, който да позволява разширения от трети страни под формата на приставки (plug-in).

## 2.2. Проектиране на процеса на създаване на аксиоми

В WSMO, възможностите на една услуга описват обобщените състояния на света, на информацията, която е необходима за правилното функциониране на услугата и на информацията, която се произвежда от услугата като резултат от дейността й. В по-сложните случаи възможностите на услугата могат да включват някакъв вид релация между двете състояния на света – преди и след изпълнението на услугата. Всяко обобщено състояние може да се разглежда като конюнкция и/или дизюнкция на твърдения за съществуване или отсъствие на някакви обекти в термините на понятия, екземпляри и релации от съществуващите онтологии.

По-точно, всеки обект от обобщеното описание на света е копие на някой сложен елемент от онтология, чиито свойства са частично спецификирани. Подобна спецификация може да се направи чрез явно свързване или приравняване (binding) на стойността на някое свойство с друг обект (или обекти), или чрез неявно включване в релация, в която обектът участва като параметър.

Аксиомите, използвани в описанията на възможности на услуги, може да се разглеждат като компактни записи на сложни логически програми в плосък (т.е. не е разбит на отделни клаузи) вид. Такова плоско представяне затруднява създаването и осмислянето на аксиомите. Ето защо считаме, че едно по-добре структурирано, йерархично представяне на аксиомите значително може да улесни боравенето с тях.
За тази цел създаването на аксиоми ще се моделира като процес, който се състои от комбиниране на три основни структурни фази – *дефиниция*, *уточняване* и *логическо развиване*.

Първата фаза „дефиниция" се използва за дефиниране на общите понятия, необходими за описване на смисъла на аксиомата.

Втората фаза „уточняване" (refinement) служи за по-точно специфициране на исканите свойства на тези понятия.

Третата фаза „логическо развиване" включва усложняване на логическата структура на аксиомата с използване на логическите оператори AND, OR и NOT.

### 2.2.1. Фаза „дефиниция"

По време на тази фаза се указва естеството на основните понятия, дефиниращи аксиомата. Тази фаза е еквивалентна на създаването на WSML израз от вида:

```
?Concept memberOf Concept
```

Този запис обозначава WSML променлива с име *?Concept*, която копира структурата на понятието *Concept* от някоя от наличните WSML онтологии. Например, да предположим, че в някоя онтология съществува понятието *reservationRequest*, описано в онтологията ето така:

```
concept reservationRequest
   nonFunctionalProperties
      dc#description hasValue "This concept represents a
      reservation request for some trip for a particular person"
   endNonFunctionalProperties
   reservationItem impliesType wsml#true
   reservationHolder impliesType prs#person
```

В резултат от дефинирането на променлива от тип – това понятие, ще се получи WSML логически израз от този вид:

```
definedBy
   ?reservationRequest memberOf tr#reservationRequest [
       reservationItem hasValue ?reservationItem
       reservationHolder hasValue ?reservationHolder
     ]
```

Тази операция може да се осъществява чрез избиране на подходящо понятие от онтология и създаването на обект в графичния редактор, който да представлява променлива от тип – това понятие. Едно подходящо графично представяне на обекта за тази цел е таблично изобразяване, показващо името на променливата, нейния тип и всички нейни свойства, взети от съответното понятие от онтологията. (виж фигура 2.1).

| Name: ?reservationRequest | Type: tr#reservationRequest |
|---|---|
| reservationItem | ?reservationItem |
| reservationHolder | ?reservationHolder |

*Фигура 2.1. Визуализация на променлива в графичния редактор*

## 2.2.2. Фаза „уточняване"

Тази фаза представлява рекурсивна процедура по уточняване на стойностите на някои от атрибутите на променливи, дефинирани на предната стъпка или стъпки. Подобно уточняване може да се направи чрез явно свързване на стойността на някой атрибут с друг обект (променлива, екземпляр или цял логически под-израз).

Основният проблем тук е как да се осигури семантична съгласуваност (consistence) на резултатния логически израз. За решаването му се използва основното на онтологии - свойството на релация "включване" (subsumtion relation): разрешава се свързване само на обекти със семантично съвместими типове. Проверката за съвместимост се прави чрез онтологиите.

Един атрибут може да бъде уточнен чрез свързване с един от следните обекти:

1. Понятие от онтология, чийто тип, указан с *ofType* или *impliesType* WSML израз, съответства на този на атрибута (по подразбиране);
2. Под-понятие на понятието, чийто тип, указан с *ofType* или *impliesType* WSML израз, е наследник на този на атрибута;
3. Екземпляр на съответното понятие, чийто тип, указан с *ofType* или *impliesType* WSML израз, съответства на този на атрибута;
4. Екземпляр на под-понятие, чийто тип, указан с *ofType* или *impliesType* WSML израз, е наследник на този на атрибута;
5. Релация, която притежава поне един параметър, който е съвместим по тип с типа на атрибута;
6. Споделена променлива (shared variable) с тип, който е съвместим с типа на атрибута;
7. Цял нов сложен логически израз, конструиран рекурсивно по същия начин, включително с използване на логическите оператори OR и NOT от фаза „логическо развиване".

**Уточняване чрез свързване с обект**

Уточняване на стойността на атрибут се прави чрез операция свързване ("Bind to…"). Необходимо е да се спомене, че типове на атрибути могат да бъдат както понятия от онтологии, така и вградени (built-in) типове данни (date, integer, string, и т.н.), а също и *wsml#true* (което означава „всички типове"). За улеснение, когато типът на атрибута е понятие, по подразбиране това понятие се използва за тип на обекта. Потребителят обаче може да избере или него, или някой под-тип на този тип от онтологиите.

Ако типът на атрибута отговаря на някой от вградените типове, тогава потребителят трябва сам да въведе подходяща стойност за него в правилния формат.

Ако пък типът е *wsml#true*, тогава на потребителя се разрешава да избере произволен елемент от наличните онтологии.

И в трите случая като резултат от уточняването в графичното представяне се появява нов обект, който автоматично се свързва чрез стрелка със стария обект.

Ето как ще изглежда това чисто схематично в графичното представяне:

```
?reservationRequest=tr#reservationRequest
reservationItem    = ?reservationItem      ------------>   ?reservationItem=tr#trip
reservationHolder  = ?reservationHolder                    origin       = ?origin
                                                           destination  = ?destination
                                                           departure    = ? departure
                                                           arrival      = ?arrival
```

Нека илюстрираме какво се случва с генерирания текст на логическия израз, ако свържем променливата *?reservationItem* към понятието *tr#trip*. Като резултат ще генерираме следния WSML израз:

```
definedBy
    ?reservationRequest memberOf tr#reservationRequest [
        reservationItem    hasValue ?reservationItem
        reservationHolder  hasValue ?reservationHolder
    ]
    and
     ?reservationItem memberOf tr#trip [
            origin       hasValue ?origin
            destination  hasValue ?destination
            departure    hasValue ?departure
            arrival      hasValue ?arrival
        ]
```

Вижда се, че съвсем естествено в израза се появява операторът AND, който явно никъде няма да присъства в графичното представяне на така построения логически израз. По този начин успяваме да скрием такива детайли от синтаксиса на WSML, които само биха затруднили и объркали потенциалните потребители на редактора. Разбира се, това не винаги е възможно или удачно.

**Уточняване чрез добавяне на алтернатива**

В някои случаи е необходимо да се укаже една или няколко алтернативи към съществуващите свързвания на една променлива. Това е възможно да се направи чрез операция „Добави алтернатива" върху вече съществуваща връзка, след което се избира елемент или тип от онтологиите, който да се използва за създаване на алтернативата. Като резултат от тази операция в графичния модел автоматично ще се появи оператор OR, в който ще се пренасочи старата връзка, а две нови връзки ще излизат от този оператор и ще го свързват със стария и новия обект съответно.

Нека илюстрираме какво ще генерираме за текст на логическия израз, ако добавим алтернатива към предишния пример за променливата за местоположение *?locatedIn*, която да бъде или Австрия (*loc#austria*) или Германия (*loc#germany*) – екземпляри на понятието *country*. Като резултат трябва да генерираме следния WSML израз:

```
definedBy
?reservationRequest[
        reservationItem  hasValue ?reservationItem
        reservationHolder hasValue ?reservationHolder
    ] memberOf tr#reservationRequest
    and
        ?reservationItem[
                origin       hasValue ?origin
                destination  hasValue ?destination
                departure    hasValue ?departure
                arrival      hasValue ?arrival
                ] memberOf tr#trip
    and
            ?origin[
            stationName hasValue ?stationName
            locatedIn   hasValue ?locatedIn
                ] memberOf loc#station
        and
            (?locatedIn hasValue loc#austria
           or
            ?locatedIn    hasValue loc#germany)
```

В този случай не е уместно да се крие операторът OR в графичния модел, защото се губи прегледност и точно затова той явно ще се появява като обект в диаграмата.

**Генериране на уникални имена за променливите**

Нека да предположим, че бихме искали да приложим същата процедура по уточняване към променливата *?destination*, която има същия тип като *?origin*. За да го направим, можем да използваме уточняване с понятието *loc#station*. Обаче, трябва да избягваме използването на имена на променливи, които вече съществуват в аксиомата (освен ако потребителят изрично не го иска). За да се предпазим от проблема с уникалността на имената, редакторът трябва сам автоматично да генерира имената по подходящ начин. За такъв начин е избрано генериране, при което за основа на името се използва името на типа, а при необходимост, за избягване на дублиране се добавя номер към името, ето така: *?duplicatename2*.

Да илюстрираме какво ще генерираме като текст на логическия израз при опит за дублиране на променливите *?stationName* и *?locatedIn*:

```
definedBy
?reservationRequest[
        reservationItem hasValue ?reservationItem
        reservationHolder hasValue ?reservationHolder
    ] memberOf tr#reservationRequest
    and
        ?reservationItem[
                origin       hasValue ?origin
                destination  hasValue ?destination
                departure    hasValue ?departure
                arrival      hasValue ?arrival
                ] memberOf tr#trip
    and
            ?origin[
            loc#stationName hasValue ?stationName
```

```
                    loc#locatedIn    hasValue ?locatedIn
                        ] memberOf loc#station
          and
             (?locatedIn hasValue loc#austria
            or
            ?locatedIn hasValue loc#germany)
           and
?destination[
            loc#stationName hasValue ?stationName1
            loc#locatedIn    hasValue ?locatedIn1
                ] memberOf loc#station
         and
            (?locatedIn1 hasValue loc#austria
           or
           ?locatedIn1 hasValue loc#germany)
```

Същата схема прилагаме за генериране на имена на променливи и в случаите, когато се прави друг вид уточняване и пак възникне опасност от погрешна двойна употреба на едно и също име. Един такъв случай е генериране на уникални имена за стойност на атрибутите. Ето примерен WSML израз за илюстрация:

```
(((?locatedIn[countryName hasValue ?countryName
          countryPopulation hasValue ?countryPopulation
          ….
          ] memberOf loc#country
     and
          ?countryName hasValue "Austria")
 or
(locatedIn[countryName hasValue ?countryName1
           countryPopulation hasValue ?countryPopulation1
           ….
           ] memberOf loc#country
          and
          ?countryName1 hasValue "Germany"))
```

Автоматично генерираните имена за променливи могат, разбира се, да се редактират по-късно от потребителя, стига това да не нарушава проверката за уникалност. Промяна в името на променливата следва да се отрази на всички срещания на тази променлива в логическия израз.

**Уточняване чрез отрицание**

Възможно е да се уточни стойността на една променлива чрез логическо отрицание NOT[2]. Например, можем да използваме NOT, за да укажем, че една услуга за пътувания (travel service), *не може да обработва* гари, които не са в Германия или Австрия. За да се постигне това, трябва да се вмъкне операторът NOT пред логическия израз, който свързва обектите Австрия и Германия с оператор OR. При това Axiom Editor трябва да генерира следния логически WSML expression:

```
definedBy
?reservationRequest[
       reservationItem    hasValue ?reservationItem
```

---
[2] Конкретната имплементация на NOT – чрез **neg** или **naf** – зависи от използвания WSML вариант.

```
            reservationHolder hasValue ?reservationHolder
        ] memberOf tr#reservationRequest
    and
        ?reservationItem[
                origin      hasValue ?origin
                destination hasValue ?destination
                departure   hasValue ?departure
                arrival     hasValue ?arrival
                ] memberOf tr#trip
    and
            (?origin[
            stationName hasValue ?stationName
            locatedIn   hasValue ?locatedIn
                ] memberOf loc#station
    and
            not
              ((?locatedIn hasValue loc#austria
               or
              ?locatedIn   hasValue loc#germany))
```

Ако обаче извършим същата операция върху връзката между оператора OR и възелът на екземпляра *loc#austria*, това ще доведе до забрана Австрия да бъде изходна точка за пътуването:

```
definedBy
    ….
    and
        ( not ?locatedIn hasValue loc#austria
         or
         ?locatedIn    hasValue loc#germany)
```

За да е ясна тази разлика и на потребителя на редактора, ще показваме оператора NOT. Той ще се появява като обект в графичния модел и мястото му ще е от ключово значение за смисъла на логическия израз.

**Уточняване чрез използване на релации**

Стойностите на една или няколко променливи могат да бъдат ограничени чрез релация, описана в използваните онтологии. В редактора предвиждаме да може да се избере релация и след това свободните нейни параметри да се свържат с променливи от подходящи типове, по подобен начин на вече описания за атрибутите.

Ако искаме да посочим, че трябва да съществува физическа връзка между началната и крайната точка за пътуването, това може да стане с помощта на релация. Ето как би изглеждал съответният WSML израз:

```
    tr#connectionExists(?origin, ?destination).
```

С това завършихме описанието на фаза „уточняване". Като резултат от нейното прилагане се получава графичен модел, който би могъл да изглежда така, както схематично е показано на фигура 2.2.

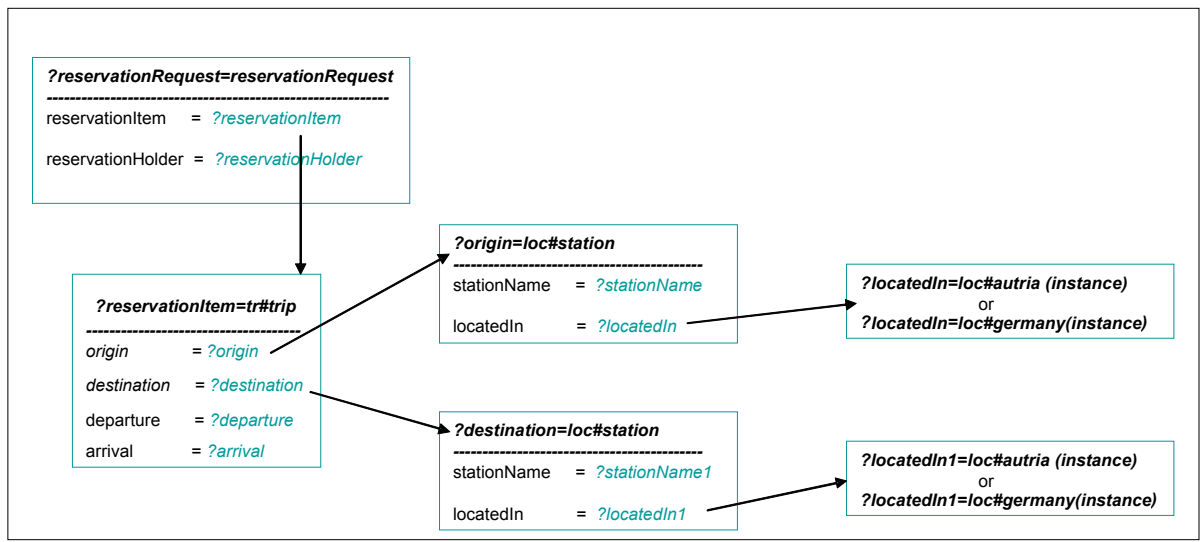

*Фигура 2.2. Схематичен изглед на графичен модел от фаза „уточняване"*

### 2.2.3. Фаза „логическо развиване"

Тази фаза всъщност включва операции по добавяне на логически оператори AND**,** OR и NOT към конструирания израз. Те може да се добавят, за да свържат две независими части от логическия израз, или пък може директно да се вмъкват във вече съществуващи изрази и да ги променят по този начин.

Директното вмъкване става задължително на мястото на съществуваща в модела връзка, която се „разцепва" на две и по средата й се вмъква новият логически оператор. Разбира се, тази операция се следи от проверките за семантична коректност и се разрешава вмъкването само на определени оператори на определени места. Проверката е контекстно-зависима, като се следи контекст далеч назад в израза, чак до корена на логическия израз.

Трябва изрично да се подчертае, че тази фаза се използва за логическото комбиниране на основните обекти, определящи аксиомата, дефинирани в първата фаза. С други думи, тези логически оператори не уточняват смисъла на някои от параметрите на вече дефинирани обекти, а усложняват аксиомата чрез указване на логическата връзка на иначе независими части от нея.

## 2.3. Проектиране на потребителския интерфейс

Редакторът за аксиоми е най-предизвикателната част от всички редактори в рамките на INFRAWEBS SWS Designer. По принцип, потребителският интерфейс трябва да скрива сложността на лежащия в основата синтаксис. В случая с аксиомите това е много трудно за постигане, защото целта е да се използва пълния потенциал на сложния логически формализъм за изразяване. А той трудно би могъл да се моделира с прости визуални контроли като падащи списъци или двумерни таблици.

Лесно се убеждаваме в това при разглеждане на съществуващите инструменти. Например, базираният на фреймове мета-модел на Protégé [Gennari et al., 2002] позволява дефиниране само на ограничен набор от логически твърдения (например ограничения на кардиналността са възможни, но квантифициране не е възможно).

Ясно е, че за нуждите на логическите изрази трябва различен подход, който да позволи по-пълното използване на езика. Освен това е добре да има възможност за работа с различно подмножество (т.е. изразителна сила) на езика, в зависимост от това колко е напреднал потребителя. Ще се опитаме да анализираме различни подходи за потребителски интерфейс, за да измислим най-удачно решение за редактора.

### 2.3.1. Подход с използване на усложнен текстов редактор

Един възможен подход за реализиране на редактор на логически изрази е да се започне с обикновен текстов редактор, към който да се добавят допълнителни функции, улесняващи писането на логическите изрази. Това са функции като синтактично оцветяване (syntax highlighting), автоматично дописване (auto-completion) и проверка на изразите за правилност. Такъв подход е използван в WSML text editor [Lausen et al., 2004], който е част от WSMO Editor (виж фигура 2.3)

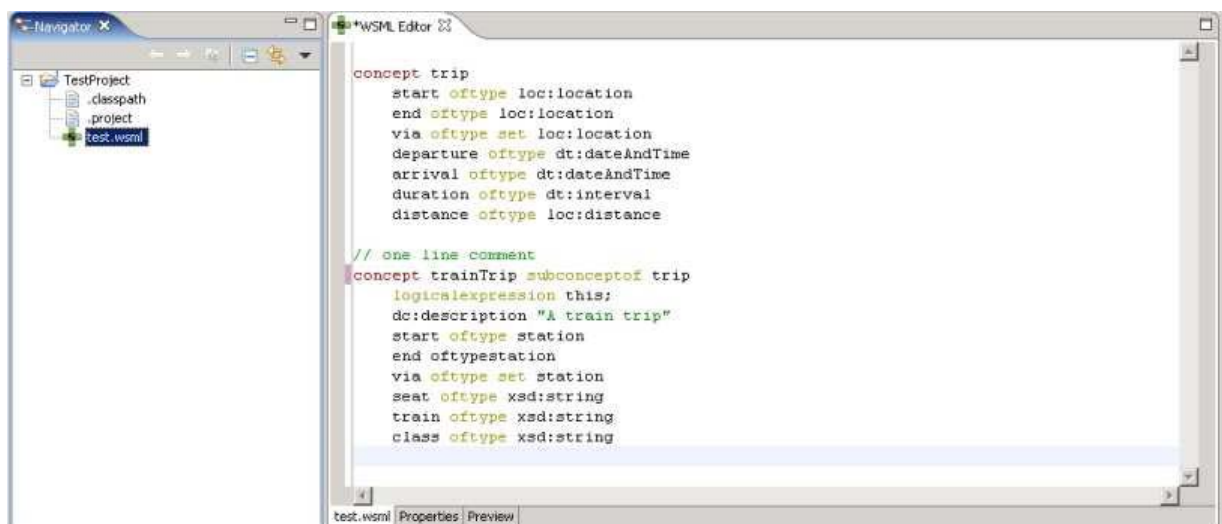

*Фигура 2.3. WSML text editor*

Очевидно, такъв подход може да се използва само за най-напредналите потребители, които знаят отлично синтаксиса на езика WSML, знаят „наизуст" имената на понятията, атрибутите им и се справят с директното записване на логическите изрази в текстов вид. Този подход обаче е напълно безполезен за по-неопитни потребители, каквито очакваме да бъдат потребителите на Axiom Editor.

## 2.3.2. Подход с използване на дървовидно представяне, подобно на редактор за онтологии

В областта на онтологиите вече съществуват множество инструменти, които значително улесняват работата по създаването и поддържането на онтологии. За съжаление, те не могат по никакъв начин да се използват наготово за създаване на логически изрази, въпреки отворената си архитектура. Това се дължи главно на потенциалната сложност на създаваните логически изрази и голямата изразителна сила на синтаксиса на използвания език (WSML).

В редакторите за онтологии обикновено визуализацията се състои в едно дървовидно представяне на йерархията на понятията и допълнителни странички с детайлна информация за избрания елемент.

Един чудесен редактор за онтологии е например Protégé2000 (фигура 2.4). Той има спечелена завидна популярност сред потребителите. Успехът му се гради на базата на:

1. отворен изходен код (open source code)
2. разширяема архитектура чрез приставки (plug-ins)
3. гъвкав мета-модел

Оценявайки важността на тези фактори, в настоящата дипломна работа те също са широко застъпени.

Поради наличието на добри инструменти за създаване на онтологии (напр. Protégé2000), счиме за уместно редакторът на аксиоми да бъде не универсално средство за създаване на онтологии, а средство за използване на такива. Така се предполага, че преди за да се създаде една аксиома с редактора, ще бъдат предварително осигурени необходимите онтологии, чиито елементи (понятия, атрибути, релации) да се използват в процеса на изграждане на логическите изрази на аксиомата.

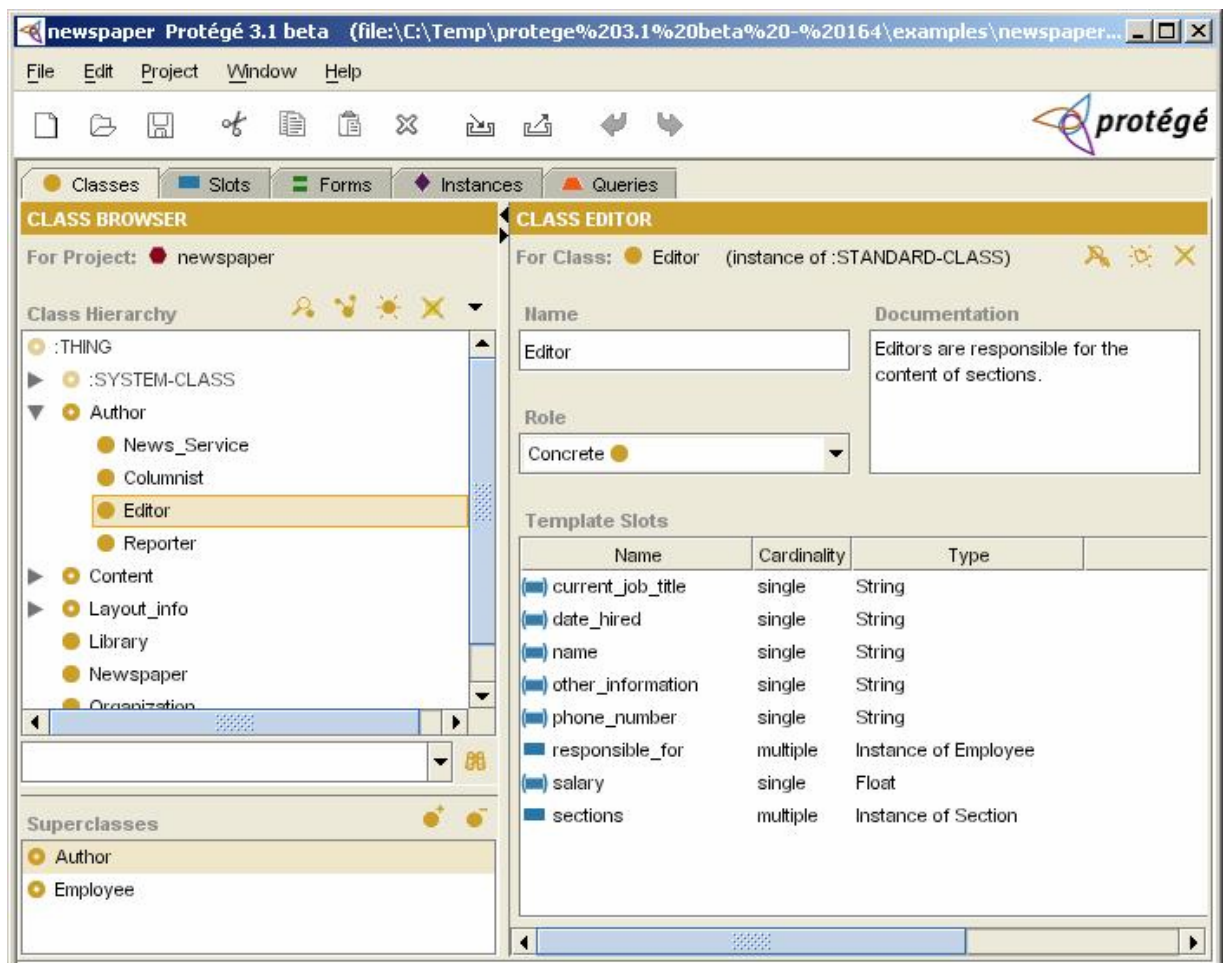

*Фигура 2.4. Визуализация на модел чрез дървовидно представяне в средата Protégé*

За съжаление, редактор за онтологии не би могъл да се използва за редактиране на логически изрази. Това е така, понеже липсата на достатъчно изразителна визуализация на сложната структура на логическите изрази ще затрудни използването му от по-широк кръг потребители, а това е едно от основните изискванията към настоящата разработка.

### 2.3.3. Подход с използване на вложени таблици, подобно на XML редактор

Друг възможен подход за реализиране на редактор на логически изрази е този, използван в популярните среди за създаване на XML документи. Това е визуализация чрез т.нар. вложени таблици. Като пример ще разгледаме средата Altova XML Spy 2005 [Altova, 2005], показана на фигура 2.5.

*Фигура 2.5. Altova XML Spy 2005*

Този подход притежава значителни предимства, като например:
- Естествено отразява сложната вложеност на логическите изрази във вложените таблици, като те могат да се "свиват", така че да се вижда във всеки един момент достатъчно компактен изглед на целия израз. Това води до по-лесно възприемане на сложни изрази.
- Не е нужно да се познава в детайли синтаксисът на езика
- Не е нужно да се знаят наизуст имената на атрибутите на понятията, тъй като те биха могли да се появяват автоматично в отделна вложена таблица.

За съжаление, този подход има и сериозни недостатъци, които пречат той да се използва за визуализация на логически изрази. Най-големият проблем е, че когато в един израз се цитира едно и също понятие на няколко места, при този подход то ще трябва да се появява на всички тези места. Това е така, защото практически зад вложените таблици стои една проста дървовидна структура, в която възлите и листата могат да са понятия и атрибути. Но в общия случай, дървовидната структура не може да спести това многократно дублиране на информацията, след като тя се цитира от много места в логическия израз.

## 2.3.4. Използване на граф за визуализация на логически изрази

И така, след анализиране на предимствата и недостатъците на няколко подхода, стигнахме до извода, че имаме нужда от по-голяма изразителна сила за компактно визуалното представяне на аксиомите. Такава по-голяма изразителна сила може да бъде осигурена от друга структура, а именно – граф.

Графът решава по естествен начин доста от споменатите в предишната секция проблеми. Например, многократно използваната информация в израза няма смисъл да се дублира, а само ще се слагат връзки (ребра на графа) към нея. Представянето на операторите и понятията може да бъде еднотипно, под формата на възли в графа, а връзките между отделните елементи да се представят чрез ребрата в графа.

Основното предимство на представянето с граф е, че за разлика от дървовидната структура тук може да има цикли, един обект може да се цитира (сочи) от няколко места, а визуално цялото представяне е много по-прегледно и лесно за възприемане от по-широк кръг потребители (фигура 2.6).

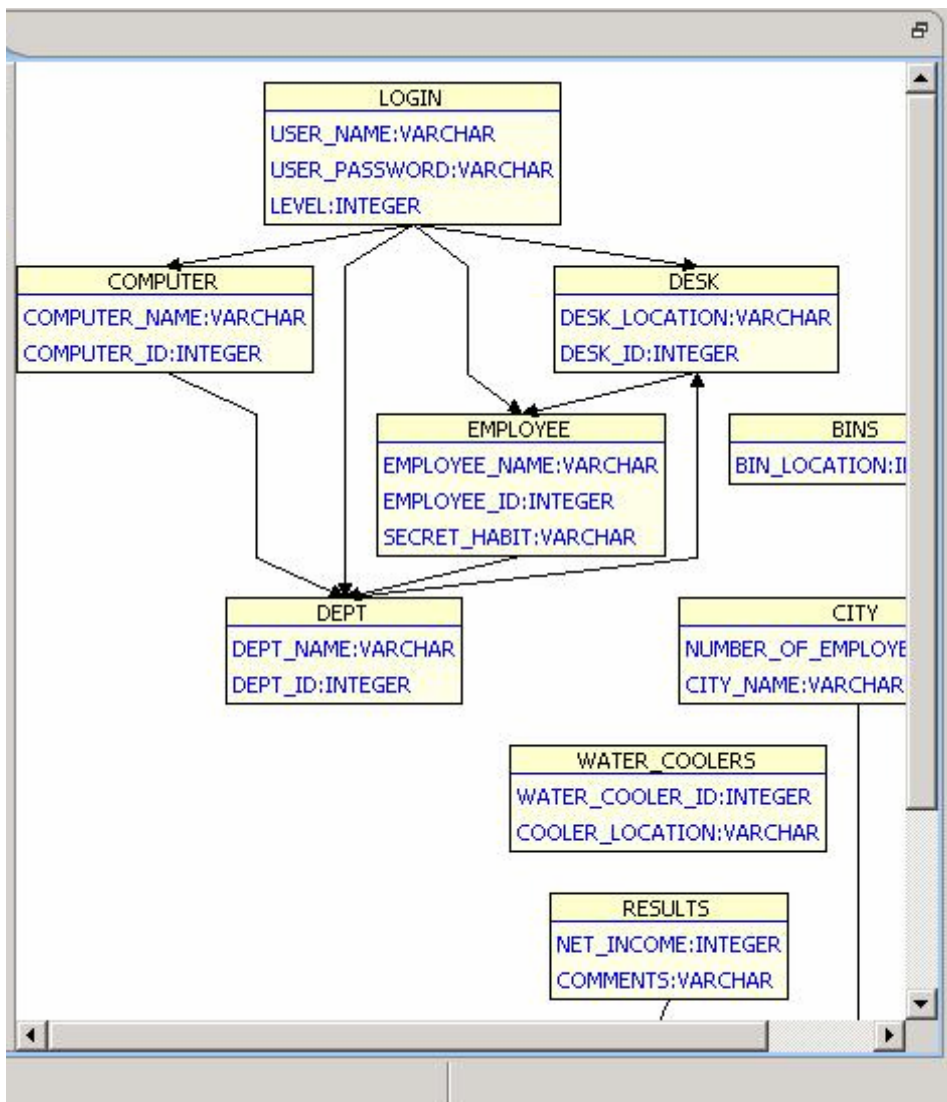

*Фигура 2.6. Примерна визуализация на граф*

# 3. Разработка на Axiom Editor

Дотук извършихме планиране от високо ниво на процеса на създаване на аксиоми и избрахме адекватен потребителски интерфейс, чрез който да се визуализира този процес. В тази глава ще разгледаме в много по-големи детайли процеса на конструиране на логическите изрази и начина на използване на онтологии в него.

## *3.1. Разработка на хранилище за онтологии*

„Хранилище за онтологии" представлява множество от онтологии, заредени в паметта, които предоставят семантичните елементи за конструирането на аксиоми. Тези елементи са: понятия и техните атрибути, екземпляри и техните стойности на атрибути, релации и техните параметри. Хранилището за онтологии е глобално за всички аксиоми, отворени в редактора.

Онтологиите считаме, че са дефинирани на езика WSML. Axiom Editor използва WSMO4J, за да чете файлове написани на този език и да зарежда дефинираните в тях онтологии в хранилището за онтологии.

### 3.1.1. Склад за онтологии

Тъй като хранилището за онтологии съществува само в паметта, необходимо е друго място, където онтологиите да се съхраняват за постоянно. За такова място Axiom Editor използва предварително зададен каталог (directory) от файловата система, който наричаме „склад за онтологии". В този каталог се съхраняват *.wsml файлове, които съдържат онтологии.

Всяка онтология има уникален идентификатор (на ресурс) (URI), който е записан в *.wsml файла, съдържащ онтологията. Когато трябва да бъде заредена онтология, на която знаем идентификатора, Axiom Editor претърсва склада за онтологии за този идентификатор и зарежда съответната онтология в хранилището за онтологии в паметта.

Всички *.wsml файлове, които трябва да бъдат достъпни за „механизма за зареждане при необходимост", трябва да се копират в склада за онтологии. За момента WSML документите, които се намират в Интернет, също трябва бъдат записани в локални файлове в този склад. Едва тогава ще могат да се зареждат от Axiom Editor, при поискване.

### 3.1.2. Зареждане на онтологиите „при необходимост"

Потребителят може да зареди онтология ръчно, като избере *.wsml файл от произволно място по файловата система. За да е възможно изграждането на една аксиома, трябва да бъде заредена поне една онтология.

Една аксиома може да цитира единствено елементи, които са напълно дефинирани в онтологиите от хранилището за онтологии. Например, ако дадена аксиома твърди нещо за хората по принцип, то цялата онтология, която дефинира понятието „Човек" трябва да бъде достъпна и заредена в хранилището за онтологии.

Axiom Editor предоставя механизъм за полу-автоматично зареждане на онтологиите, само при възникване на необходимост от тях.

### 3.1.3. Зареждане на импортирани онтологии

Онтологиите описват и взаимното наследяването на понятията. Понятията обикновено произлизат от (наследяват) едно или повече над-понятия. Над-понятията могат да бъдат дефинирани в други онтологии. Например, понятието "Личност", дефинирано в онтологията "Социология", може да произлиза от понятието „Човек", дефинирано в онтологията "Биология". В такъв случай казваме, че онтологията "Социология" декларира онтологията "Биология" като *импортирана онтология*.

За да може да се използва понятието "Личност", е необходимо само онтологията "Социология" да присъства в хранилището за онтологии. Тъй като на екрана се изписват и преките над-понятия на всяко дефинирано понятие, то понятието "Човек" също ще се вижда, но само като наименование – то няма да има структура. Потребителят няма право директно да създаде променлива от тип „Човек", тъй като той е от външна онтология. Първо трябва да се зареди онтологията „Биология".

Операцията по "зареждане на импортирана онтология" може да се прилага върху над-понятия, видими на екрана, които са дефинирани в импортираните онтологии. Декларацията на импортирана онтология съдържа нейния идентификатор. По него онтологията може да бъде намерена в склада за онтологии и да бъде заредена в хранилището за онтологии. Понятието, което е повод за зареждането, автоматично се маркира в новото дърво. Вече са достъпни атрибутите му и може да се създават променливи от този тип.

### 3.1.4. Допълване на списъка от атрибути

Понятията наследяват всички атрибути на своите над-понятия. Когато дадено над-понятие е дефинирано в импортирана онтология, която не е заредена в хранилището за онтологии, атрибутите на това над-понятия са неизвестни.

Ако "произлиза-от" е атрибут на понятието „Човек", а онтологията „Биология" все още не е заредена, тогава понятието „Личност" няма да има атрибут „произлиза-от", който да е видим и да може да се избира. (Виж примера в 3.2.2 Зареждане на импортирани онтологии)

Ако по-късно онтологията "Биология" бъде заредена (при необходимост), понятието „Човек" вече ще бъде дефинирано заедно с всичките си атрибути. Понятието "Личност" допълва списъка си с атрибути, като добавя "произлиза-от", както и всички останали атрибути, които са въведени от "Човек", но не са предефинирани от "Личност".

След зареждането на онтологията "Биология", вече е възможно уточняването на атрибута „произлиза-от" за всички променливи от тип „Личност" в модела на аксиомата.

### 3.1.5. Онтология от базови типове

Базовите типове са XML типове данни, като xsd:integer (целочислен), xsd:float (дробни числа) и xsd:string (текстов). Всички те са валидни WSML типове данни. Атрибутите и параметрите могат да бъдат от такъв тип. Понятията могат да наследяват тези типове. Изобщо, те са широко използвани, но липсва онтология, която да ги дефинира формално като понятия.

За удобство на потребителя Axiom Editor автоматично добавя в хранилището за онтологии своя собствена онтология от базови типове. Тя има идентификатор *http://www.w3.org/2001/XMLSchema#*, а съответния *.wsml файл се намира в склада за онтологии.

### 3.1.6. Кратки имена

Елементите на онтологиите се определят еднозначно с техния уникален идентификатор на ресурс (URI). URI-тата не се възприемат добре от крайния потребител, защото са прекалено дълги. Освен това, всички идентификатори на елементи от една и съща онтология имат общо начало, което е самият идентификатор на онтологията По тази причина, когато използваме ограничен набор от онтологии, по-удобно е да заместим дългите общи начала с кратки имена, които определят еднозначно онтологията в рамките на хранилището за онтологии. Този подход се използва, например, в езиците XML и WSML, където се декларират области от имена (namespaces) и се дават съкратени имена на дадени адреси (URLs).

Например, идентификаторът на онтология *http://www.wsmo.org/ontologies/dateTime* може да бъде заменен с краткото име "dt", а елементите от тази онтология могат да се цитират като "dt:елемент", вместо с пълния си идентификатор *http://www.wsmo.org/ontologies/dateTime#елемент*.

Обаче, едно и също локално име може да бъде използвано в различни онтологии и да означава различни елементи. По тази причина, Axiom Editor поражда уникални кратки имена за всяка онтология, която е заредена в хранилището за онтологии. Краткото име се базира на последната част от идентификатора на онтологията. Ако то вече е използвано, се добавят поредни номера в края, докато не се намери свободно име.

Краткото име за онтологията *http://www.wsmo.org/ontologies/dateTime* ще бъде "dateTime". Ако бъде заредена друга онтология с идентификатор *http://infrawebs.org/repository/dateTime*, то нейното кратко име ще бъде "dateTime1" и т.н.

### 3.1.7. Имената в модела на аксиомата

За да се различават еднаквите локални имена, използвани в различни онтологии, елементите на аксиомата, които цитират елементи от онтологиите изписват имената им във вида <кратко име>:<локално име>. Например, ако имаме две променливи, една от които от тип *http://www.sample.org/ontology/biology/Human*, а другата от тип *http://www.example.net/ontology/scifi/Human*, то понятията на тези променливи ще бъдат изписани съответно като 'biology:Human" и "scifi:Human". Това правило важи за имената на типовете на променливите, имената на типовете на екземплярите, имената на самите екземпляри, имената на релациите.

Кратките имена на идентификаторите на онтологии се използват при пораждането на WSML представянето на аксиомата. Това намалява количеството текст и прави експортираната аксиома по-четима за хората. Списъкът от кратки имена се експортира като списък от декларации на области от имена (namespaces).

### *3.2. Разработка на процеса на създаване на аксиоми*

Тъй като се предполага, че потребителите на този инструмент няма да са специалисти в областта на логиката от първи ред, основна цел на редактора е да гарантира семантичната коректност на конструираните логически изрази. За тази цел разработваме процес на конструиране на аксиоми, който е съобразен със семантиката. На всяка негова стъпка, потребителят може да извършва само такива действия, които са съгласувани с частта от аксиомата, конструирана до момента.

Предлагаме два режима на работа:

- Първият, *стандартен режим*, позволява само разширяване на съществуващата част от аксиомата, като новите елементи, за да са семантично съвместими, се избират от контекстни менюта. Този подход е „градивен" и е подходящ за начинаещи потребители.

- Вторият, *режим за напреднали*, позволява да се добавят самостоятелни елементи към работната площ, които по-късно могат да се комбинират по различни, семантично коректни начини. Това прави работата на по-опитните потребители по-ефективна.

По наши наблюдения, основните проблеми, които възникват по време на създаването на една аксиома, са свързани с използването на правилните имена за понятията, атрибутите, релациите и параметрите, както и с избирането на техните типове, а не толкова с изразяването на логическите връзки между тях.

Axiom Editor се справя с тези проблеми по два начина:

- Първо: потребителят е ограничен в избора си – той може да избира само елементи от онтологии и не може да пише произволни наименования на ръка.

- Второ: разрешено е добавянето само на семантично-коректни зависимости между елементите. Това означава, че на всяка стъпка от процеса, потребителят може да свързва само такива елементи от използваните онтологии, които са съвместими с изградената до момента част от аксиомата.

В резултат, създадените аксиоми са винаги семантично коректни спрямо онтологиите, които се използват, а процесът на конструиране на логическите изрази с този инструмент е изцяло *онтологичен (ontology-driven)*.

След внимателен анализ на всички възможни операции, които могат да се извършат с модела на аксиомата, избрахме подмножество, което съдържа само семантично коректни операции. Всяка такава операция се извършва върху някой вече съществуващ обект от модела на аксиомата.

В началото, всяка нова аксиома има един единствен елемент, който наричаме „Корен на аксиомата(Axiom root)". Той ще се отбелязва с надпис "Start" в графичния модел.

Процесът на изграждане започва с избирането на понятие от *хранилището за онтологии*. По избраното понятие се създава първата променлива в модела на аксиомата. Типът на променливата е именно понятието. Още с добавянето си, първата променлива се свързва с коренния елемент на аксиомата, "Start".

От този момент процесът на изграждане може да продължи с разширяване на аксиомата посредством семантично коректни операции. Тези операции се изпълняват върху наличните елементи в модела на аксиомата, а те са: *променливи, атрибути на променливите, екземпляри, връзки, оператори, релации* и *параметри на релации*. Освен върху тези седем елемента, операции могат да се изпълняват и върху целия модел на аксиомата.

### 3.2.1. Операции върху модела на аксиомата

Тези операции се изпълняват върху работната площ (полето, съдържащо графичния модел на аксиомата), а не върху конкретен елемент от модела на аксиомата.

    Създаване на променлива

Тази операция създава нова променлива в модела на аксиомата. Типът на променливата се избира от потребителя от хранилището за онтологии. Изборът не е ограничен – потребителят може да избере произволно понятие за тип на променливата.

Името на променливата се генерира автоматично от името на избраното понятие. Например, ако понятието е „Лице", то първоначалното име за променливата, което Axiom Editor ще предложи ще бъде "?Лице". Ако вече съществува променлива с това име, ще се добавят поредни номера към името по следния начин: "?Лице1", „?Лице2" и т.н. Това гарантира уникалността на променливите в рамките на една аксиома.

Ако това е първата променлива в модела, тя ще бъде автоматично свързана с коренния елемент на аксиомата, "Start". В противен случай, променливата просто ще се появи в работната площ без връзки към други елементи на аксиомата.

    Създаване на оператор

Тази операция добавя в работната площ нов оператор от указан тип. Типът на оператора се избира от меню – той може да бъде OR(дизюнкция), AND(конюнкция) или NOT(отрицание). Потребителят може лесно да промени типа с операцията върху оператори „Смяна на типа на оператор", но преди този оператор да бъде свързан с елементите от модела на аксиомата.

    Създаване на екземпляр

Тази операция добавя готов екземпляр към модела на аксиомата. Потребителят има възможност да избере екземпляр от хранилището за онтологии. Изборът не е ограничен – потребителят може да избере произволен екземпляр.

### Създаване на релация

Тази операция добавя релация към модела на аксиомата. Потребителят има възможност да избере релацията от хранилището за онтологии. Изборът не е ограничен – потребителят може да избере произволна релация.

### Създаване на връзка (режим за напреднали)

Тази операция добавя нова връзка в модела на аксиомата. Потребителят избира началния и крайния елемент на връзката. Изборът е ограничен само до семантично съвместими начален и краен елементи.

## 3.2.2. Операции върху променливи

Следните операции могат да се прилагат върху променливи. Всяка променлива има тип (който представлява понятие от хранилището за онтологии или е базов WSML тип данни), наименование (от вида: ?наименование) и списък с атрибути.

### Преименуване на променлива

Когато нови променливи се добавят в модела, Axiom Editor генерира автоматично имената им. След това, потребителят може да променя тези имена. Axiom Editor се грижи променливата да се промени името си от старото на новото на всички места в модела, където се използва тази променлива.

Обръщаме внимание, че автоматично генерираните от Axiom Editor имена на променливи са *уникални*. Потребителят може да променя имената, но без да нарушава уникалността им. Axiom Editor постоянно следи за запазването на уникалните имена, по време на всички операции.

### Копиране на променлива (добавяне на алтернатива)

Копирането на променлива води до създаването на неин дубликат в модела на аксиомата. Копието има също име и същия тип като оригинала. Axiom Editor автоматично свързва двете променливи с оператор OR. Целта на тази операция е да позволи бързо създаване на алтернативен набор от ограничения върху същата променлива в друг клон от логическия израз.

### Изтриване на променлива

Изтриването на променлива води до изтриването и на всички входящи и изходящи връзки от тази променлива към други елементи. По този начин се запазва съгласуваността на аксиомата.

### 3.2.3. Операции върху атрибут на променлива

Следните операции могат да се прилагат върху атрибутите на променливите в модела на аксиомата. Всеки атрибут има наименование (например "цвятНаОчите") и тип (наприимер „Цвят"), който е понятие от хранилището за онтологии.

#### Уточняване на атрибут с нова променлива от типа по подразбиране

Тази операция добавя нова променлива и поставя връзка от атрибута към променливата. Семантиката на връзката е, че стойността на атрибута се свързва с (приравнява на) новата променлива.

Типът на променливата се поставя автоматично, като се взима типа на атрибута. Това гарантира семантичната съвместимост на атрибута и променливата.

Името на променливата се генерира автоматично по името на атрибута. Например, ако атрибута е "цвятНаОчите", то за име на променливата първоначално Axiom Editor ще избере "?цвятНаОчите".Ако вече съществува променлива с такова име, ще се добавят поредни номера към името по следния начин: "?цвятНаОчите1", „?цвятНаОчите2" и т.н.

#### Уточняване на атрибут с променлива от друг тип

Тази операция също добавя нова променлива и поставя връзка от атрибута към променливата. Разликата от операцията с „типа по подразбиране" е, че тук потребителят има възможността да избере типа за новата променлива.

Появява се диалог, който съдържа подмножество на съдържанието на хранилището за онтологии. Изборът е ограничен само до тези понятия, които съвпадат или са под-понятия на типа на атрибута. Това гарантира семантичната съвместимост между типа на атрибута и избраното понятие.

Отново името на променливата се основава на името на атрибута, генерира се автоматично и се гарантира уникалността му.

#### Уточняване на атрибут с екземпляр от сложен тип

Тази операция поставя връзка между атрибут и нов елемент за екземпляр (на понятие). Потребителят има възможността да избере екземпляр от диалога с филтрирания списък с онтологии – за да се запази семантичната коректност на аксиомата, изборът е ограничен само до екземпляри на понятия, които са равни или са под-понятия на типа на атрибута.

Генерира се уникално име на променлива, което не се изобразява в екземпляра, а само до името на атрибута.

#### Уточняване на атрибут с екземпляр от базовия тип по подразбиране

Потребителят може да въведе конкретна стойност на атрибута. Това е разрешено само за атрибути, чийто тип е измежду базовите WSML типове данни. Такива са, например,

xsd:integer (цяло число), xsd:string (низ от букви/текст) и т.н. или техни подтипове, като dayOfMonth (число, което е ден от месеца).

Тази операция поставя връзка между атрибут и нов елемент за екземпляр. Потребителят въвежда стойност директно в екземпляра и тази стойност се асоциира с атрибута. Стойността може да бъде целочислена, текстова и т.н. в зависимост от типа на атрибута.

Типът на екземпляра автоматично се избира да съвпада с типа на атрибута. Това гарантира семантичната съвместимост между типа на атрибута и неговата редактируема стойност.

### Уточняване на атрибут с екземпляр от друг базов тип

Тази операция също добавя нов елемент за екземпляр, поставя връзка от атрибут към екземпляра и е разрешена само за атрибути, чийто тип е измежду базовите WSML типове данни и техните под-типове.

Разликата от операцията с "базовия тип по подразбиране" е, че потребителят има възможност да избере понятието, което ще е типа на новия екземпляр.

Изборът е ограничен само до онези *понятия* (не екземпляри), които съвпадат или са под-понятия на типа на атрибута. Това гарантира семантичната съвместимост между типа на атрибута и неговата редактируема стойност.

### Уточняване на атрибут с променлива от модела

Тази операция позволява на потребителя да свърже атрибут с някоя вече съществуваща променлива от модела на аксиомата. За да се запази семантичната коректност, изборът е ограничен само до променливи с тип, съвместим с типа на атрибута.

### Уточняване на атрибут с екземпляр от модела

Тази операция позволява на потребителя да свърже атрибут с някоя вече съществуващ екземпляр от модела на аксиомата. За да се запази семантичната коректност, изборът е ограничен само до екземпляри на понятия, съвместими с типа на атрибута.

## 3.2.4. Операции върху екземпляри от сложен тип

Следните операции могат да се прилагат върху екземпляри от модела на аксиомата. Всеки екземпляр е дефиниран в някоя онтология от хранилището за онтологии, има фиксирана стойност (например "Германия") и тип, който е понятие от хранилището за онтологии. За разлика от променливите, екземплярите нямат собствени имена. Вместо това, при обръщение към екземпляра, директно се използва стойността му.

### Изтриване на екземпляр

Изтриването на екземпляр води до изтриването и на всички входящи и изходящи връзки от него към други елементи. По този начин се запазва съгласуваността на аксиомата.

### Редактиране на екземпляри от базови WSML типове данни

Следните операции могат да се прилагат върху редактируеми екземпляри от модела, (които имат базов WSML тип данни или съответен подтип).

Стойността на тези екземпляри се въвежда директно от потребителя и може да се редактира. Например, текста "Петър Иванов" може да се въведе за стойност на екземпляр от тип xsd:string. По-късно тази стойност може да бъде променена на "Стоян Иванов".

Типът на екземплярите се изписва на работната площ и този тип определя какви стойности могат да бъдат въвеждани.

## 3.2.5. Операции върху връзки

Следните операции могат да се прилагат върху връзките. Всяка връзка има точно по един начален и един краен елемент.

### Добавяне на алтернатива

Тази операция вмъква оператор OR по средата на връзката. Това е позволено само за следните видове връзки:

- връзка от атрибут към променлива или екземпляр
- част от верига от връзки с начало – корена на аксиомата (Start)

### Добавяне на OR към връзка от атрибут към променлива или екземпляр

Тази операция добавя алтернативно уточнение на стойността на даден атрибут. Предназначението й е същото като на операцията, описана в "Операции върху атрибут на променлива" и трябва да има същите разновидности. Това може да се постигне с подобно функционално подменю (Уточняване на атрибут с. . .). След като се избере и изпълни съответната разновидност на операцията, към модела на аксиомата се добавя и оператора OR. Първият му операнд се свързва със старото уточнение на атрибута, а вторият – с новодобавеното уточнение, което може да бъде избрано от онтология, генерирано с тип по подразбиране или посочено измежду съществуващите елементи.

### Добавяне на OR към верига от връзки с начало – корена на аксиомата (Start), но не и с начало атрибут

Тази операция разделя една съществуваща връзка на две алтернативи. За втората алтернатива са възможни всички разновидности на избор – избор на променлива или екземпляр от онтология, генериране на променлива с тип по подразбиране, избиране на типа за екземпляр от някой базов тип, генериране на екземпляр от базов тип по подразбиране или избиране измежду съществуващите елементи.

### Добавяне на оператор AND

Тази операция вмъква конюнкция (оператор AND) по средата на връзка, която е част от верига връзки с начало – корена на аксиомата (Start), но не и с начало атрибут. За втория операнд на AND са възможни всички разновидности на избор, като тези при добавяне на OR в същия контекст.

### Вмъкване на оператор NOT

Тази операция добавя оператор NOT по средата на която и да е връзка.

### Промяна на началната точка (режим за напреднали)

Тази операция премества началната точка на връзката от един на друг елемент в модела на аксиомата. Крайният елемент на връзката не се променя. За нови начални елементи се допускат само елементи, съвместими с контекста на връзката, за да се запази семантичната коректност на аксиомата.

### Промяна на крайната точка (режим за напреднали)

Тази операция премества крайната точка на връзката от един на друг елемент в модела на аксиомата. Началният елемент на връзката не се променя.

За нови крайни елементи се допускат само елементи, съвместими с контекста на връзката, за да се запази семантичната коректност на аксиомата.

### Изтриване на връзка

Тази операция премахва само връзката между началния и крайния елемент. Елементите не се изтриват, защото потребителят може да поиска да ги свърже в друга конфигурация по-късно. За да се запази семантичната коректност на аксиомата, подобни несвързани елементи не се включват в генерирания WSML текст на аксиомата.

## 3.2.6. Операции върху оператори

Следните операции могат да се прилагат върху оператори. Поддържат се три типа оператори – OR (дизюнкция), AND (конюнкция), NOT (отрицание). Всеки оператор има точно една входяща връзка и няколко изходящи. Изходящите връзки завършват в операндите на оператора. Операторът NOT има точно един операнд. Операторите AND и OR имат поне два операнда.

Тъй като потребителят може да изтрива всяка връзка, операторите може да останат невалидни – да имат недостатъчен брой операнди или да нямат входяща връзка. Ако това се случи, операторът се изобразява с променен цвят в диаграмата и не се включва в WSML текста на аксиомата.

### Избор на типа оператор

Тази операция се използва, за да се промени типа на съществуващ оператор на OR, AND или NOT.

Това може да крие много рискове за семантичната коректност. Затова, операцията е разрешена само когато операторът няма нито входящи, нито изходящи връзки.

### Изтриване на оператор

Тази операция премахва оператора от модела на аксиомата. Всички входящи и изходящи връзки също се изтриват. Това може да доведе до появата на несвързани елементи в модела. За да се запази семантичната коректност на аксиомата, подобни елементи не се включват в генерирания WSML текст на аксиомата.

### Добавяне на операнд

Тази операция добавя изходяща връзка към оператор. Появява се същото подменю с операции за уточняване, което се използва при добавяне на алтернатива на връзка. Разликата е в това, че не се добавя нов оператор, а се използва съществуващия. Както обикновено, новият операнд може да бъде избран от онтология, да бъде генериран с тип по подразбиране или да бъде посочен измежду наличните елементи.

Възможните операции се определят от типа на веригата, в която входящата връзка участва. Тя може да започва от корена на аксиомата или от атрибут. Ако започва от атрибут, е разрешено само добавянето на операнд, който е съвместим по тип с типа на атрибута.

Забележка: Операторът NOT не може да има повече от един операнд.

# 4. Имплементация на Axiom Editor

Този раздел съдържа описание на реализацията на избраните в етапи проектиране и разработка решения за процеса на редактиране и визуализация на аксиоми.

Едни от най-тежките за удовлетворяване изисквания към настоящия редактор са свързани именно с интерфейса и това доколко той е интуитивен и лесен за използване от широк кръг потребители. Ето защо изборът на средства за имплементацията е толкова важен за постигане на желаните резултати.

## *4.1. Избор на средства за имплементацията*

За реализиране на предложената архитектура трябва да се избере първо език за програмиране, а след това и набор от програмни средства, които да се използват.

### Език за програмиране

Тъй като повечето съществуващи инструменти в тази сфера (напр. Средства за създаване на онтологии) са Java-базирани, затова Java е като че ли най-удачният език за програмиране, на който да се напише имплементацията. Редакторът работи в средата J2SDK 1.4.2 или по-нова.

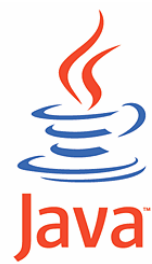

### Среда за програмиране

Платформата Eclipse [Eclipse, 2005] не е насочена към разработване на конкретен вид приложения, а е по-скоро универсална платформа за разработване на всякакви инструменти. Тя работи изцяло на принципа на отворения изходен код (open source).

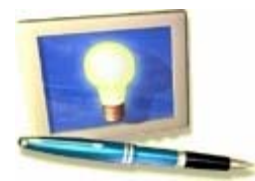

Eclipse поддържа разработването на приложения под формата на приставки (plug-in) и има библиотеки за изграждане на мощни редактори [Hudson, 2003]. Ето защо Eclipse е идеална платформа за имплементиране на разширяемата архитектура, която възприехме.

### Платформа за приложението

Платформата Rich Client Platform (RCP) на Eclipse позволява създаването на Java приложения, които могат успешно да се конкурират със собствените (native) приложения на която и да е платформа. RCP приложенията се базират на архитектурата с приставки (plug-in) на Eclipse. Целта е да се изпълнява самостоятелна програма с пълнофункционален интерфейс, без потребителят да знае за присъствието на Java или Eclipse.

## Достъп до онтологиите

Достъпът до WSMO-базирани хранилища с WSML онтологии се осъществява чрез WSMO4J [WSMO4J, 2005], което представлява WSMO API. От него се използва WSML Parser за зареждане на онтологиите от външни файлове.

## Платформа за визуализация

За визуализация се използва платформата GEF (Graphical Editing Framework) в комбинация с Draw2D за изрисуване на графичните фигури.

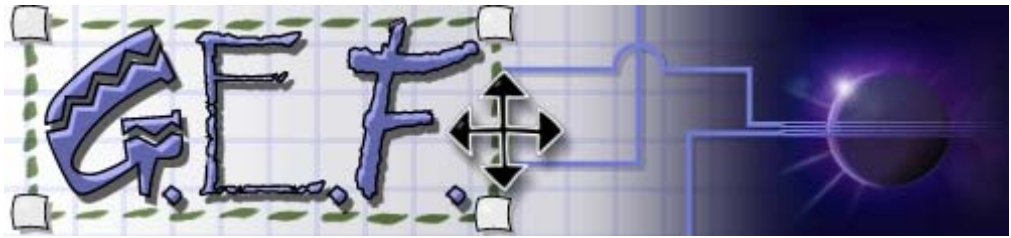

Graphical Editing Framework (GEF) представлява платформа за графично редактиране [Mehaut et al., 2004]. GEF предоставя мощни средства за създаване на визуални редактори за произволни модели. Ефективността на платформата се крепи на модулната й структура и ефективност, идваща от добре употребените шаблони за проектиране (design patterns) [Gamma et al., 1995].

За начинаещи ползватели, платформата GEF изглежда стряскаща, заради огромното количество термини и технологии, които трябва да се научат. Но веднъж научени и правилно използвани, те помагат да се разработи изключително сложен, гъвкав и в същото време лесен за поддържане софтуер [Moore et al., 2004].

За да не се товари излишно изложението, в *Приложение 2* се разглежда свързаната с GEF имплементация на графичния редактор Axiom Editor. В него са покрити голяма част от принципите на GEF, които са от голямо значение, за да бъде разбрана правилно имплементацията на настоящия редактор.

Дефинираните в Приложение 4 термини се използват наготово в тази глава, така че е важно те да бъдат ясни. Основното, на което трябва да се наблегне, е, че в GEF се използва изчистен шаблон Модел-Изглед-Контролер, който налага ред ограничения върху имплементацията на Axiom Editor.

## *4.2. Имплементация на потребителския интерфейс*

Външният вид на програмата се изгражда от няколко основни компонента. Работата с редактора става чрез взаимодействие с тези интерфейсни компоненти. Един типичен общ изглед на Axiom Editor е показан на фигура 4.1.

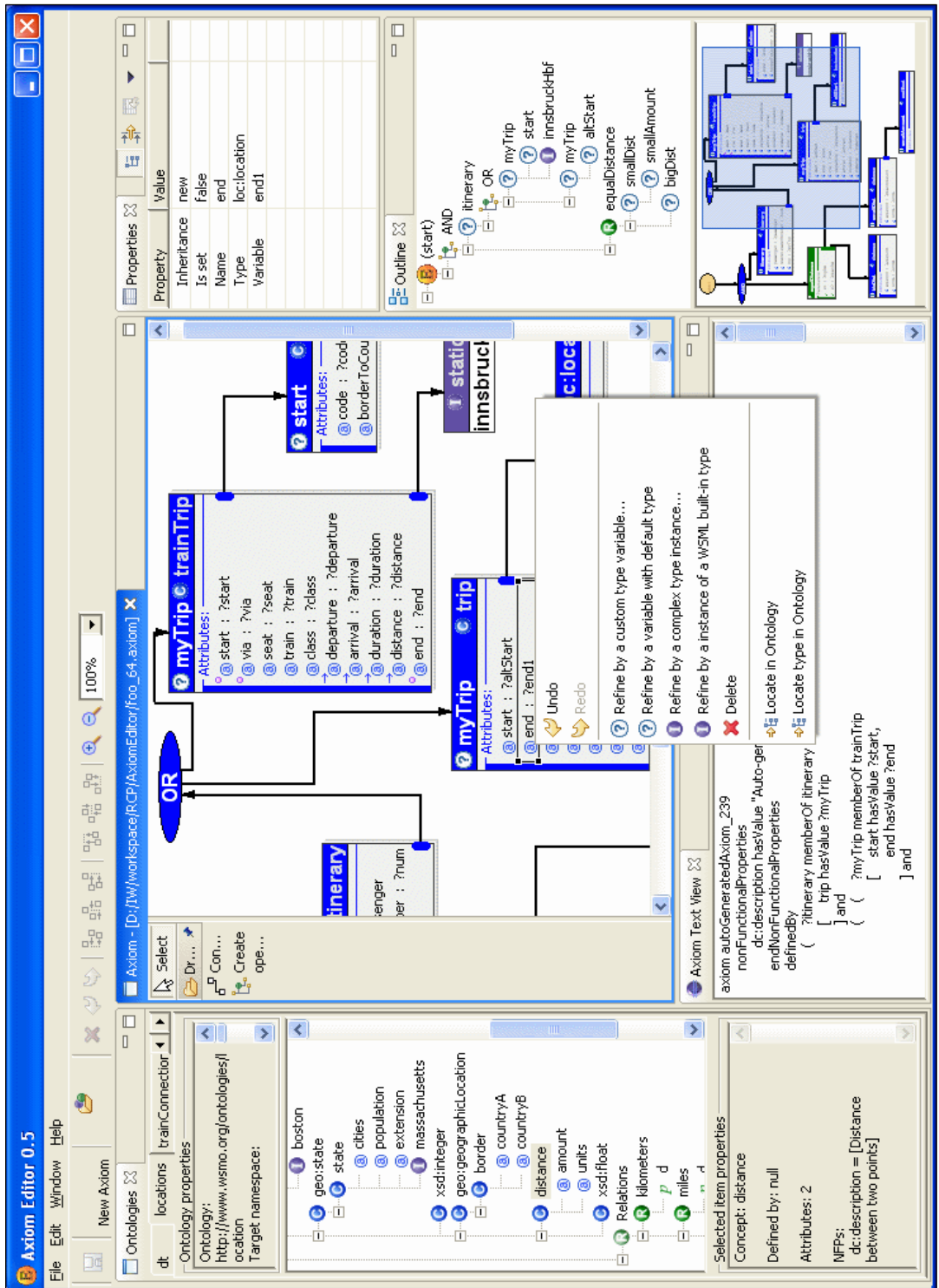

*Фигура 4.1. Общ изглед на Axiom Editor.*

## 4.2.1 Изглед на онтологиите (Ontology view)

Този изглед (прозорец) осигурява функционалността по зареждане и визуализация на онтологии. В него не може да се редактира, а само да се използват елементите за създаване на нови обекти в модела.

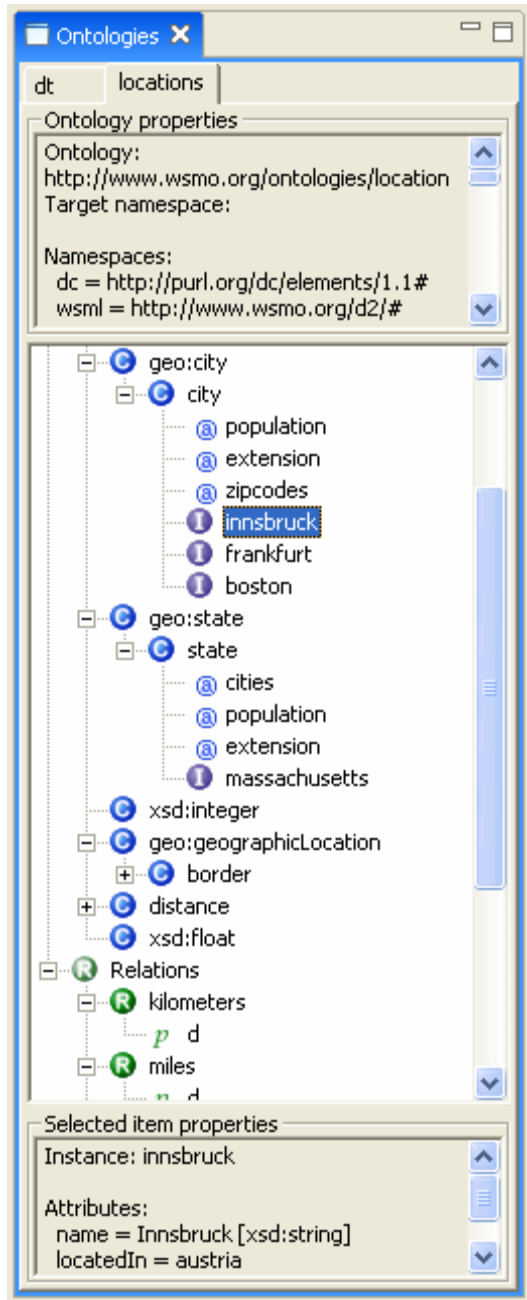

В изгледа на онтологиите се визуализират следните елементи: понятия (C), атрибути (@), екземпляри (I), релации (R) и параметри (p). Тези елементи се изобразяват в дървовидна структура, която отразява тяхната йерархичност. Атрибутите на всяко понятие и екземпляр се изобразяват в дървото като наследници на съответния възел на понятие или екземпляр. Аналогично параметрите на релациите се визуализират като наследници на релациите.

Йерархичната структура на понятията се определя от над-понятията на всяко понятие. Всяко понятие се изобразява в дървото като наследник на своите над-понятия. Това означава, че всяко понятие се среща в дървовидното представяне толкова на брой пъти, колкото над-понятия има то.

Екземплярите се показват в дървото като наследници на съответните понятия, на които те са екземпляри. Релациите се изобразяват в отделно поддърво. При избиране на елемент за него се изписва детайлна информация в полето под дървовидната структура.

*Фигура 4.2. Изглед на онтологиите с 2 заредени онтологии - DateTime и Locations.*

### 4.2.2. Изглед на свойствата (Object properties)

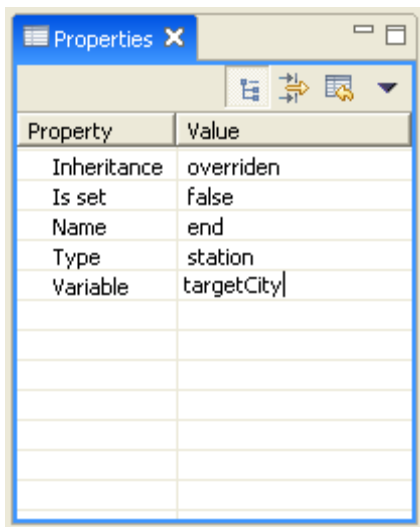

Този изглед осигурява визуализация и възможност за редактиране на свойствата на избрания обект от графичния модел на аксиомата. Обектът може да бъде избран или от модела на аксиомата, или от съкратения изглед.

*Фигура 4.3. Изглед на свойствата при избран обект – променлива от модела на аксиомата.*

### 4.2.3. Площ за редактиране

Площта за редактиране се използва за редактиране на графичния модел на аксиомата. Тя се използва за създаване на променливи, свързването им, конструиране на сложна логическа структура и т.н. Повечето елементи вътре в нея са представени чрез съставни фигури, които могат да бъдат произволно размествани заедно със съдържанието си от потребителя.

Връзките между елементите на модела са представени от стрелки в т.нар. **Манхатън стил** (с прави чупки). Връзките автоматично се преместват при местене на елементите от потребителя, за да запазят логическата структура на модела.

В лявата част на площта за редактиране се намира **палитрата с инструменти**, която се използва при редактиране в режим за напреднали (advanced mode).

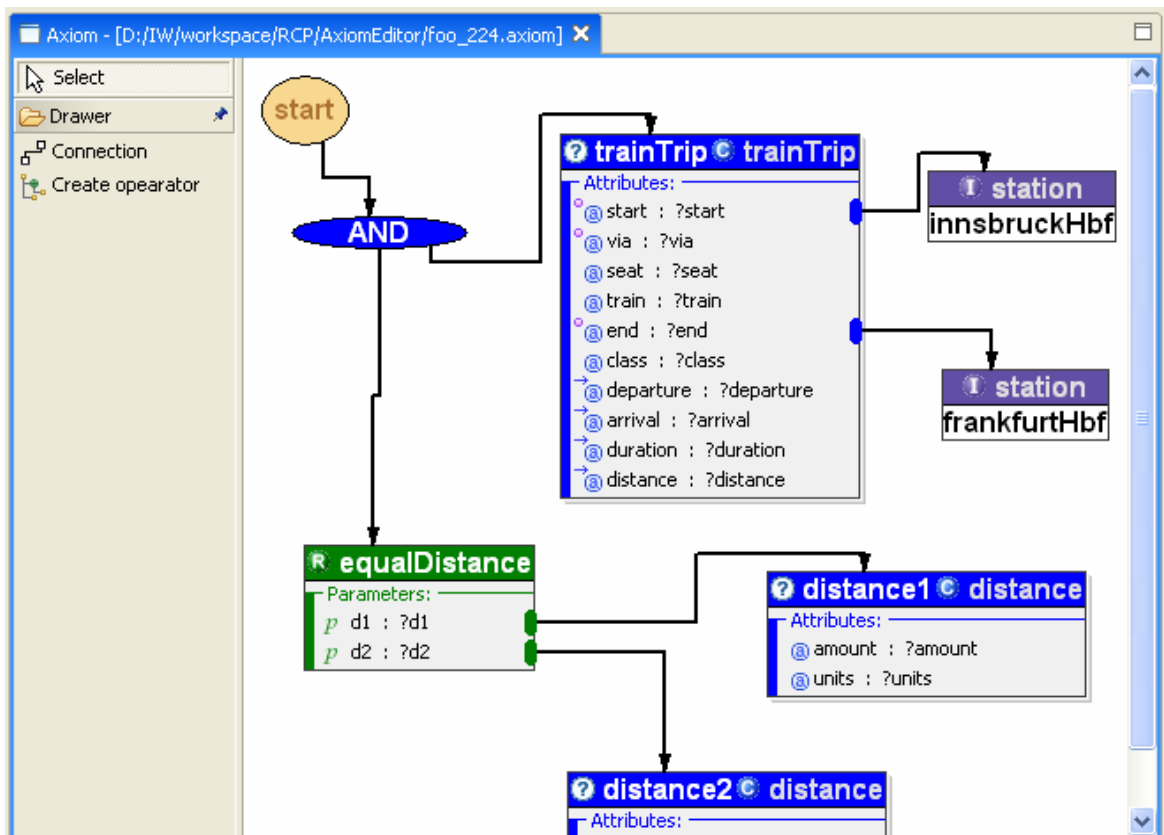

*Фигура 4.4. Графичен модел на аксиомата в площта за редактиране, заедно с палитра в лявата част*

### 4.2.4. Елементи на графичния модел

В тази секция описваме начина на изобразяване на всеки вид елементи на графичния модел. Тук понятието „фигура" се използва в смисъла на GEF фигурите, дефинирани в главата „Платформа за графично редактиране GEF". Повечето от елементите на графичния модел всъщност представляват съставни фигури, изградени от 2, 3 или повече други фигури. Като пример можем да посочим фигурата за променлива, която използва отделни фигури за всеки свой атрибут и фигура за името и типа си.

**Променливи**

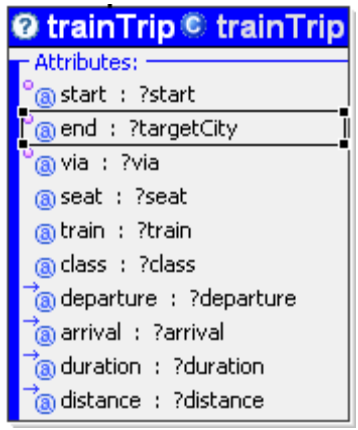

Променливата се визуализира като правоъгълник. В горния край се изписват името на променливата със знак (?) и името на нейния тип със знак (C). В останалата част от правоъгълника се изобразяват всички атрибути на понятието, което е тип на тази променлива. Всеки атрибут се изписва на един ред, като последователно се показват: тип наследяване, име на атрибута, име на променливата, с която той е уточнен/свързан. Типът наследяване може да приема три стойности, които се изобразяват с различни икони – собствен атрибут, наследен атрибут от над-понятие или наследен и предефиниран атрибут.

**Екземпляри на сложни типове от онгология**

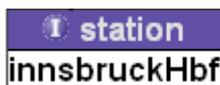

Екземплярът се визуализира като правоъгълник, който е разделен хоризонтално на две части. В горната част се изписва името на понятието, което е тип на този екземпляр. В долната част се изписва името на самия екземпляр.

**Екземпляри на WSML вградени типове**

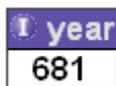

Аналогично на екземпляри на сложни типове, но с тази разлика, че тук в долната част се изписва директно стойността на екземпляра. Тази стойност може да се редактира от потребителя.

**Релация**

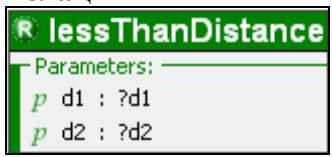

Релацията се визуализира като правоъгълник. В горния край се изписват името на релацията със знак (R). В останалата част от правоъгълника се изобразяват всички параметри на релацията. Всеки параметър се изписва на един ред, като последователно се изобразяват името на параметъра и името на променливата, с която той е уточнен/свързан.

**Логически оператор**

Логическият оператор се визуализира като елипса, в която е написан видът на оператора (AND, OR, NOT). В оператора може да влиза точно една връзка и могат да излизат две или повече връзки (за AND и OR) или точно една връзка (за NOT).

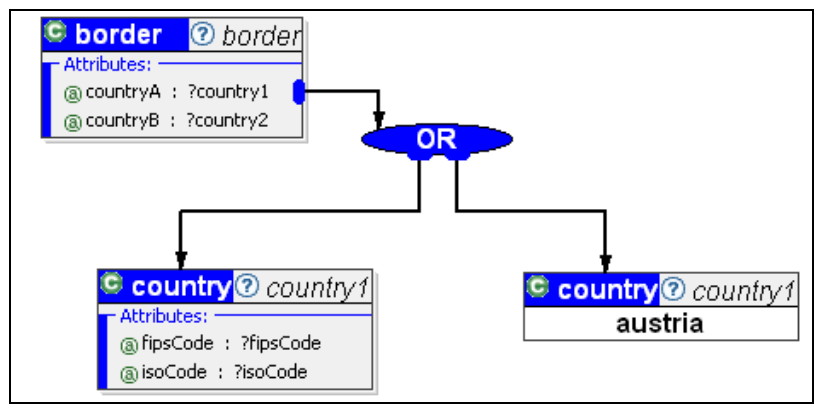

**Диалог за избор на елемент от онтологиите, който е съвместим по тип с дадено понятие**

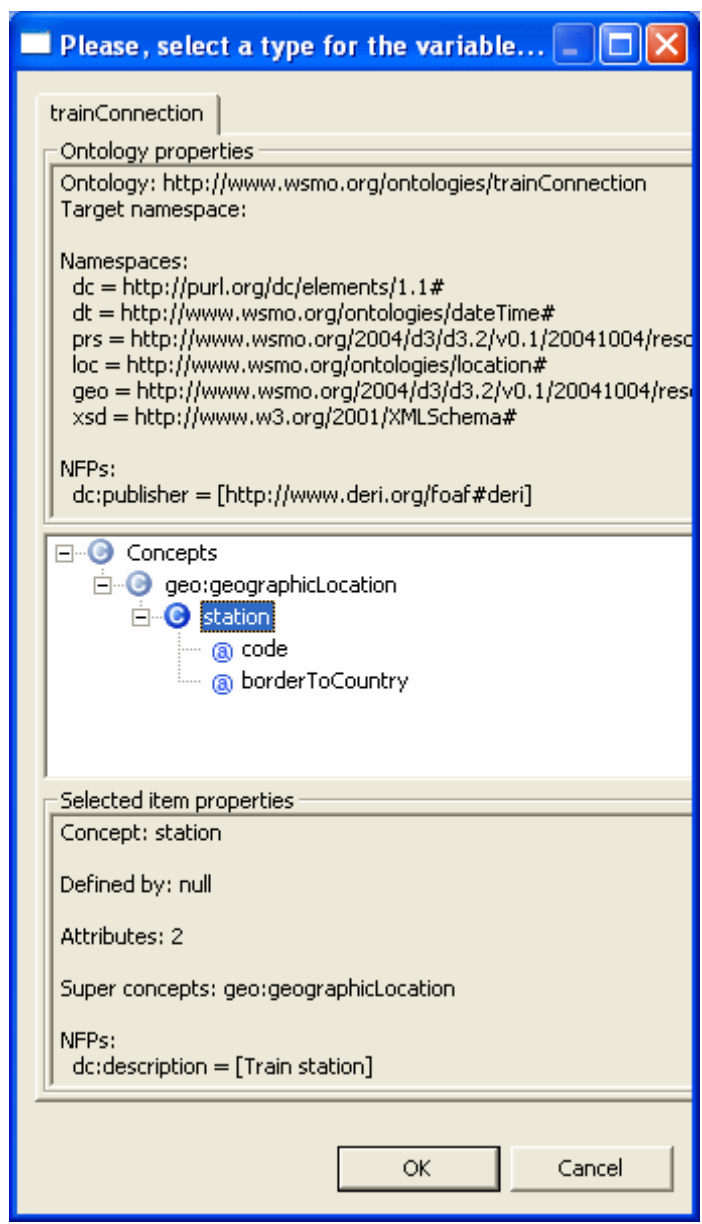

Този диалог показва филтрирано копие на хранилището за онтологии, в което се съдържат само понятия. При това в него се визуализират само тези понятия, които са съвместими по тип с дадено понятие. За по-голяма информативност в диалога се показват и атрибутите на понятията, но не могат да се избират. Диалогът се затваря, когато потребителят избере едно от понятията или даде отказ (Cancel).

**Диалог за избор на екземпляр от онтологиите, който е съвместим по тип с дадено понятие**

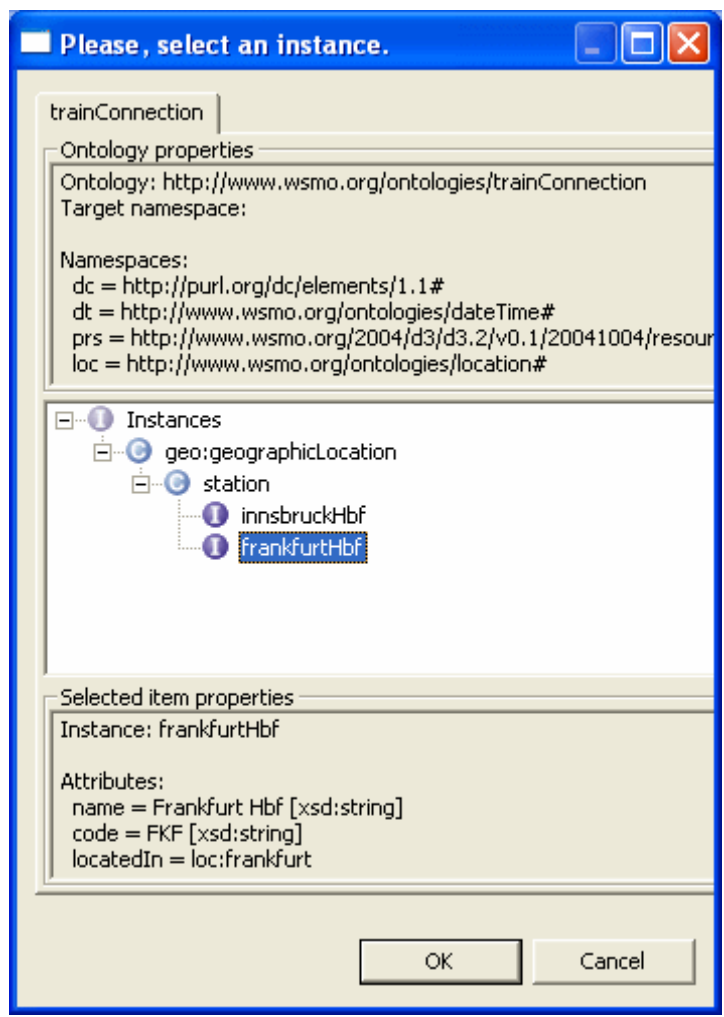

Този диалог показва филтрирано копие на хранилището за онтологии, в което се съдържат само екземпляри. При това в него се визуализират само тези екземпляри, които са съвместими по тип с дадено понятие. За по-голяма информативност в диалога се показват и понятията, на които принадлежат показаните екземпляри, но те не могат да се избират. Диалогът се затваря, когато потребителят избере един от екземплярите или даде отказ (Cancel).

**Диалог за избор на релация от онтологиите**

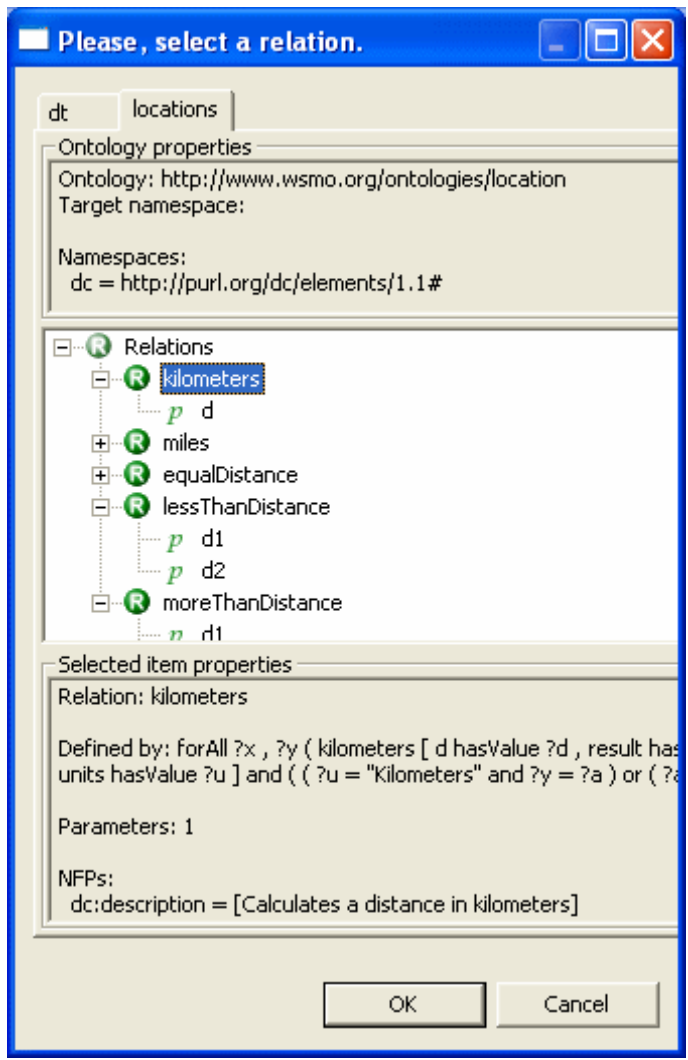

Този диалог показва филтрирано копие на хранилището за онтологии, в което се съдържат само релации. За по-голяма информативност в диалога се показват и параметрите на релациите. Диалогът се затваря, когато потребителят избере една от релациите или даде отказ (Cancel).

**Контекстно-зависимо меню с операции според избрания елемент от модела**

Операциите, които бяха описани подробно в разработката на Axiom Editor, се изпълняват върху различни елементи от модела на аксиомата. В имплементацията това е реализирано чрез контекстно-зависими изникващи (pop-up) менюта, които се извикват с натискане на десния бутон на мишката. По този начин за всеки избран елемент се появяват само тези операции, които са приложими за него и то в точно този конкретен момент. На фигура 4.5 се вижда контекстно-зависимото меню на един атрибут, който все още не е уточнен/свързан с нищо.

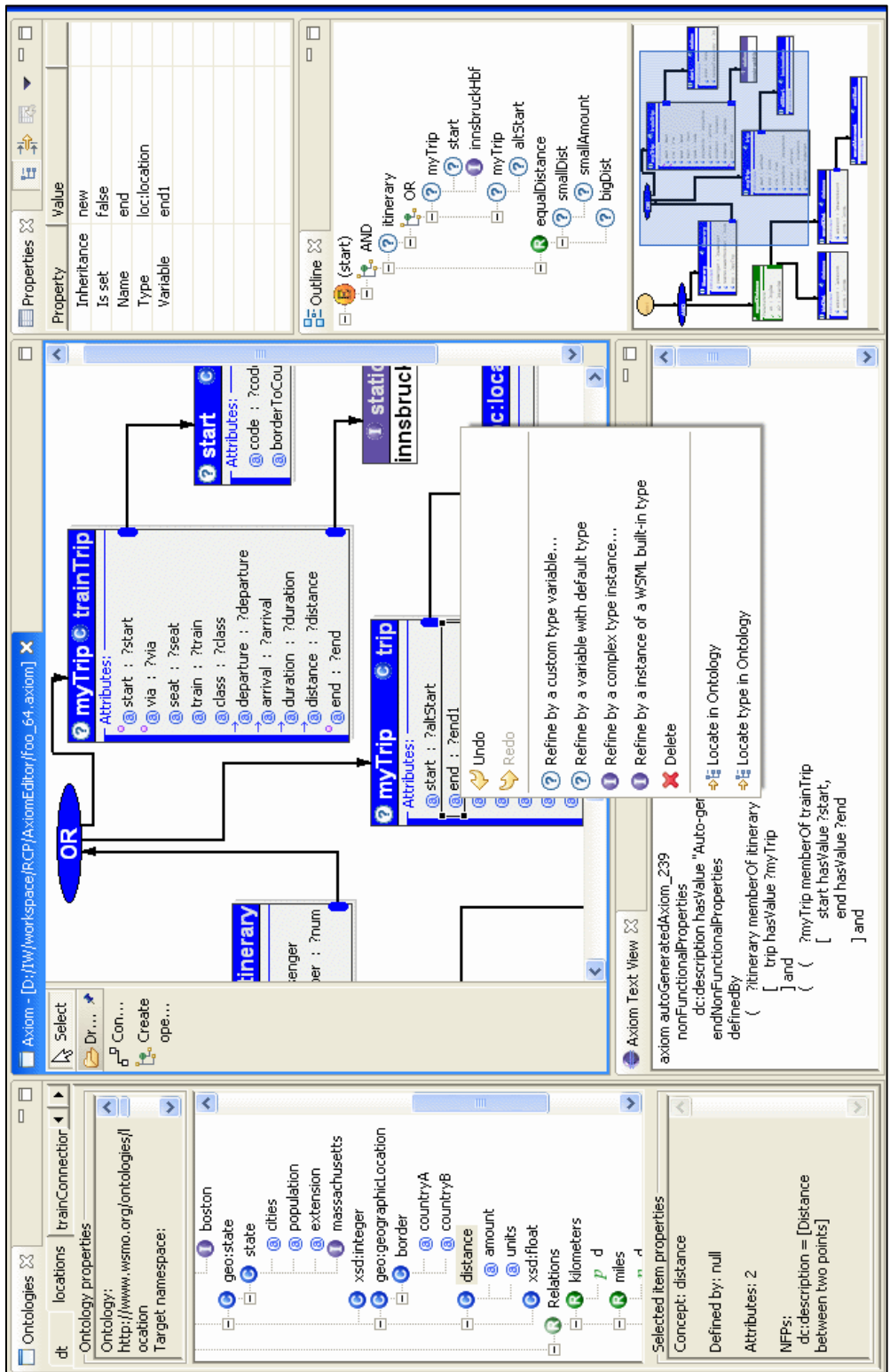

*Фигура 4.5. Контекстно-зависимо меню с операции според избрания елемент от модела*

### 4.2.5. Съкратен изглед на аксиомата (Axiom outline)

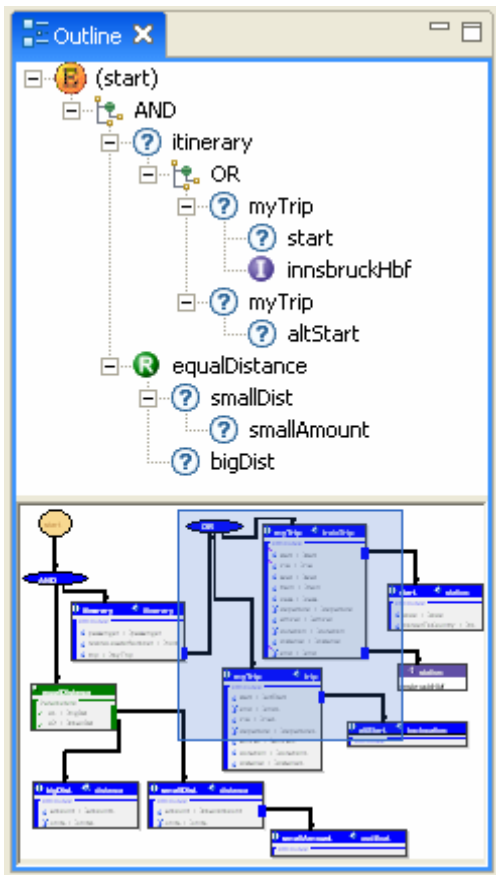

В този изглед се помества по-компактно представяне на графичния модел на аксиомата. Този изглед се състои от 2 части: текстова и графична.

Текстовата част съдържа дървовидно представяне на конструираната до момента аксиома (логически израз), като операторите се използват във възлите, а понятията и екземплярите са в листата на дървото.

Графичната част показва миниатюра (thumbnail) на целия графичен модел на аксиомата и служи за лесна навигация при по-сложни логически изрази, които не се събират на един екран и се налага скролиране.

*Фигура 4.6. Съкратен изглед на аксиомата*

## 4.2.6. Текстов изглед на аксиомата (Axiom text view)

В този изглед се появява автоматично генерирания текст на логическия израз (аксиомата). Представя се в чист текстов вид, като за момента се поддържа генериране на езика WSML (виж Фигура 4.7). По принцип е възможно в бъдеще да се добави поддръжка и за други езици, но само при условие, че и те следват WSMO методологията.

Голямото удобство на този изглед идва от факта, че то може да се генерира в реално време едновременно с протичащото в също време редактиране на модела на аксиомата. По този начин по-напредналите потребители могат „изкъсо" да следят работата на редактора и генерирания изход. В също време начинаещите потребители пък могат да се обучават в тънкостите на WSML, стига да следят внимателно как промените по модела се отразяват на генерирания текст.

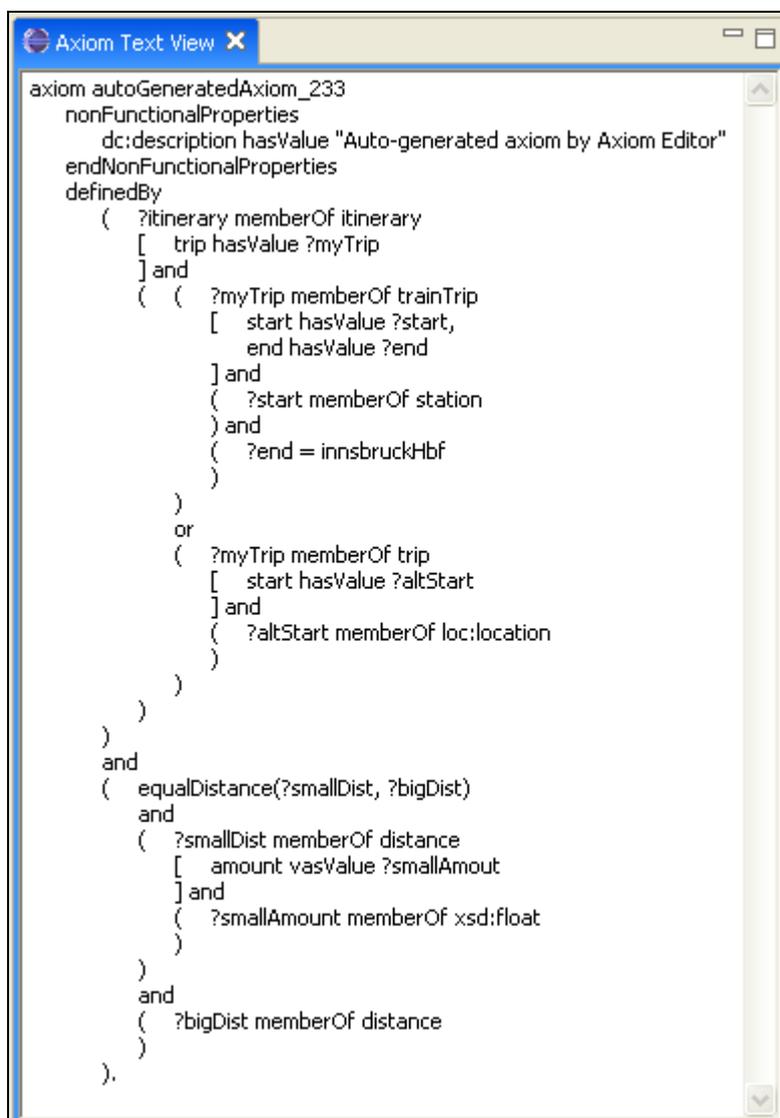

*Фигура 4.7. Текстов изглед на аксиомата*

Показаният автоматично генериран текст по-горе съответства на следния графичен модел, създаден с редактора:

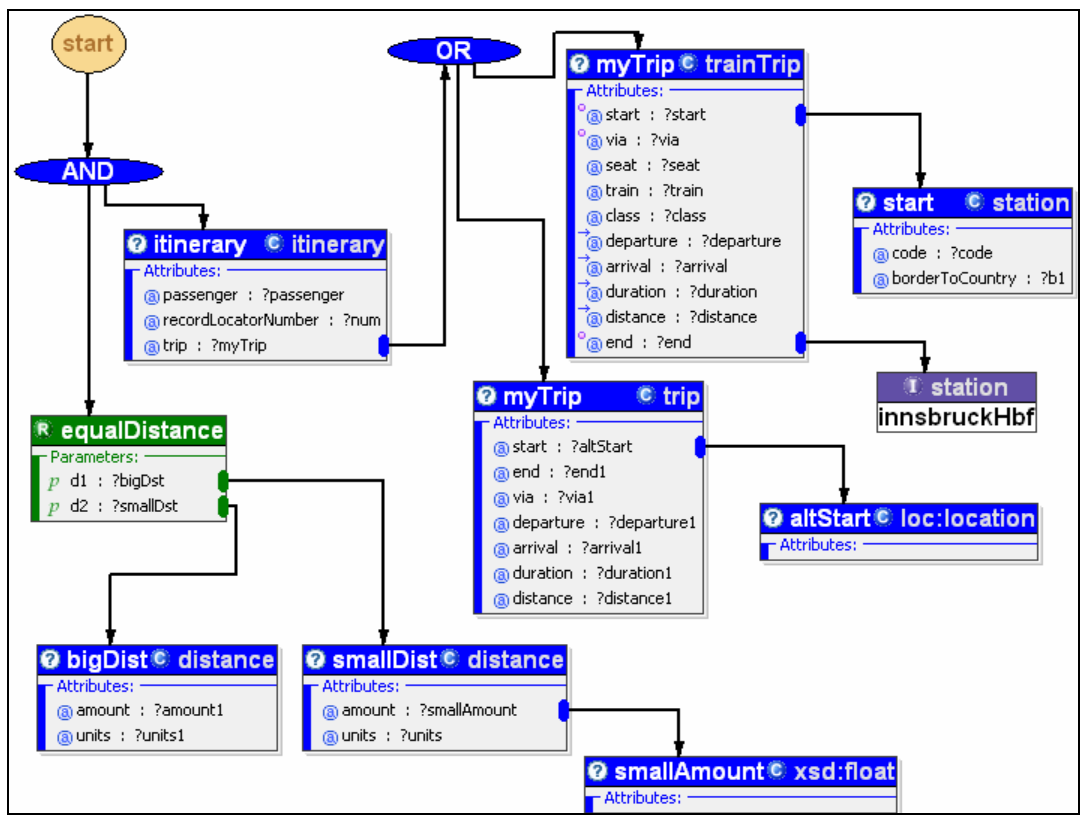

При създаване на нова аксиома, когато в графичния модел все още няма никакви елементи, автоматично генерираният текст на аксиомата изглежда така:

```
axiom autoGeneratedAxiom_1
   nonFunctionalProperties
         dc:description hasValue "Auto-generated axiom by Axiom Editor"
   endNonFunctionalProperties
   definedBy
    .
```

## *4.3. Имплементация на функционалността*

Тази секция съдържа обзор на имплементацията на функционалността на INFRAWEBS Axiom Editor. Реализацията е написана изцяло на езика java. За реализирането на нужната функционалност са създадени *над 150 класа*, които са организирани логически в *над 20 пакета*. Поради големия обем на изходния код на проекта, тук няма да се спираме в детайли на всички конкретни класове и методи. Ще опишем подробно само пакетите, в които са групирани класовете и ще акцентираме само върху няколко по-важни класа.

### **4.3.1. Дървовидна структура на файловете с изходния код**

Общо в имплементацията има *над 200 файла*, разпределени в *над 30 директории*.
Ето как изглежда дървовидната структура на проекта:

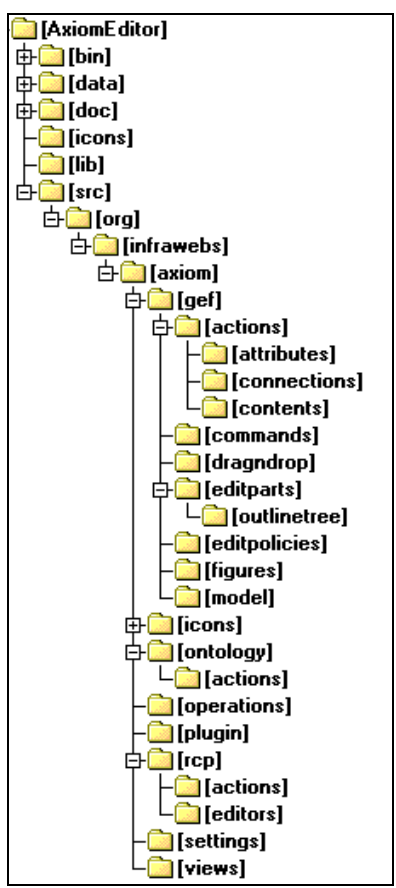

## 4.3.2. Логическа структура на пакетите с класове

Аналогично на файловата структура, класовете в проекта са подредени в логически пакети, чиято йерархия на най-високото ниво е следната:

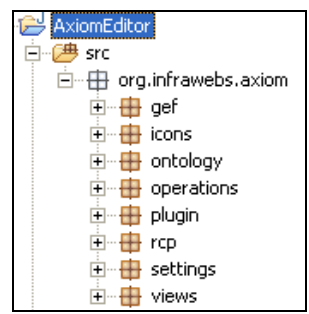

В следващата таблица има съвсем кратко описание на пакетите и функционалността, която те изпълняват

| Пакет | Функционалност |
|---|---|
| org.infrawebs.axiom.gef | Графичен редактор за аксиомния модел |
| org.infrawebs.axiom.gef.actions | Операции в графичния редактор |
| org.infrawebs.axiom.gef.actions.attributes | Контекстни операции върху атрибути |
| org.infrawebs.axiom.gef.actions.connections | Контекстни операции върху връзки |
| org.infrawebs.axiom.gef.actions.contents | Контекстни операции върху самия модел |
| org.infrawebs.axiom.gef.commands | Команди за редактора GEF |
| org.infrawebs.axiom.gef.dragndrop | Drag & drop функционалност |
| org.infrawebs.axiom.gef.editparts | Контролери за редактора GEF |
| org.infrawebs.axiom.gef.editparts.outlinetree | Контролери за съкратения изглед |
| org.infrawebs.axiom.gef.editpolicies | Политики за редактора GEF |
| org.infrawebs.axiom.gef.figures | Фигури за графично представяне в GEF |
| org.infrawebs.axiom.gef.model | Модел на аксиомите за GEF |
| org.infrawebs.axiom.icons | Изображения, използвани в редактора |
| org.infrawebs.axiom.ontology | Хранилището за онтологии |
| org.infrawebs.axiom.ontology.actions | Операции върху хранилището |
| org.infrawebs.axiom.operations | Генериране на текста на аксиомата и др. |
| org.infrawebs.axiom.plugin | Plug-in реализация за Eclipse |
| org.infrawebs.axiom.rcp | Имплементация на Rich Client Platform |
| org.infrawebs.axiom.rcp.actions | Действия за RCP |
| org.infrawebs.axiom.rcp.editors | Редактори за RCP |
| org.infrawebs.axiom.settings | Настройки на редактора |
| org.infrawebs.axiom.views | Създаване на изгледите в интерфейса |

## Пакет GEF

В този пакет са събрани всички класове, които имплементират основната част от редактора, а именно – *графичната област за редактиране на модела на аксиомата*. В него са реализирани всички техники, които бяха описани в главата „Платформа за графично редактиране GEF".

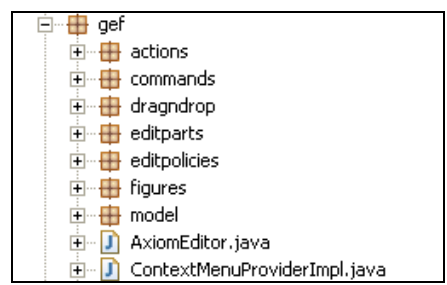

Ще обърнем внимание на два важни класа в този пакет.
Единият е *AxiomEditor* и представлява имплементация на обвивката на графичния редактор. Тази обвивка съдържа в себе си целия механизъм на GEF редактора.
Вторият е *ContextMenuProviderImpl* и представлява имплементация на контекстно-зависимите операции върху модела в GEF редактора.

## Пакет Actions

В този пакет са събрани всички класове, които имплементират *контекстно-зависимите операции* върху обектите от модела на аксиомата. Те са групирани в няколко под-пакета: за операции над атрибути, връзки, операции над основния модел, над онтологиите и базови операции.

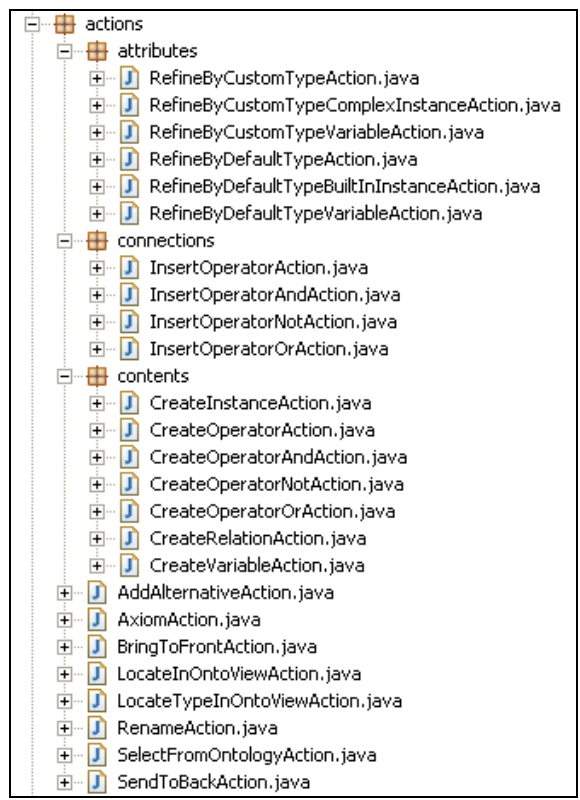

## Пакет Commands

В този пакет са събрани всички класове, които имплементират GEF *командите*, които бяха описани в главата „Платформа за графично редактиране GEF". Те включват имплементация на команди за елементи на модела (обекти и връзки между тях), операции по контейнери в модела, операции за уникалното именуване на променливите и др.

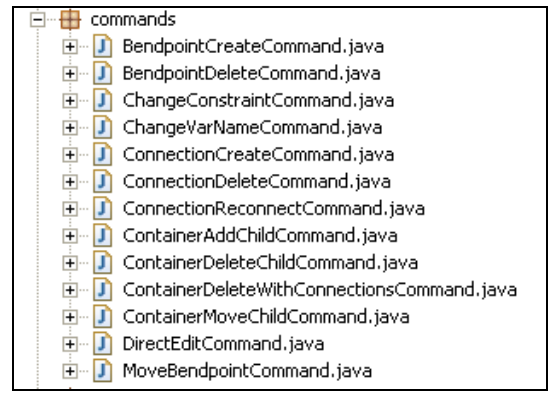

## Пакет EditParts

В този пакет са събрани всички класове, които имплементират *GEF контролерите от Модел-Изглед-Контролер шаблона*, който беше описан в главата „Платформа за графично редактиране GEF".

Контролерите са разделени на два вида. Едните се използват за графичния модел на редактора, където се извършват всички основни операции. Другите се използват за дървовидното представяне на модела в съкратения изглед (Outline), където също могат да се прилагат операции.

Вижда се, че този пакет е най-обемист, тъй като в него се съдържа основния код за управление на поведението на редактора и той става предпоставка за реализиране на всички операции над модела.

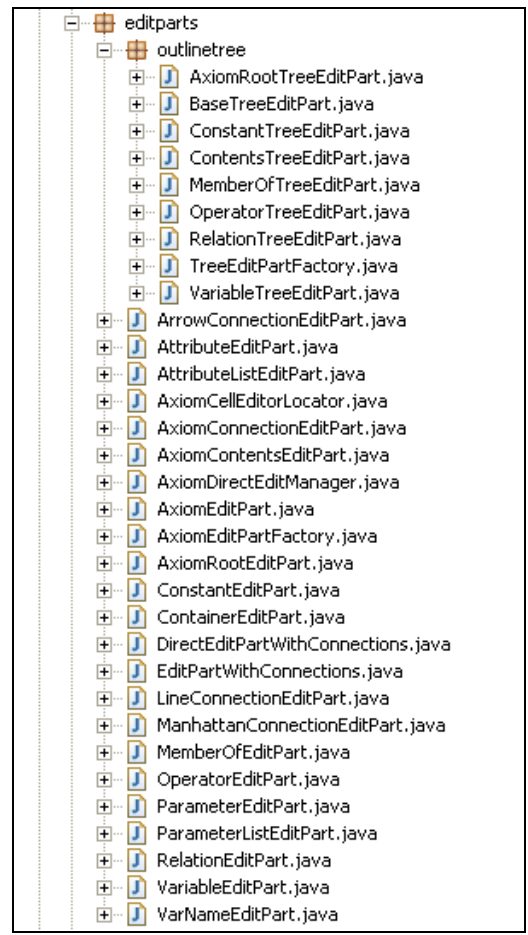

## Пакет EditPolicies

В този пакет са събрани всички класове, които имплементират GEF *политиките*, които бяха описани в главата „Платформа за графично редактиране GEF". Те включват имплементация на политики за създаване, изтриване, подреждане, редактиране и други операции над модела.

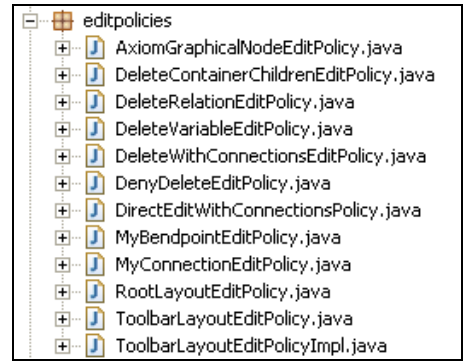

## Пакет Figures

В този пакет са събрани всички класове, които имплементират *Draw2D фигурите*, които служат за визуализация, както беше описано в главата „Платформа за графично редактиране GEF". Те включват имплементация на фигурите за понятия, атрибути, екземпляри, релации, параметри, оператори, имена на променливи и др.

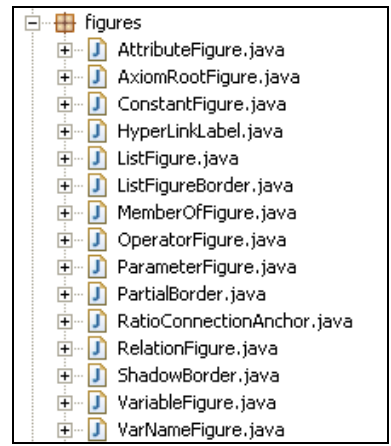

## Пакет Ontology

В този пакет се намира имплементацията на изгледа за онтологии и всички операции над него.

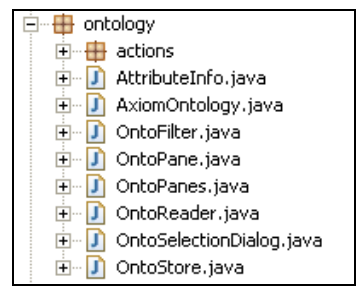

## Пакет Operations

В този пакет се намира имплементацията на най-важните операции в редактора, а именно: генериране на текста на аксиомата, изпълняване на операции върху модела и всички операции по проверка на семантичната съгласуваност на модела.

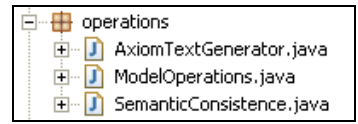

## Пакети Plugin и RCP

В този пакет се намира имплементацията на всичко, свързано с начина на изграждане на редактора като приставка (plug-in) на Eclipse или като отделно напълно функционално приложение на база на платформата Rich Client Application на Eclipse.

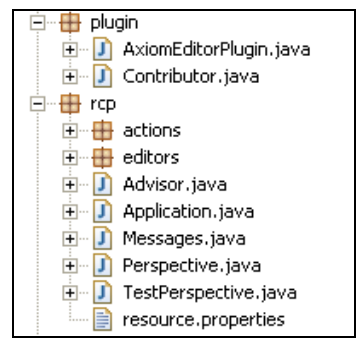

В пакета RCP се намират още два други пакета. Единият е Actions и в него са дефинирани всички операции върху цели аксиоми (отваряне, създаване на нова и др.). Другият е Editors и в него са включени допълнителни редактори, които могат да се използват в RCP приложението, като например прост текстов редактор.

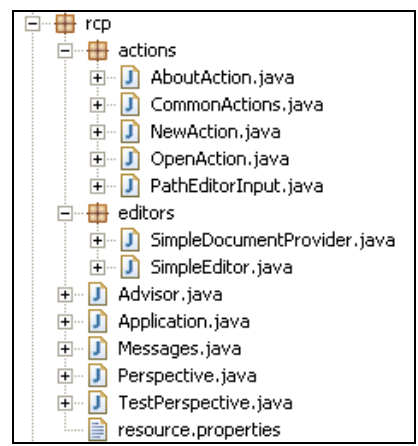

## Пакети Settings и Views

В този пакет се намират имплементациите на свързаните с изгледите функции, настройките на цветовете, начинът на извеждане на отделните прозорци в редактора и др.

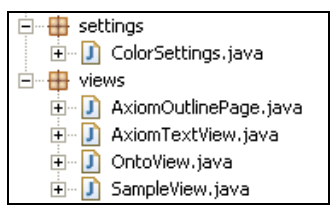

## Най-главният клас - AxiomEditor

Класът AxiomEditor всъщност капсулира имплементацията на един цял графичен редактор на логически изрази. Неговото място в йерархията на класовете в имплементацията е следното:

```
java.lang.Object
  └org.eclipse.ui.part.WorkbenchPart
      └org.eclipse.ui.part.EditorPart
          └org.eclipse.gef.ui.parts.GraphicalEditor
              └org.eclipse.gef.ui.parts.GraphicalEditorWithPalette
                  └org.infrawebs.axiom.gef.AxiomEditor
```

По време на работата на RCP приложението могат да бъдат инстанцирани много обекти от този клас, като всеки един заема цял нов прозорец. Това позволява едновременната работа върху няколко аксиоми в редактора.

В *Приложение 1* има включен нагледен пример за работа с Axiom Editor. В него е показано конструирането на една аксиома стъпка по стъпка заедно с описание на извършваните операции.

# Възможности за бъдещо развитие

В съответствие с целта на дипломната работа, успешно беше имплементирано средство за създаване на сложни логически изрази за семантично описание на мрежови услуги.

В уводната част направихме кратко въведение в предметната област и показахме необходимостта от решение на съществуващите проблеми. Описахме проекта INFRAWEBS и уточнихме мястото, което настоящата разработка заема в него. В края на въведението дефинирахме точно и ясно целта на дипломната работа.

В хода на проектирането направихме анализ на изискванията към редактора и предложихме един възможен процес за създаване на аксиоми, както и потребителски интерфейс за неговото визуализиране. В разработката на Axiom Editor описахме детайлно операциите върху хранилището за онтологии и операциите за създаване на аксиоми.

В частта за имплементация обосновахме избора на средства за написване на редактора, описахме реализирания потребителски интерфейс и посочихме функционалността на пакетите с класове от имплементацията. В *Приложение 1* дадохме пример за работата с Axiom Editor, който може да служи и за потребителска документация. В *Приложение 2* навлезнахме по-дълбоко в използваните техники за имплементиране на графичния редактор, базиран на платформата GEF.

Проектът INFRAWEBS е дългосрочен проект. Общата му продължителност е около 28 месеца. До момента на завършване на дипломната работа от това време са изминали едва 11 месеца, а вече първите резултати са налице. Те станаха основа и за публикация в бюлетина на международната конференция i.TECH 2005, която се проведе през юни 2005 година във Варна [Agre et al., 2005].

Текущата имплементация на Axiom Editor е напълно функционираща и с нея могат да се създават сложни логически изрази. Поддържа се хранилище от онтологии, решава се проблемът за именуване на променливите, максимално се улеснява потребителят при работа и в същото време се ограничава, за да не допуска грешки. Имплементирани са почти всички описани в изложението операции, заедно с контекстно-зависимите менюта за всеки един елемент от графичния модел. Автоматичното генериране на текста на аксиомата работи паралелно с процеса на нейното конструиране, като по този начин потребителят има непрекъснат поглед над крайния резултат.

Имплементацията има и редица недостатъци. Такъв недостатък безспорно е липсата на начин за записване и зареждане на графичния модел на аксиомата заедно с информация за използваните онтологии от хранилището. Работата по тази функционалност тепърва предстои. В момента се разработва идея за създаване на хранилище с аксиоми, което да предоставя удобен механизъм за търсене и повторно използване на вече конструирани логически изрази.

Друг недостатък на текущата имплементация е фактът, че тя може да работи само с локални онтологии от *.wsml файлове, записани в Склада за онтологии. Добре би било в бъдеще да се добави възможност за директна работа с публикувани в Интернет онтологии. За да се реализира това, трябва да се решат множество проблеми с начина на достъп, наличието на различни версии на една и съща онтология и т.н.

В момента редакторът поддържа само три логически оператора: AND, OR и NOT. Една насока за бъдещо развитие е добавянето на още оператори, по-сложни, които да се използват само в режима за напреднали. Могат да се добавят в графичния модел и нов вид обекти – функции, които да се използват по подобен на релациите начин. Особеното при функциите е, че те връщат резултат, който може да бъде присвоен на променлива и използван на други места в логическия израз.

Трябва да се мисли в бъдеще и за интеграция на Axiom Editor с други програми. Платформата Eclipse предоставя добра възможност за това. Например, възможно е да се комбинира редакторът с инструмент за създаване на онтологии и с инструмент за публикуване на семантични мрежови услуги.

Крайната цел за бъдещо развитие на INFRAWEBS Axiom Editor е неговото разширяване и превръщане в пълнофункционален INFRAWEBS Semantic Web Service Designer, който с използване на допълнителна информация от WSDL описания на съществуващи мрежови услуги да помага за преобразуването им в семантични мрежови услуги.

# Литература

## *Проложение 1. Пример за работа с Axiom Editor*

Това приложение показва начина на работа с редактора Axiom Editor. Ще създадем една примерна аксиома с него стъпка по стъпка, ще опишем извършваните операции и ще показваме автоматично генерирания текст на аксиомата след всяка стъпка.

### *Стъпка 1*

Създаваме нова празна аксиома, като избираме от менюто File -> New Axiom (Ctrl+N). Ето как изглежда графичният модел веднага след тази операция:

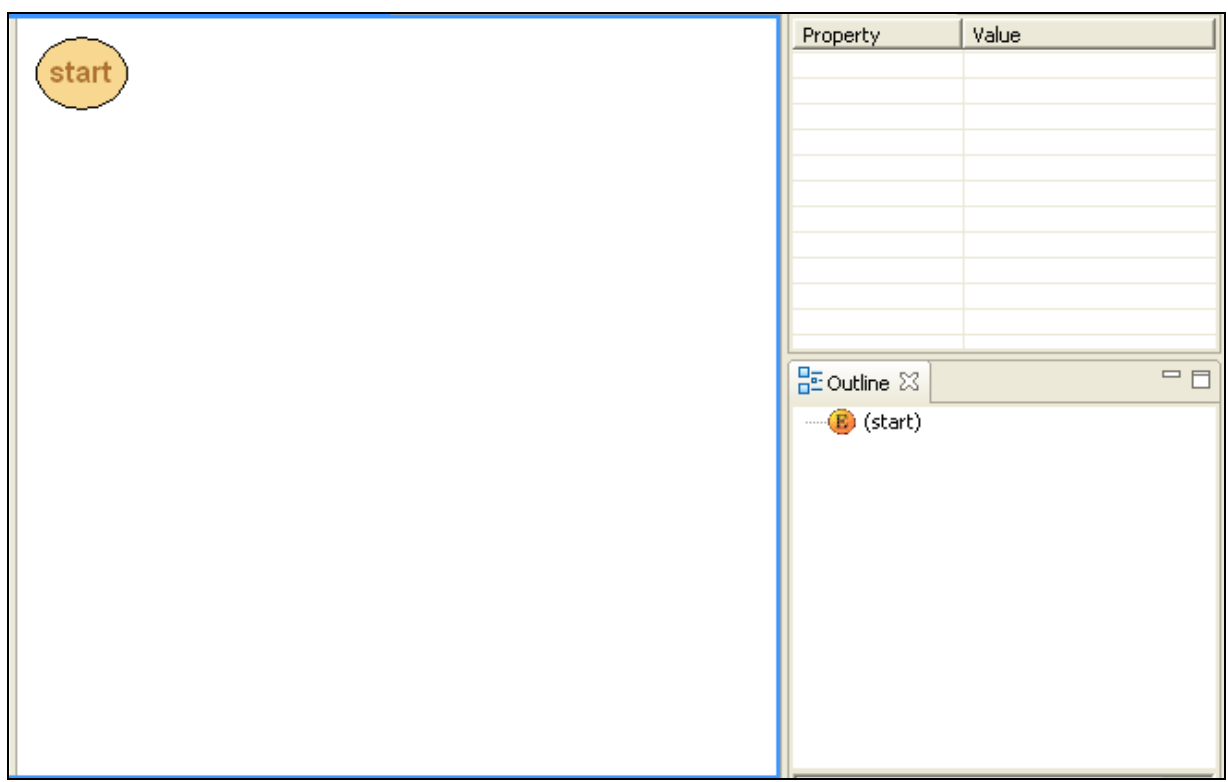

Ето автоматично генерираният текст на аксиомата от Axiom Editor в този момент:

```
axiom autoGeneratedAxiom_61
      nonFunctionalProperties
            dc:description hasValue "Auto-generated axiom by Axiom Editor"
      endNonFunctionalProperties
      definedBy
             .
```

### *Стъпка 2*

Зареждаме онтологиите *locations* и *trainConnection* от Склада за онтологии с операцията File -> Open…(Ctrl+O). Избираме понятието *itinerary* от онтологията *trainConnection* и правим двойно щракане с мишката върху него. Автоматично в модела на аксиомата се появява променлива с тип – избраното понятие, генерира се име за нея (*?itinerary*) и се свързва коренът на аксиомата (start) с нея.

Ето как изглежда графичният модел веднага след тази операция:

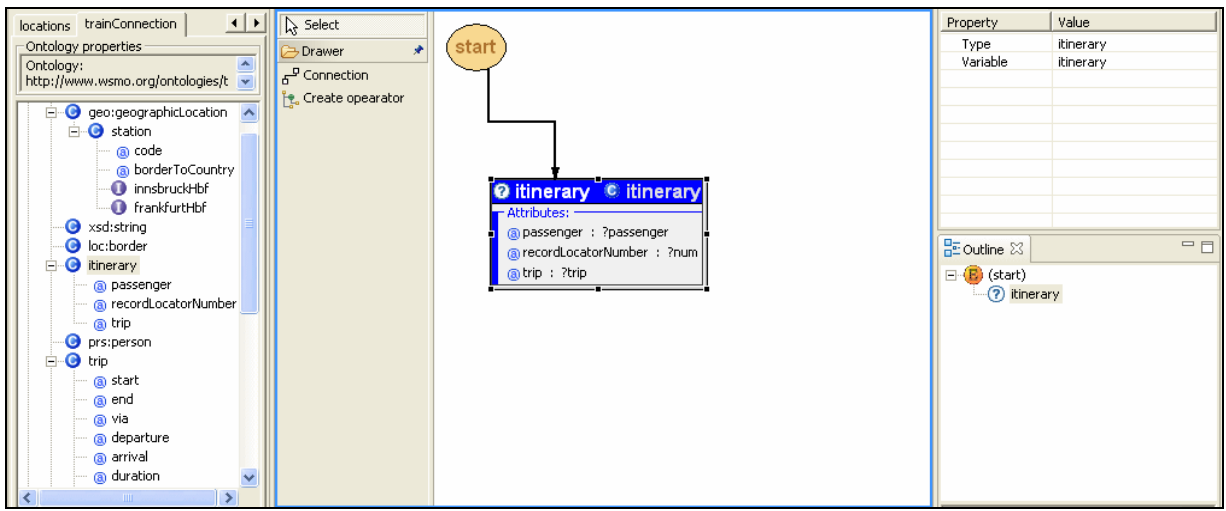

Ето как изглежда автоматично генерираният текст на аксиомата в този момент:

```
axiom autoGeneratedAxiom_61
      nonFunctionalProperties
            dc:description hasValue "Auto-generated axiom by Axiom Editor"
      endNonFunctionalProperties
      definedBy
            ?itinerary memberOf itinerary.
```

## *Стъпка 3*

Уточняваме атрибута *trip* на променливата *?itinerary*, като избираме от неговото контекстно меню точката "Refine by a custom type variable…". Ето как изглежда тази операция:

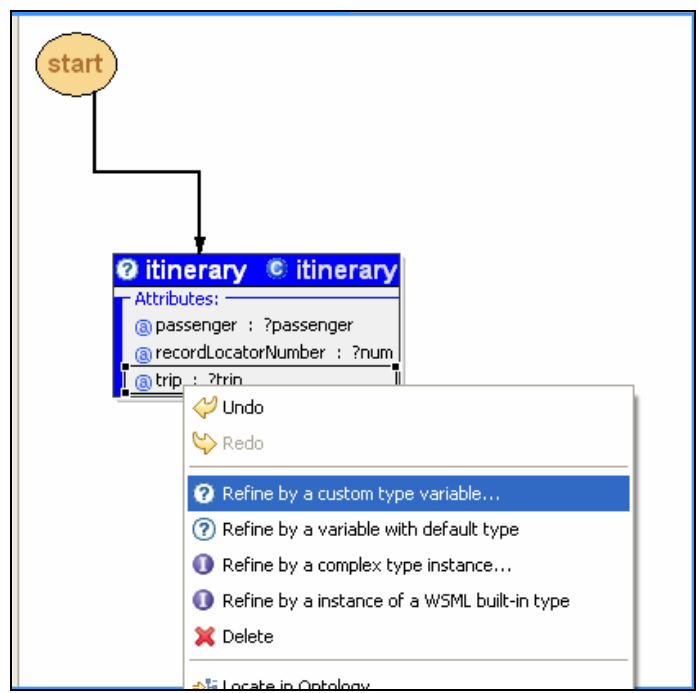

*Стъпка 4*

В този момент на екрана се появява диалогът за избор на понятие, което да се използва за тип на новата променлива, с която ще се свърже избраният атрибут:

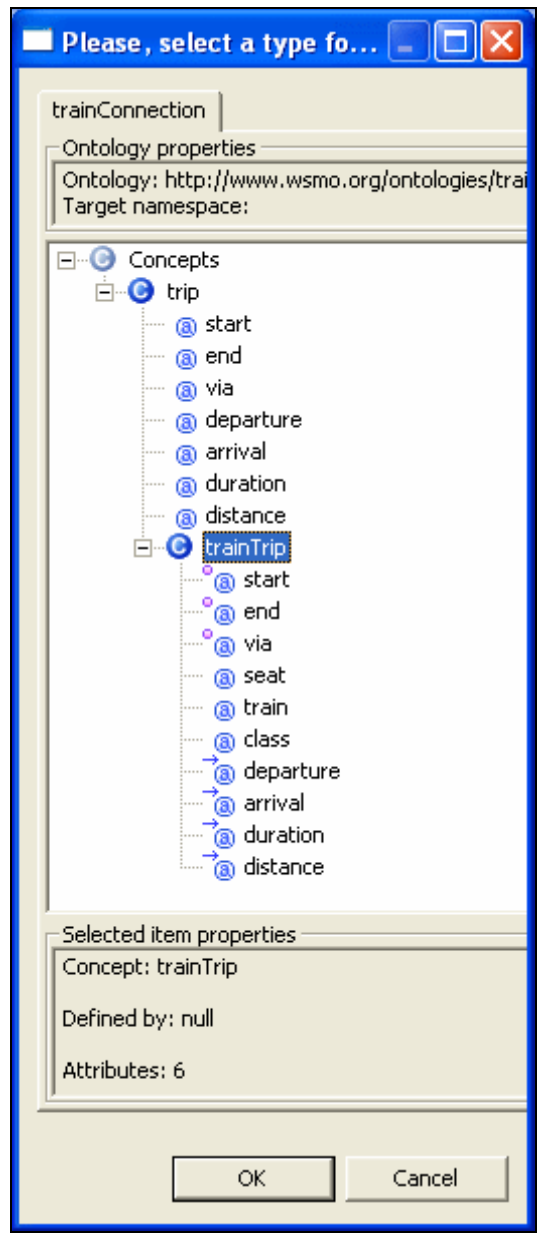

Избираме от диалога понятието *trainTrip*.

## Стъпка 5

След затварянето на диалога автоматично в модела се появява новата променлива и се създава връзка, започваща от избрания атрибут и завършваща в новата променлива:

Обръщаме внимание на името на новата променлива – то е генерирано от името на атрибута *trip* и има стойност "?trip".

Ето как изглежда автоматично генерираният текст на аксиомата в този момент:

```
axiom autoGeneratedAxiom_61
      nonFunctionalProperties
            dc:description hasValue "Auto-generated axiom by Axiom Editor"
      endNonFunctionalProperties
      definedBy
            ?itinerary memberOf itinerary
            [      trip hasValue ?trip
            ] and
            (      ?trip memberOf trainTrip
            ).
```

## Стъпка 6

Избираме да уточним атрибута *start* на новата променлива. От неговото контекстно меню избираме „Refine by a variable with default type" ето така:

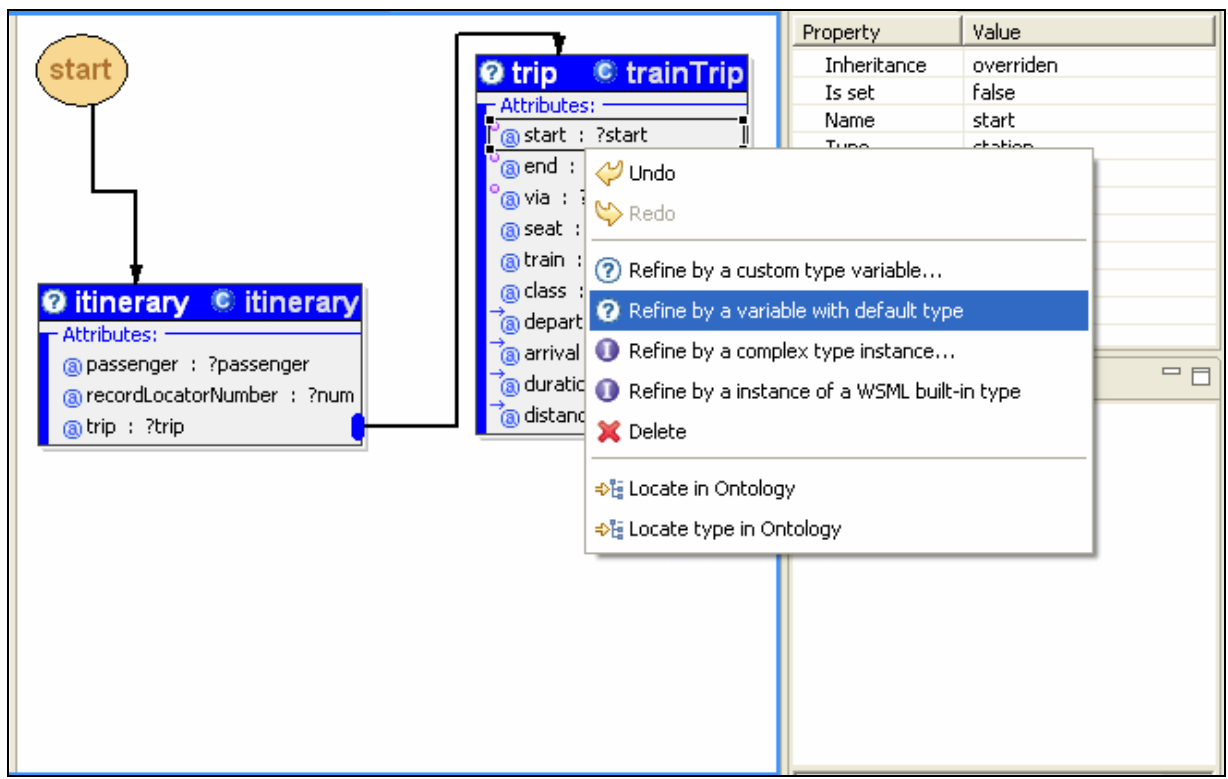

## Стъпка 7

Този път не се появява никакъв диалог за избор, а веднага се появява нова променлива, чийто тип е взет от типа на атрибута *start* (в случая той е *station*). Ето как изглежда моделът на аксиомата сега:

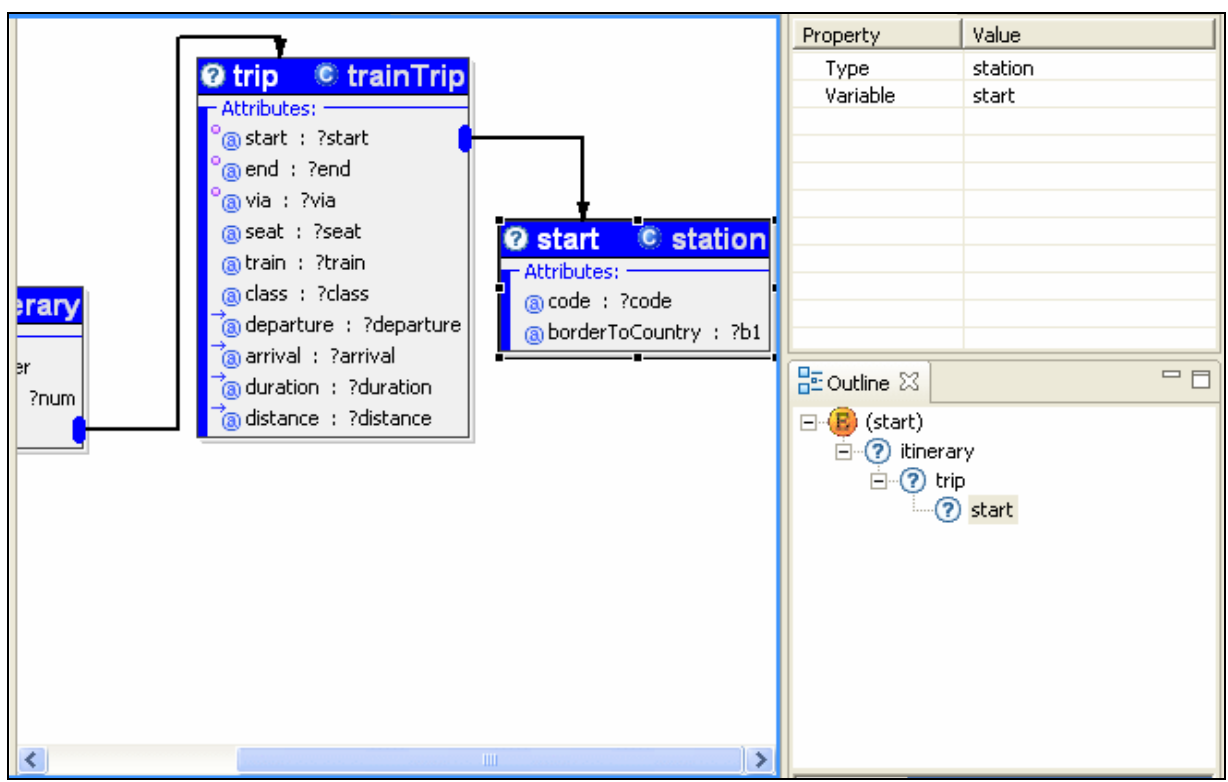

Ето как изглежда автоматично генерираният текст на аксиомата в този момент:

```
axiom autoGeneratedAxiom_61
     nonFunctionalProperties
          dc:description hasValue "Auto-generated axiom by Axiom Editor"
     endNonFunctionalProperties
     definedBy
          ?itinerary memberOf itinerary
          [     trip hasValue ?trip
          ] and
          (     ?trip memberOf trainTrip
                [     start hasValue ?start
                ] and
                (     ?start memberOf station
                )
          ).
```

## Стъпка 8

Избираме да уточним атрибута *end* с екземпляр. От неговото контекстно меню избираме „Refine by a complex type instance…" ето така:

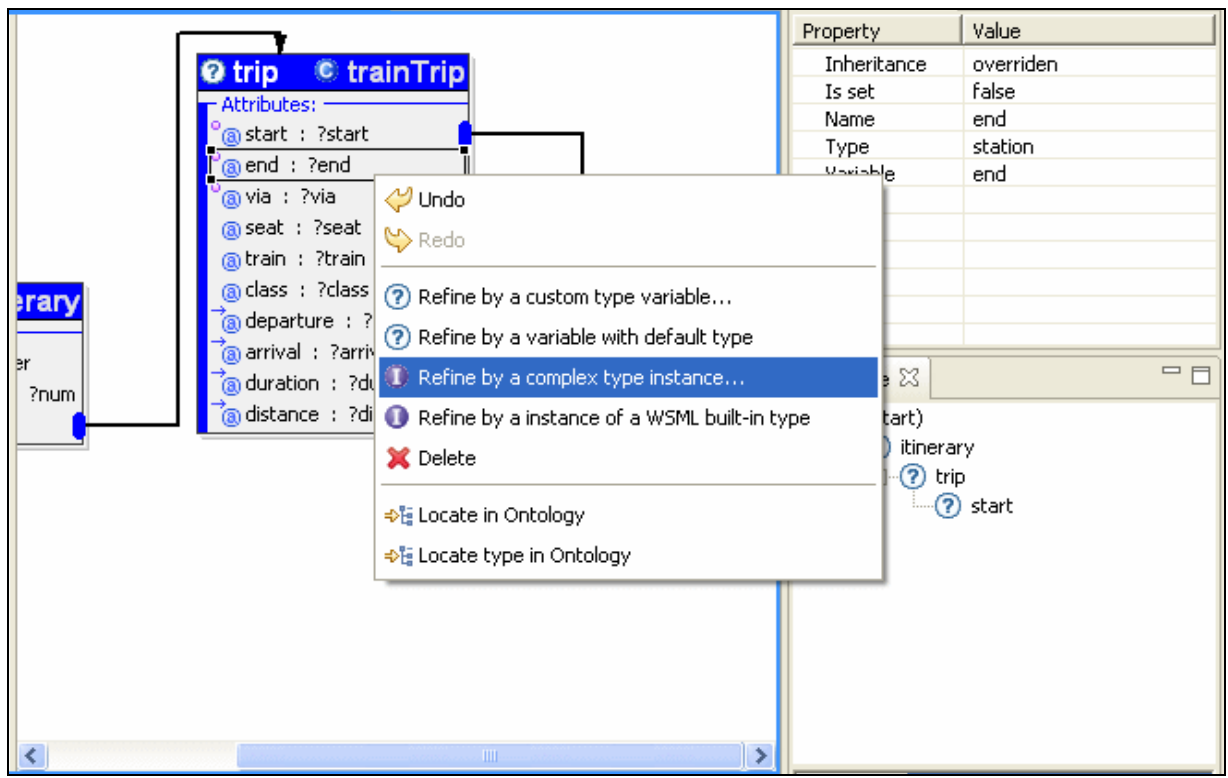

## Стъпка 9

В този момент се появява познатия диалог за избор, но този път в него са филтрирани само екземплярите, които са съвместими по тип с типа на избрания атрибут (*end*).

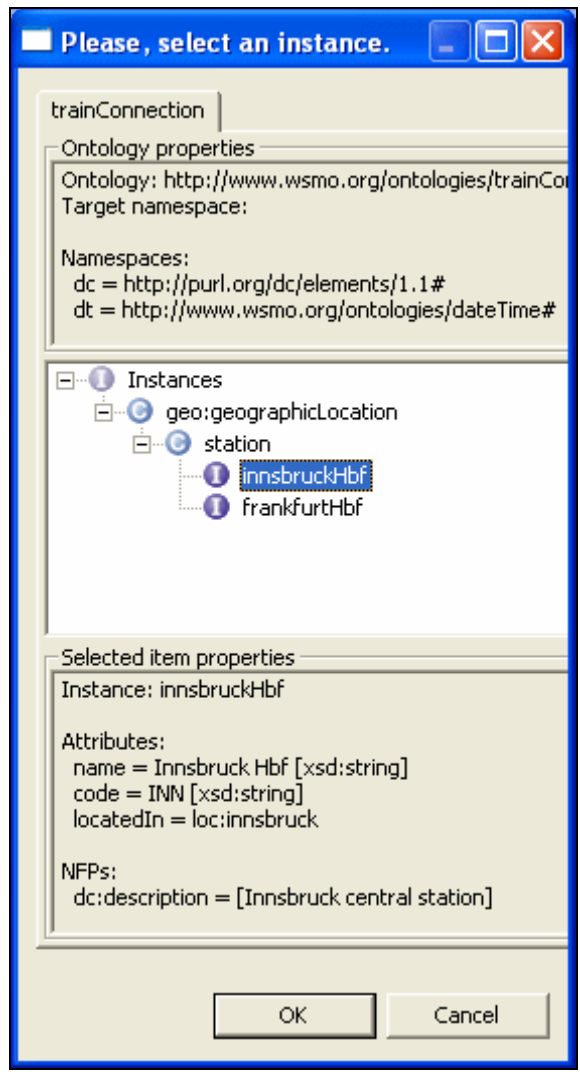

Избираме екземплярът *innsbruckHbf*.

## Стъпка 10

Axiom Editor автоматично създава елемент за този екземпляр в модела и го свързва с атрибута ето така:

*Стъпка 11*

Понеже от естетическа гледна точка не е приятно да се пресичат връзките в модела, затова ще направим разместване на реда на атрибутите. С операция drag & drop с мишката преместваме атрибута *end* по-надолу, за да стане по-прегледен модела:

## *Стъпка 12*

Ето как изглежда моделът на аксиомата сега:

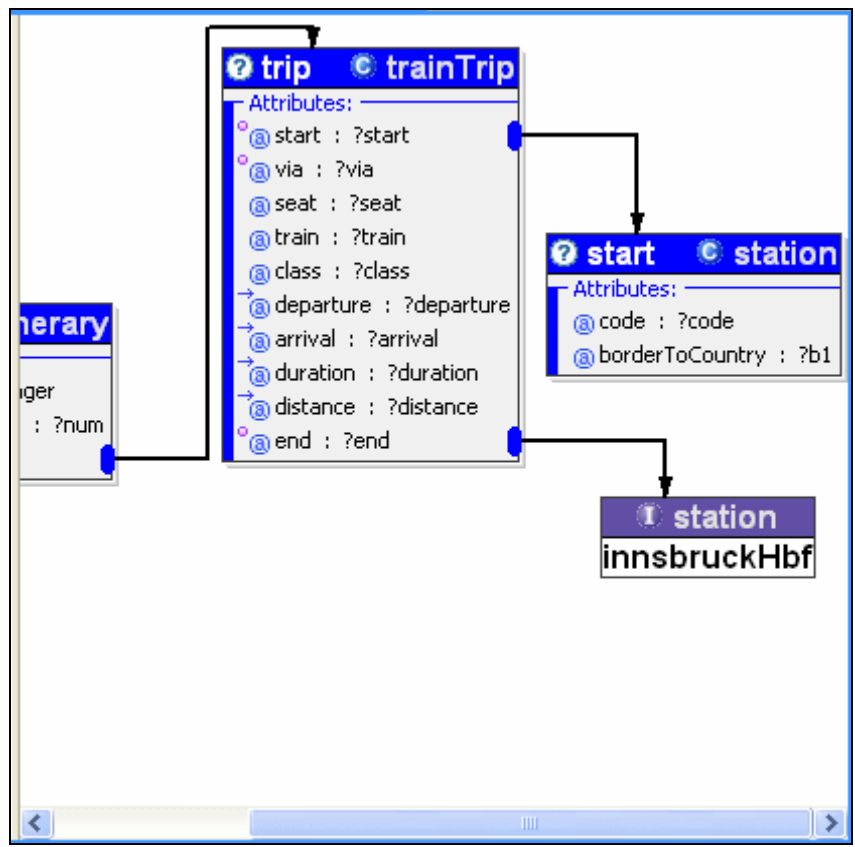

Ето и генерирания до момента текст:

```
axiom autoGeneratedAxiom_61
      nonFunctionalProperties
            dc:description hasValue "Auto-generated axiom by Axiom Editor"
      endNonFunctionalProperties
      definedBy
            ?itinerary memberOf itinerary
            [     trip hasValue ?trip
            ] and
            (     ?trip memberOf trainTrip
                  [     start hasValue ?start,
                        end hasValue ?end
                  ] and
                  (     ?start memberOf station
                  ) and
                  (     ?end = innsbruckHbf
                  )
            ).
```

*Стъпка 13*

Решаваме да развием логическата структура на аксиомата, като включим алтернатива за стойността на атрибута *trip*. За целта ще вмъкнем оператор OR по средата на връзката между атрибута *trip* и променливата *?trip*. Използваме контекстното меню на самата връзка ето така:

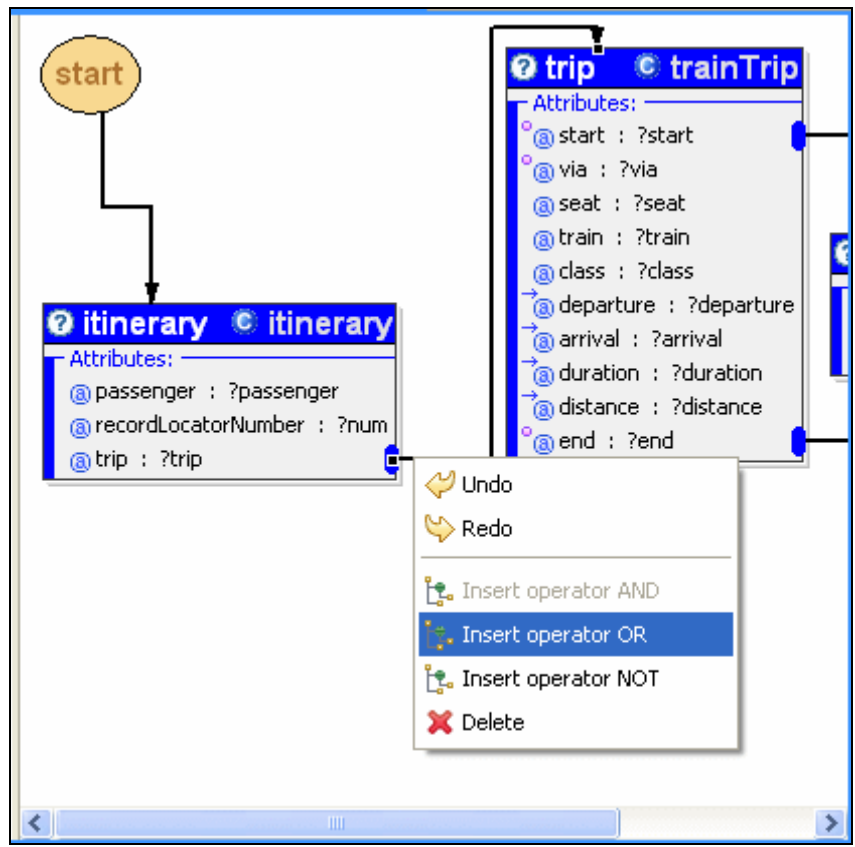

## Стъпка 14

За нова алтернатива ще изберем над-понятие на понятието *trainTrip*, а именно – *trip*. Това става от диалога за избор, в който се показват само съвместимите понятия с типа на атрибута:

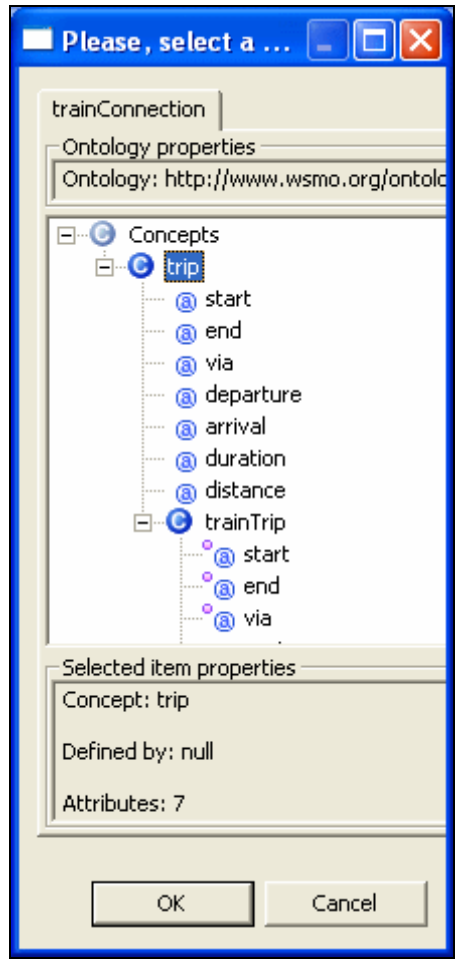

## Стъпка 15

Axiom Editor автоматично слага същото име *?trip* за новата променлива като името на старата алтернатива (другия операнд на оператора OR). Двете имена трябва задължително да са еднакви, защото представляват алтернатива за един и същи атрибут на променливата *?itinerary*.

Ето как изглежда моделът на аксиомата сега:

Ето и генерирания до момента текст:

```
axiom autoGeneratedAxiom_61
      nonFunctionalProperties
            dc:description hasValue "Auto-generated axiom by Axiom Editor"
      endNonFunctionalProperties
      definedBy
            ?itinerary memberOf itinerary
            [     trip hasValue ?trip
            ] and
            (     (     ?trip memberOf trainTrip
                        [     start hasValue ?start,
                              end hasValue ?end
                        ] and
                        (     ?start memberOf station
                        ) and
                        (     ?end = innsbruckHbf
                        )
                  )
                  or
                  (     ?trip memberOf trip
                  )
            ).
```

## Стъпка 16

Решаваме да уточним атрибута *start* на новата променлива *?trip*. Ще използваме уточняване с променлива от подразбиращия се тип. Използваме отново контекстните операции от менюто:

## Стъпка 17

В модела на аксиомата се появява променливата *?start1*, защото Axiom Editor е добавил автоматично число към името, за да не се дублира името с име на променливата *?start*, която вече фигурира в модела. Ние обаче искаме да сложим по-смислено име на новата променлива и с операция „Rename variable…" от контекстното меню извършваме преименуване на променливата направо в модела. Пишем за име *?altStart*. Ето как изглежда на екрана момента на редактиране (ограден е в кръгче):

## Стъпка 18

Axiom Editor автоматично разпространява редакцията, която направихме, навсякъде в модела. Променят се името на свързаната с атрибута променлива, съответното свойство с изгледа на свойствата (горе вдясно) и съкратения изглед на аксиомата (долу вдясно). С кръгчета са оградени тези места, които автоматично се обновяват:

Ето и генерирания до момента текст:

```
axiom autoGeneratedAxiom_61
      nonFunctionalProperties
            dc:description hasValue "Auto-generated axiom by Axiom Editor"
      endNonFunctionalProperties
      definedBy
            ?itinerary memberOf itinerary
            [     trip hasValue ?trip
            ] and
            (     (     ?trip memberOf trainTrip
                        [     start hasValue ?start,
                              end hasValue ?end
                        ] and
                        (     ?start memberOf station
                        ) and
                        (     ?end = innsbruckHbf
                        )
                  )
                  or
                  (     ?trip memberOf trip
                        [     start hasValue ?altStart
                        ] and
                        (     ?altStart memberOf loc:location
                        )
                  )
            ).
```

## Стъпка 19

Само за да илюстрираме по-пълно механизма за разпространение на редакциите в модела, ще променим името на свързаната с атрибута *?trip* променлива директно чрез свойствата на атрибута. Променяме името от *?trip* на *?myTrip* в изгледа на свойствата на атрибута (горе вдясно, заградено в кръгче).

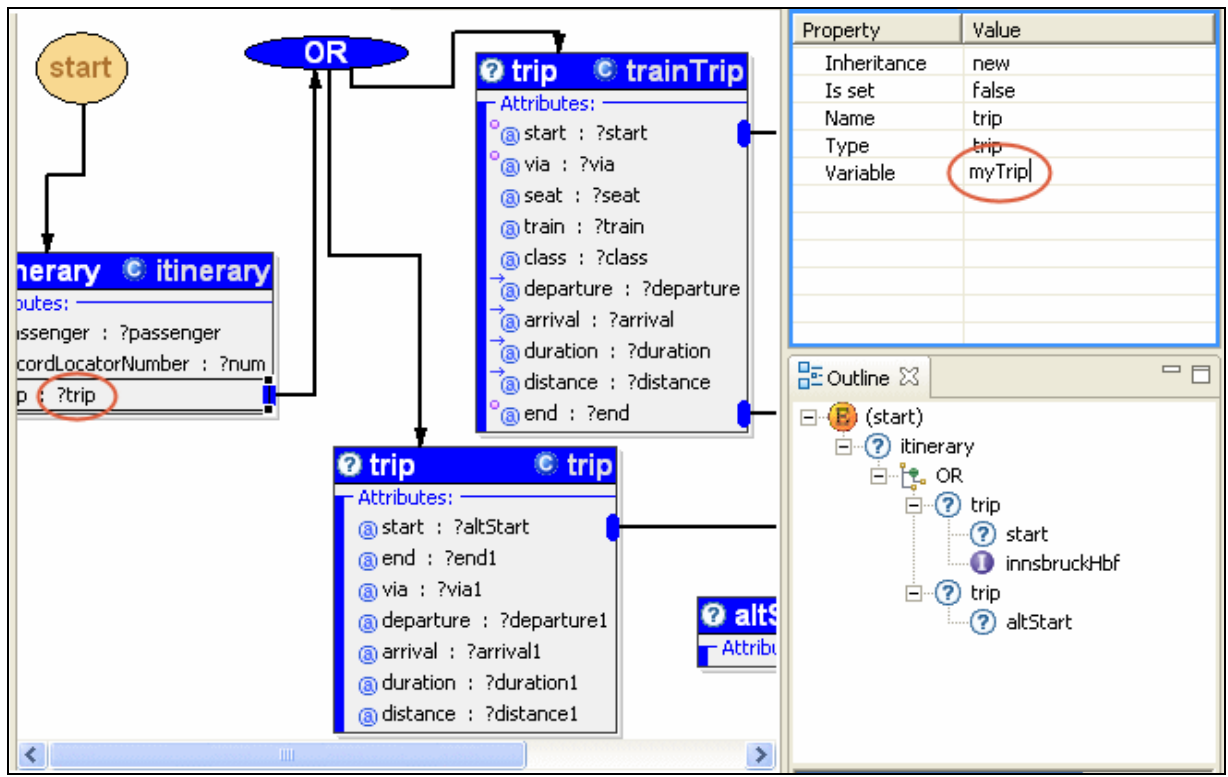

## Стъпка 20

В този момент нашата редакция се разпространява автоматично от Axiom Editor по цялата верига от алтернативи, които сме свързали с атрибута *trip*. Местата, които се променят в този момента, са отбелязани с кръгчета:

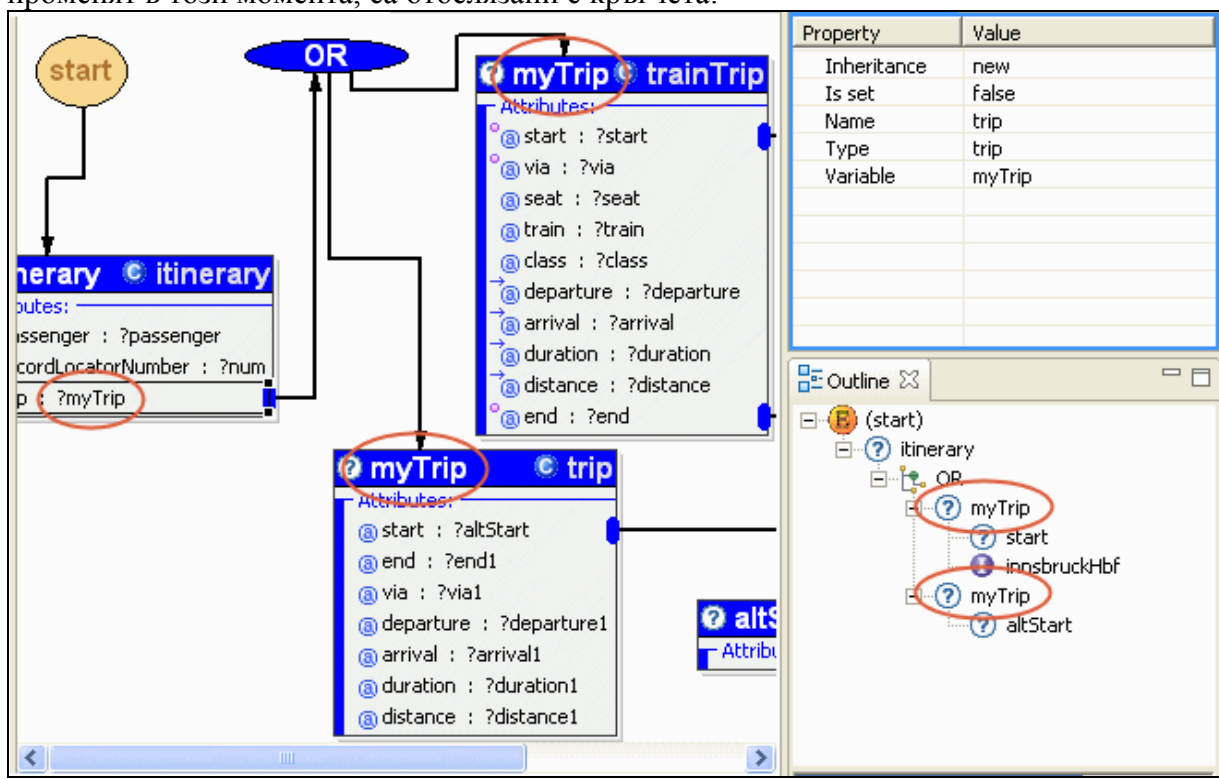

Тези промени веднага се отразяват и на генерирания текст на аксиомата:

```
axiom autoGeneratedAxiom_61
      nonFunctionalProperties
            dc:description hasValue "Auto-generated axiom by Axiom Editor"
      endNonFunctionalProperties
      definedBy
            ?itinerary memberOf itinerary
            [     trip hasValue ?myTrip
            ] and
            (     (     ?myTrip memberOf trainTrip
                        [     start hasValue ?start,
                              end hasValue ?end
                        ] and
                        (     ?start memberOf station
                        ) and
                        (     ?end = innsbruckHbf
                        )
                  )
                  or
                  (     ?myTrip memberOf trip
                        [     start hasValue ?altStart
                        ] and
                        (     ?altStart memberOf loc:location
                        )
                  )
            ).
```

## Стъпка 21

Продължаваме да развиваме логическата структура на аксиомата. Този път решаваме да добавим конюнкция в началото на аксиомата, веднага след коренния елемент *start*. Правим го, като използваме контекстното меню на връзката от *start* към променливата *?itinerary*. Ето как изглежда операцията:

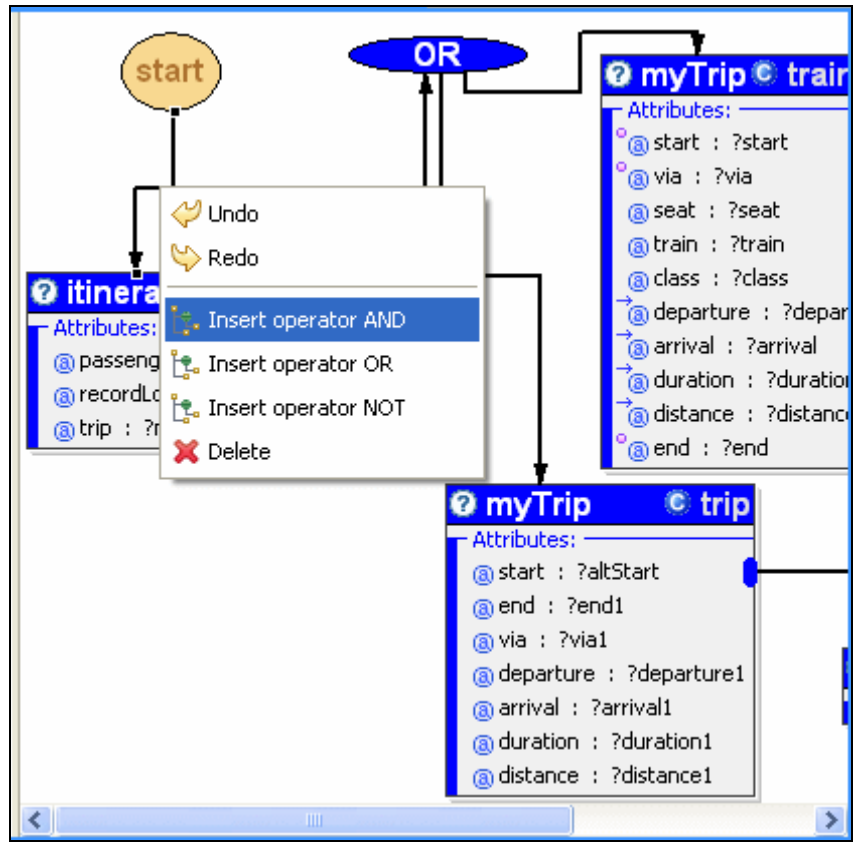

## Стъпка 22

В този момент на екрана се появява операторът AND, който се вмъква в избраната връзка. Забележете, че в момента той има само един операнд, което не е валидно, тъй като операторите AND и OR изискват поне два операнда. Ето защо в този момент генерираният текст на аксиомата все още не съдържа никакъв текст за оператора AND, въпреки че той се появява в модела и в съкратения изглед като възел. Ето как изглежда в този момент моделът (въпросният AND е показан с по-различен цвят):

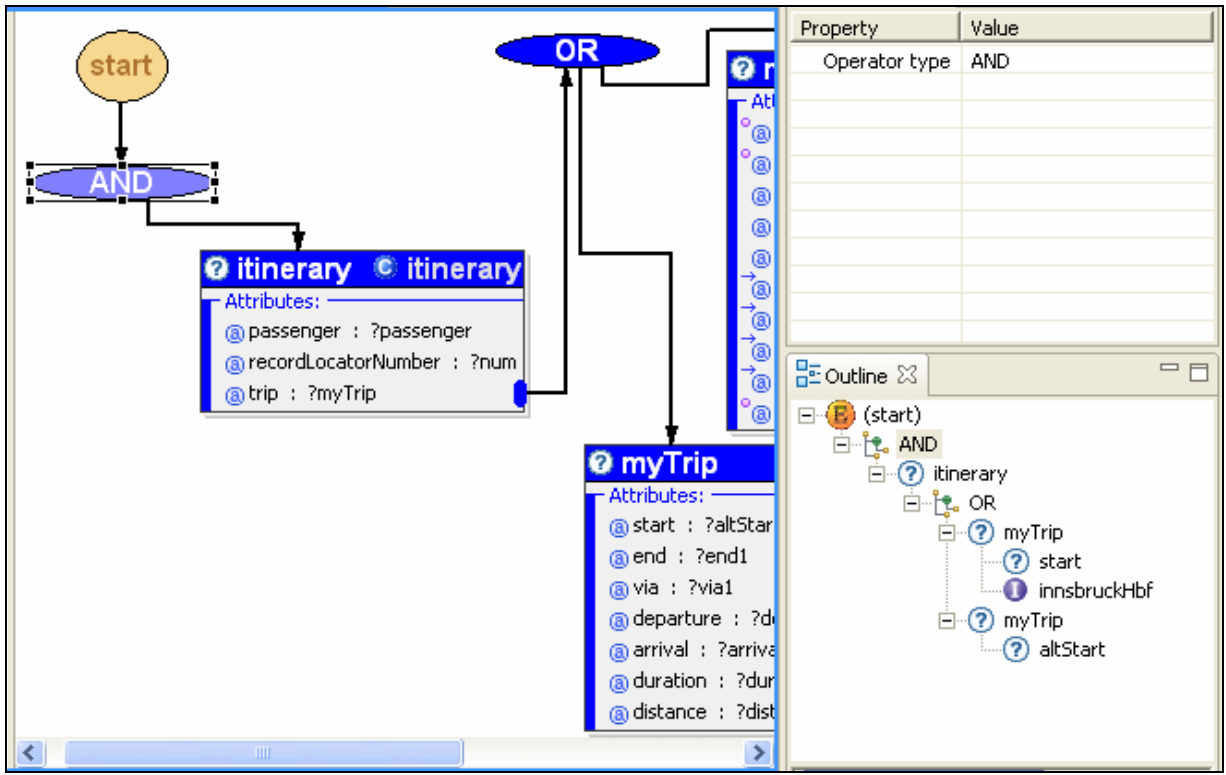

## Стъпка 23

Избираме от хранилището за онтологии някоя релация, например релацията *equalDistance* и я добавяме към модела. Тя ще се появи в него, като все още няма да е свързана с никой елемент от графичното представяне на аксиомата. Всички такива несвързани елементи в графичния модел се игнорират от Axiom Editor и не се включват нито в съкратения изглед (долу вдясно), нито в текста на генерираната аксиома.

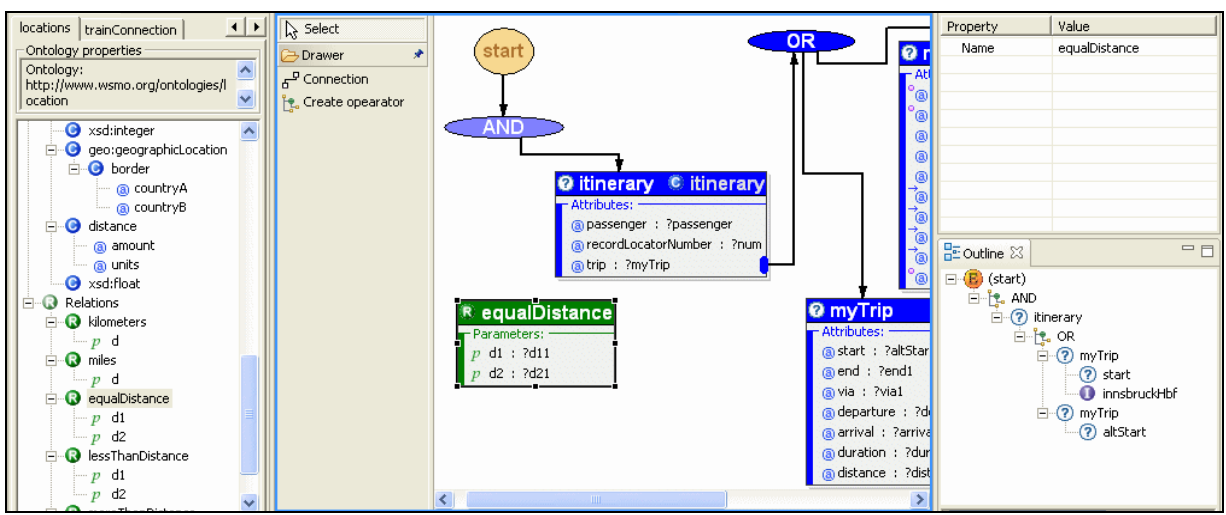

*Стъпка 24*

Ще използваме операцията по добавяне на връзка между елементи от графичния модел. За целта избираме "Connection" от палитрата с инструменти и избираме с мишката за начало на връзката оператора AND. В този момент Axiom Editor преминава в режим на свързване, като непрекъснато показва при движение на мишката дали е семантично правилно да се свърже началната точка с обекта под курсора на мишката. Ето как изглежда този „висящ край" на връзката, който още не сме свързали с нищо:

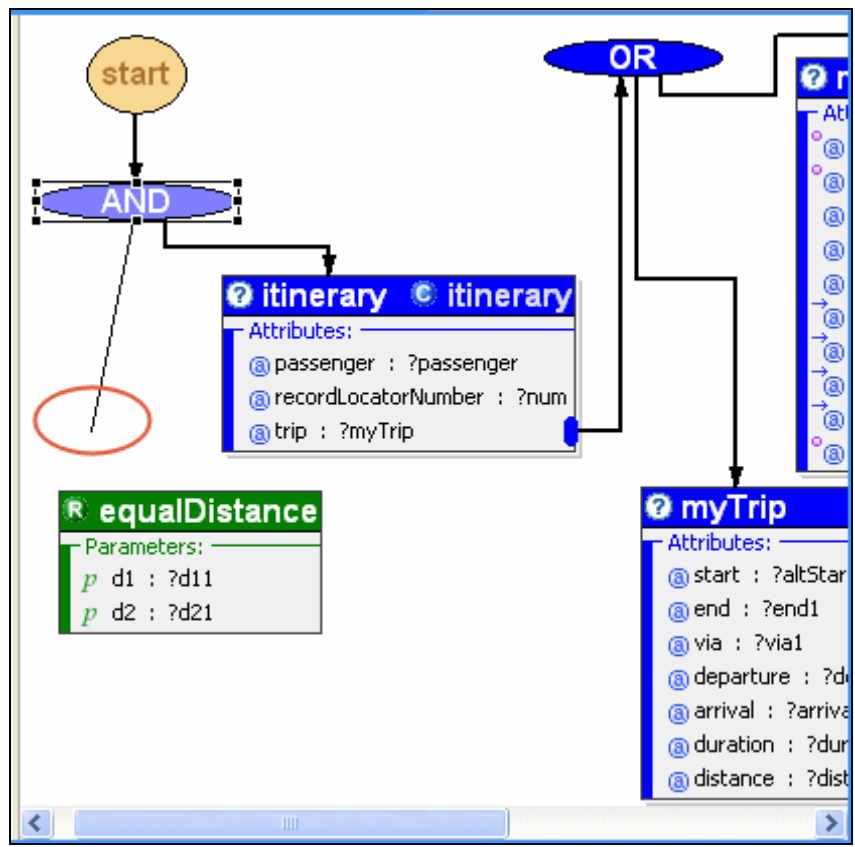

*Стъпка 25*

Избираме за крайна точка на връзката релацията *equalDistance*. Доскоро несвързана, сега вече тази релация става част от аксиомата и съответно се появява и в съкратения изглед, ето така:

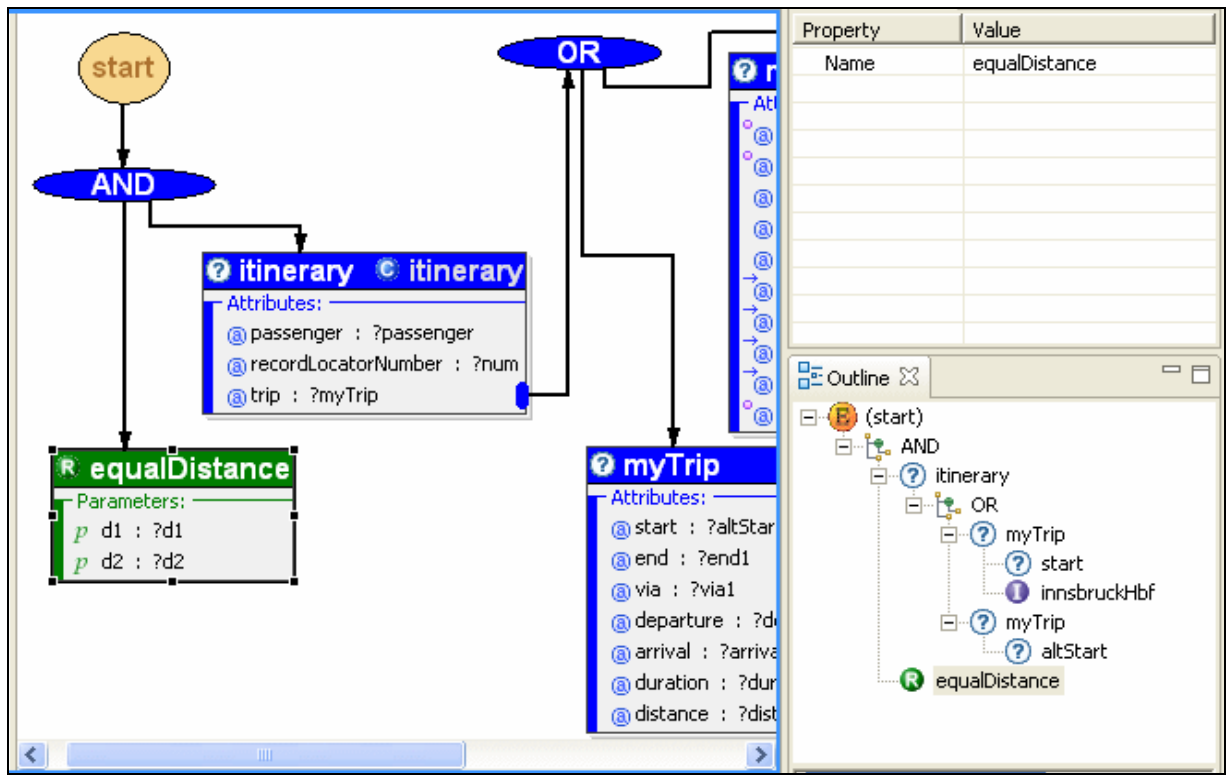

Забележете, че параметрите на релацията не са уточнени. Ето защо в генерирания текст на аксиомата те са заместени с анонимни променливи (?#), ето така:

```
axiom autoGeneratedAxiom_61
      nonFunctionalProperties
            dc:description hasValue "Auto-generated axiom by Axiom Editor"
      endNonFunctionalProperties
      definedBy
            (        ?itinerary memberOf itinerary
                  [     trip hasValue ?myTrip
                  ] and
                  (     (     ?myTrip memberOf trainTrip
                              [     start hasValue ?start,
                                    end hasValue ?end
                              ] and
                              (     ?start memberOf station
                              ) and
                              (     ?end = innsbruckHbf
                              )
                        )
                        or
                        (     ?myTrip memberOf trip
                              [     start hasValue ?altStart
                              ] and
                              (     ?altStart memberOf loc:location
                              )
```

```
                                    )
                                )
                            )
                        and
                    (        equalDistance(?#, ?#)
                    ).
```

## Стъпка 26

Все пак нека свържем тези параметри с конкретни променливи. Ще използваме контекстното меню на параметъра *d1*, за да създадем нова променлива и да я свържем с него:

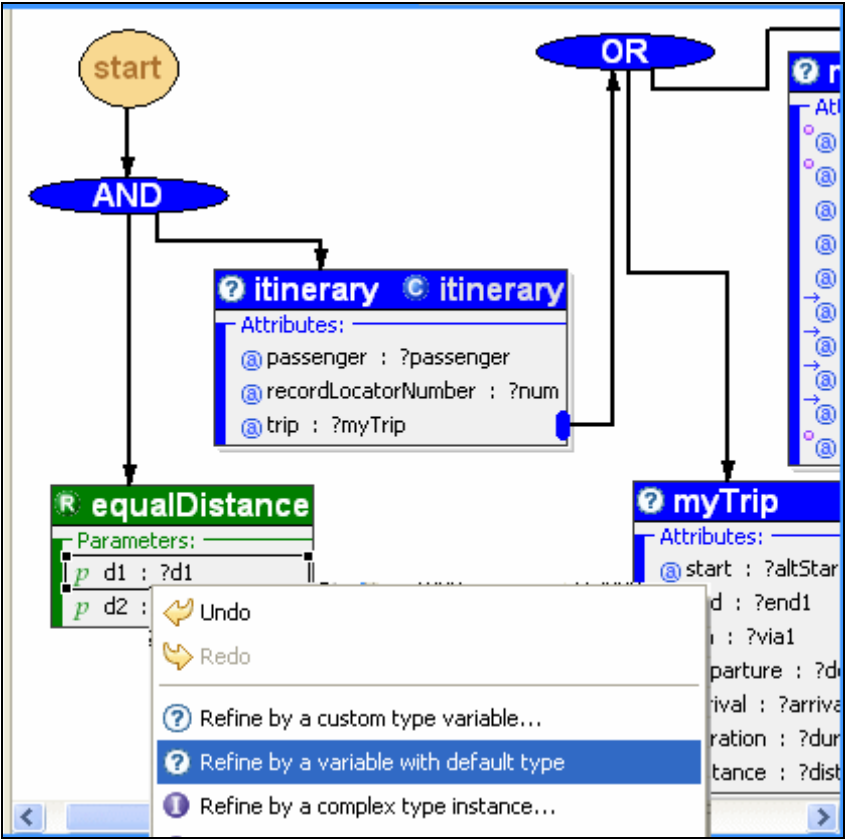

## Стъпка 27

Новата променлива получава името *?d1*, взето от името на параметъра, към който е свързана. Ето как изглежда графичният модел на аксиомата:

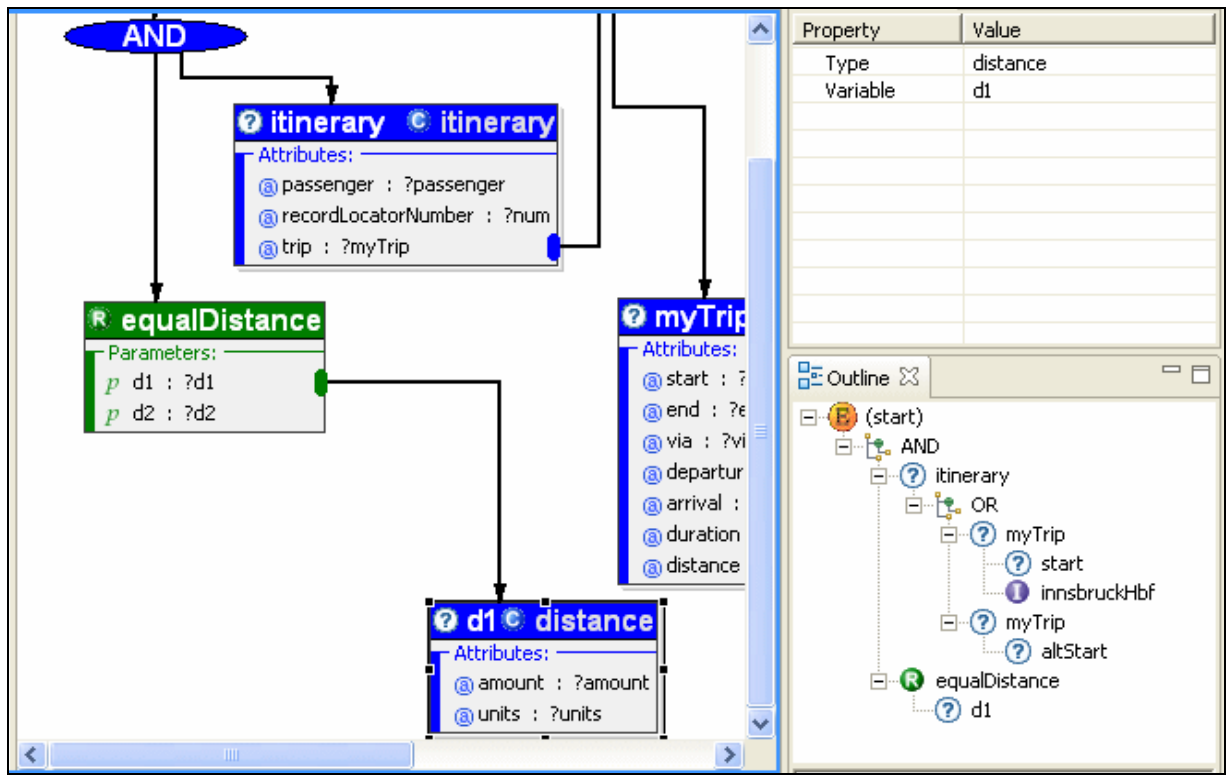

В текста на аксиомата вече първият параметър на релацията *equalDistance* е уточнен с конкретна променлива:

```
axiom autoGeneratedAxiom_61
      nonFunctionalProperties
            dc:description hasValue "Auto-generated axiom by Axiom Editor"
      endNonFunctionalProperties
      definedBy
            (       ?itinerary memberOf itinerary
                    [      trip hasValue ?myTrip
                    ] and
                    (      (      ?myTrip memberOf trainTrip
                                  [      start hasValue ?start,
                                         end hasValue ?end
                                  ] and
                                  (      ?start memberOf station
                                  ) and
                                  (      ?end = innsbruckHbf
                                  )
                           )
                           or
                           (      ?myTrip memberOf trip
                                  [      start hasValue ?altStart
                                  ] and
                                  (      ?altStart memberOf loc:location
                                  )
```

```
                            )
                       )
                  )
             and
             (      equalDistance(?d1, ?#)
                  and
                  (       ?d1 memberOf distance
                  )
             ).
```

## Стъпка 28

Ако редактираме името на променливата, свързана с параметъра, Axiom Editor автоматично ще разпространи тази промяна навсякъде в модела. Нека сменим името от *?d1* на по-смисленото *?smallDist*. Ето къде се отразява това в средата:

## Стъпка 29

Завършваме аксиомата, като задаваме конкретна стойност на атрибута *amount* от променливата *?smallDist* и също така уточняваме с променливата *?bigDist* втория параметър *d2* на релацията *equalDistance*. Ето окончателния графичен вид на модела на аксиомата в Axiom Editor:

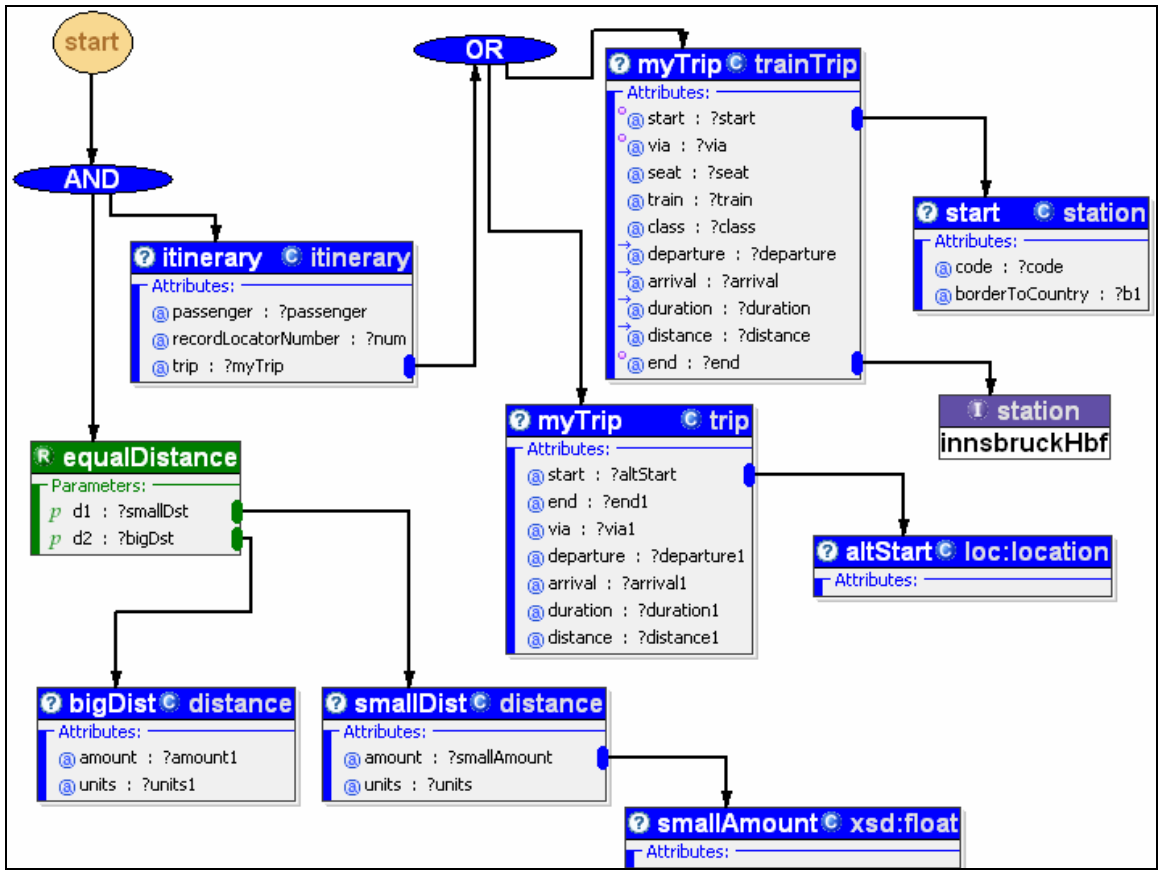

Ето как изглежда съкратения изглед на аксиомата в нейния завършен вид:

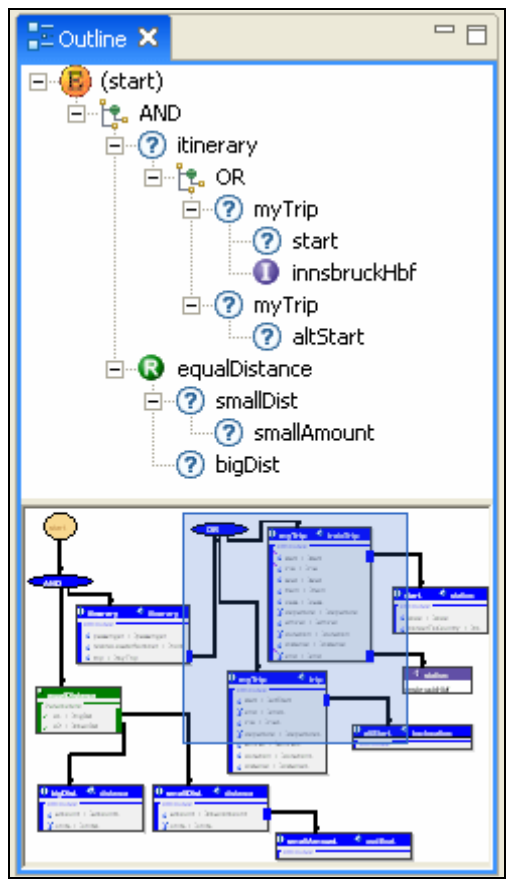

Това е автоматично генерираният текст на аксиомата, който съответства на графичния модел на аксиомата:

```
axiom autoGeneratedAxiom_61
        nonFunctionalProperties
                dc:description hasValue "Auto-generated axiom by Axiom Editor"
        endNonFunctionalProperties
        definedBy
                (       ?itinerary memberOf itinerary
                        [       trip hasValue ?myTrip
                        ] and
                        (       (       ?myTrip memberOf trainTrip
                                        [       start hasValue ?start,
                                                end hasValue ?end
                                        ] and
                                        (       ?start memberOf station
                                        ) and
                                        (       ?end = innsbruckHbf
                                        )
                                )
                                or
                                (       ?myTrip memberOf trip
                                        [       start hasValue ?altStart
                                        ] and
                                        (       ?altStart memberOf loc:location
                                        )
                                )
                        )
                )
                and
                (       equalDistance(?smallDist, ?bigDist)
                        and
                        (       ?smallDist memberOf distance
                                [       amount hasValue ?smallAmount
                                ] and
                                (       ?smallAmount memberOf xsd:float
                                )
                        )
                        and
                        (       ?bigDist memberOf distance
                        )
                ).
```

С този пример илюстрирахме създаването на една малка аксиома чрез Axiom Editor. Приложихме около 30 операции върху графичния модел на аксиомата и така получихме един сложен логически израз, който щеше да е доста по-трудно да се напише на ръка.

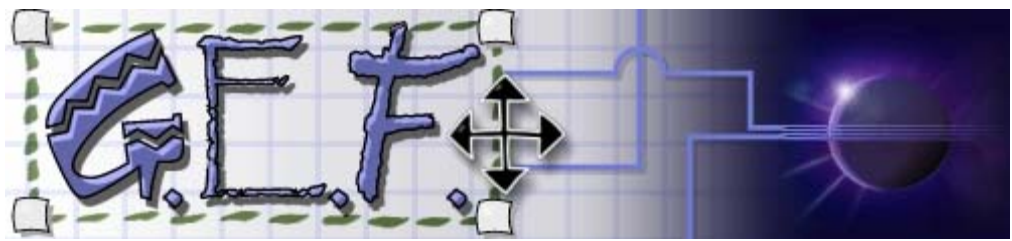

## *Приложение 2. GEF - платформа за графично редактиране*

Graphical Editing Framework (GEF) представлява платформа за графично редактиране. GEF предоставя мощни средства за създаване на визуални редактори за произволни модели. Ефективността на платформата се крепи на модулната й структура и ефективност, идваща от добре употребените шаблони за проектиране (design patterns)

За начинаещи ползватели, платформата GEF изглежда стряскаща, заради огромното количество термини и технологии, които трябва да се научат. Но веднъж научени и правилно използвани, те помагат да се разработи изключително сложен, гъвкав и в същото време лесен за поддържане софтуер. В тази секция се разглеждат най-основните принципи и техники на GEF, които се използват в имплементацията на INFRAWEBS Axiom Editor.

### Въведение в GEF

GEF е създаден за да улесни редактирането на потребителски данни, обикновено наричани *модел,* чрез използване на графични средства вместо текстов формат. Това е изключително ценни, когато става дума за случаи на много-към-много връзки в сложни модели. Примери за подобни приложения са редактори на схеми за бази данни, редактори за електронни схеми, редактори за блок-схеми и т.н.

Бедата с всяка подобна платформа (и GEF не е изключение) е, че сложният й дизайн я прави много трудна за научаване. Като пример ще посочим, че дори най-елементарният пример, който е наличен за GEF, включва над 75 потребителски класа на java. Опитът да се разбере в детайли механизма на работа на GEF при многото потребителски класове и стотиците вградени в GEF представлява истинско предизвикателство към проницателността дори и на най-прилежните програмисти.

На фигура 1 е показан екран от примерен редактор за геометрични форми и връзки между тях.

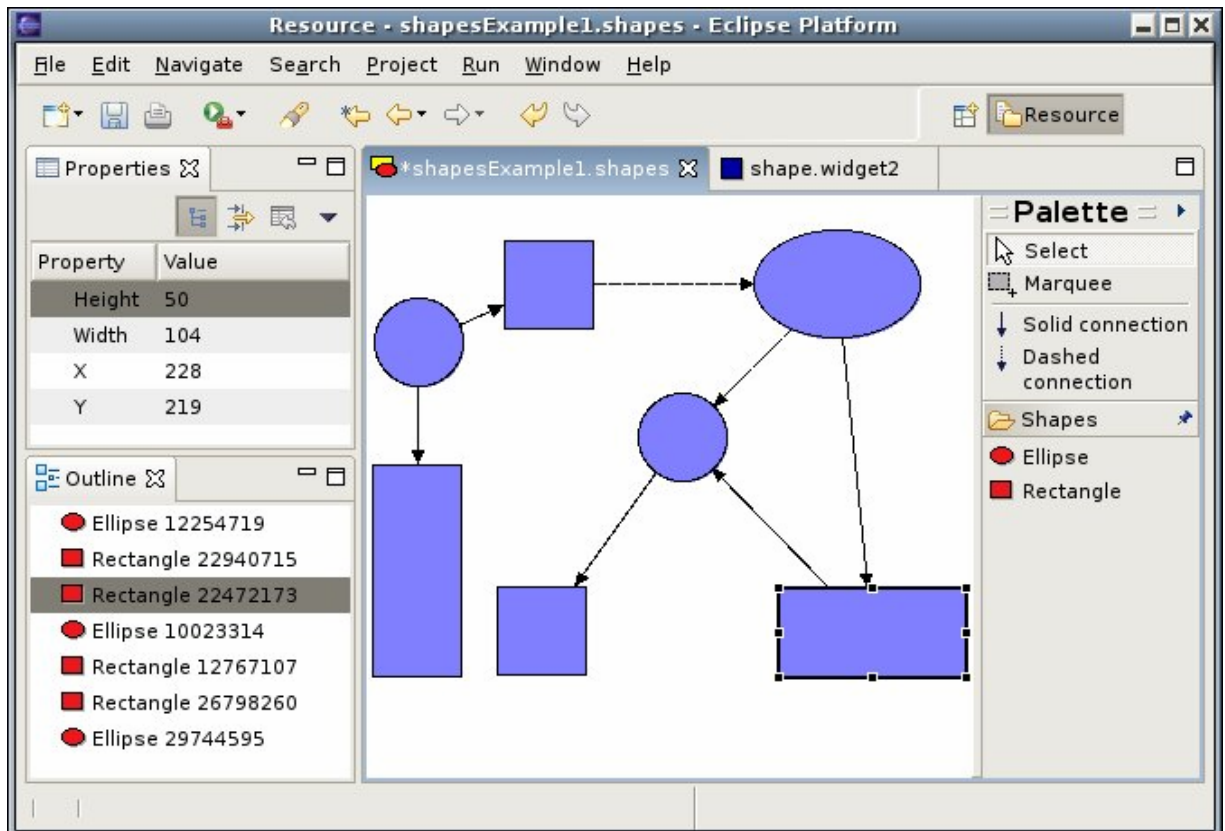

*Фигура 1. Редактор за геометрични форми и връзки между тях.*

## Основни понятия в GEF

GEF улеснява построяването на визуален редактор за потребителски данни. Данните могат да бъдат съвсем прости, като например термостат с единствено копче, или много сложни, като например частна мрежа със стотици рутери, окабеляване и друго оборудване. Благодарение на умелия си дизайн, GEF успешно се справя с каквито и да е данни, или казано в терминологията на GEF, с какъвто и да е *модел*.

Това се постига чрез стриктното спазване на т.нар. шаблон Модел-Изглед-Контролер (Model-View-Controller pattern). Следва описание на концептуалните му елементи: модел, изглед и контролер.

*Моделът* е изграден от потребителските данни. За GEF моделът е просто съвкупност от Java обекти. Моделът не трябва да знае нищо нито за контролера, нито за изгледа.

*Изгледът* е визуалното представяне на модела или негови части на екрана. Тя може да съвсем проста като правоъгълник, линия или елипса, или пък много сложна като сложна електрическа верига. По подобие с модела, изгледът не трябва да знае нищо нито за модела, нито за контролера. GEF използва *Draw2D фигури* за да създаде изгледа.

*Контролерът* посредничи между модела и изгледа (виж Фигура 2). Всеки контролер (нарича се с термина EditPart) отговаря както за съответствието между модела и изгледа, така и за отразяване на всички промени в модела. Контролерът наблюдава всички промени в модела и обновява изгледа, така че да отразява текущото състояние на модела. Всъщност контролерът е този, с който потребителят взаимодейства.

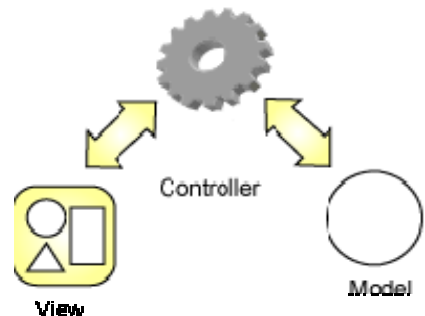

*Фигура 2*. Модел-Изглед-Контролер (Model-View-Controller pattern)

## Техники на работа в GEF

Тук ще опишем още някои техники в GEF, които са от значение за пълно използване на възможностите на платформата. Те са свързани с понятията: *инструмент, заявка, политика* и *команда*.

*Инструментите* представляват обекти, които преобразуват събития на ниско ниво, като например местене на мишката или натискания по клавиатурата, в обекти от по-високо ниво, които се наричат *заявки*.

Това, коя заявка ще се създаде, зависи от това, кой инструмент е избран в дадения момент. Например, инстументът за създаване на връзки между обекти при получаване на съобщение за натискане на мишката създава заявка за начало на връзка и фиксира началото й. Ако обаче е активен инструментът за създаване на нов обект, то за същия вход щеше да се генерира заявка за създаване на обект. В GEF има няколко предварително дефинирани инструмента, като могат да се дефинират допълнително от програмистите. Инструментите могат да се активират програмно или по желание на потребителя.

През повечето време, инструментите изпращат заявките към контролерите на съответните фигури, върху които е било кликнато с мишката например. Има и изключения от това правило. Във всички случаи обаче, самият контролер не обработва директно заявките. Вместо това той делегира обработката им на регистрираните *политики*.

Към всяка политика се отправя запитване за създаване на *команда*, която съответства на дошлата заявка. Всяка политика има две опции: или да върне валидна команда, или да откаже да обработи заявката. Механизмът на политиките позволява за всяко логически свързано подмножество от заявки да се създаде политика, която да знае необходимите команди за обслужване на тези заявки. След това тези политики могат да се инсталират към различните контролери.

Последното „парче от пъзела" са *командите*. Вместо да се модифицира директно моделът, GEF изисква програмистът сам да дефинира какво се случва на модела в резултат от изпълнението на всяка команда. Тук е мястото, което осигурява прилагане или отмяна на извършена команда (undo/redo) чрез запомняне на предишното състояние на модифицираната част от модела.

## Интегриране с Eclipse

Едно от основните предимства на GEF, освен че кара програмистите да повишат уменията си и да научат шаблони за проектиране, е възможността за пълна интеграция с платформата на Eclipse.

На избраните обекти в редактора на GEF могат да се визуализират свойствата в прозореца *Properties* на Eclipse. Функциите *Undo* и *Redo* от менюто *Edit* също функционират и извикват съответната функционалност на GEF командите.

Казано по-просто, редакторите на GEF се възползват пълноценно от предимствата на платформата Eclipse.

## Начини за визуализация в GEF

GEF поддържа два вида визуализации: *графична* и *дървовидна*.

*Графичната визуализация* използва Draw2D фигури, които се изчертават върху SWT Canvas (платно). Фигурите се създават с използване на Draw2D приставката (plug-in), която се разпространява заедно с GEF. Обикновено тази визуализация се използва в основната работна област на редактора, в която се извършва самото редактиране на модела. Като пример за такава визуализация можем да посочим един редактор за схеми на бази данни [Zoio, 2004] (виж фигура 3).

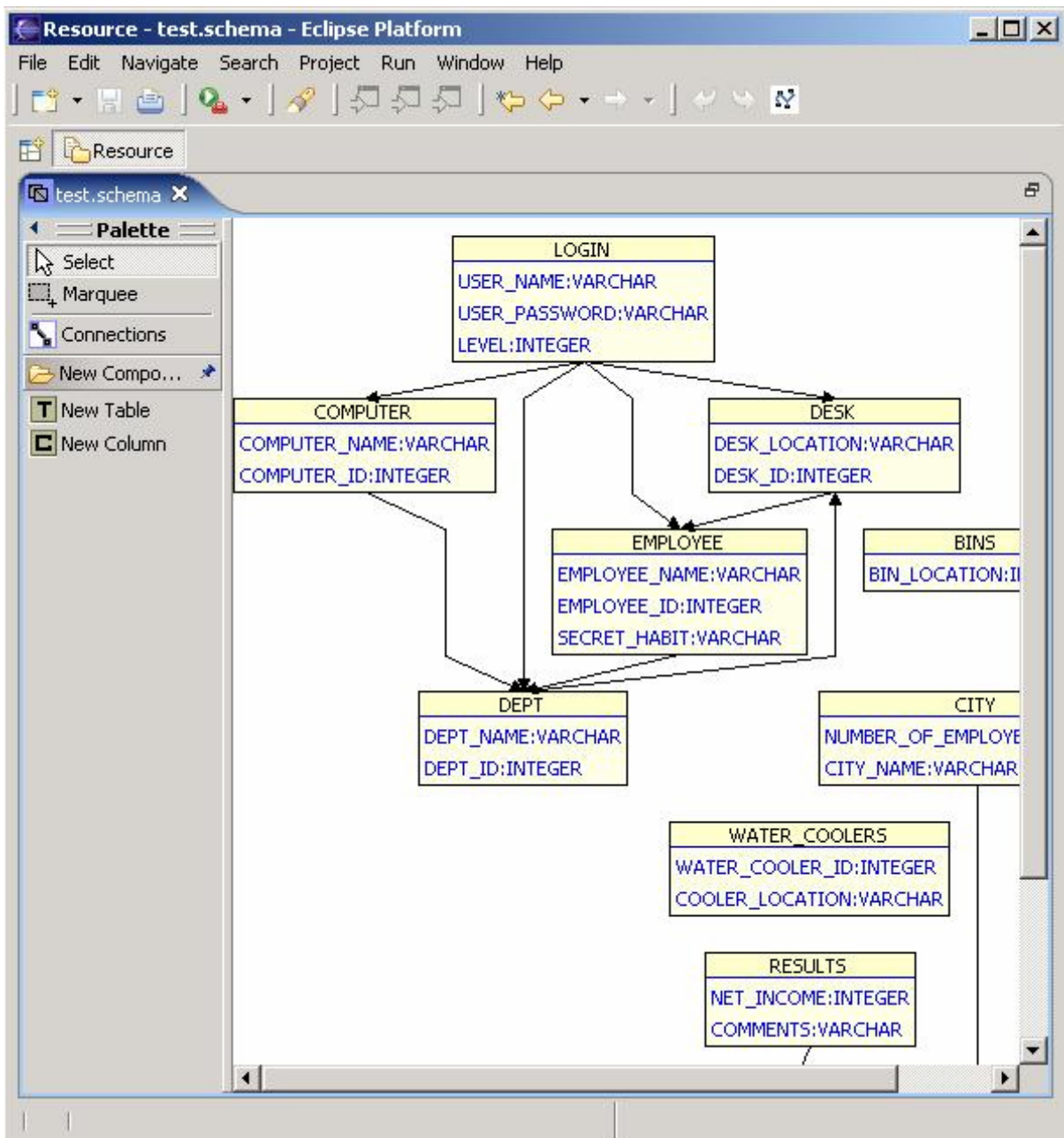

*Фигура 3. Графична визуализация на схема на база данни, реализирана с GEF*

*Дървовидната визуализация* използва SWT Tree (дърво) за да се изчертае на екрана. Обикновено тази визуализация се използва в компактното представяне на модела в прозореца *Outline*. Като пример за такава визуализация можем да посочим дървовидно представяне на йерархия от класове и техните методи (виж фигура 4).

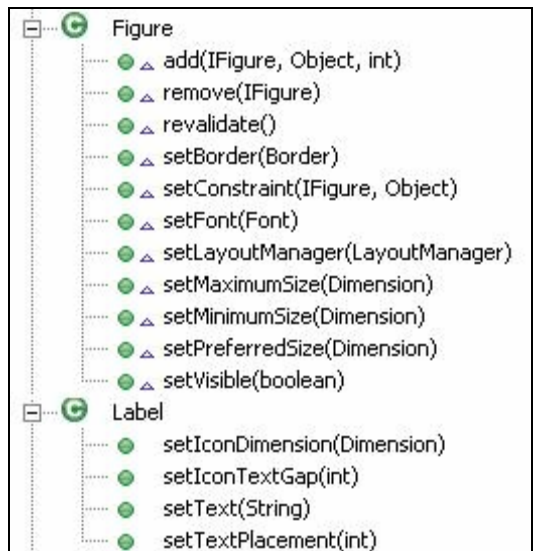

*Фигура 4. Дървовидна визуализация, реализирана с GEF*

## Спомагателни елементи в GEF

GEF разполага и с много други по-малки или по-големи интерфейсни елементи, които могат да се използват според нуждите за конкретни редактори. Тук ще споменем само някои от тях: *палитра, лента с операции, общ изглед* и *страница със свойства*.

Палитрата (Palette) показва множеството от налични инструменти за създаване на обекти в диаграмата или за редакцията им. Потребителят може да активира инструменти от нея с мишката.

Лентата с операции (Action bar) показва операциите, които са разрешени за текущо избраните обекти. Тези операции също могат да се показват в главното меню или в контекстни менюто, в зависимост от реализацията и търпението на програмистите.

Страницата със свойства (Property sheet) показва детайли за избрания обект. Някои от свойствата са само за четене, други могат и да се променят от там, като се осъществява валидиране на новите стойности.

Общият изглед (Outline) предоставя съкратен поглед върху структурата на диаграмата. Обикновено в него се използва дървовиден вид визуализация чрез GEF TreeViewer.

## Обобщение

Успяхме да покрием голяма част от принципите на GEF. Това е от голямо значение, за да бъде разбрана правилно реализацията на настоящият редактор за сложни логически изрази. Неговото разглеждане ще използва наготово дефинираните тук термини и е важно те да бъдат ясни.

Основното, което трябва да се наблегне е, че в GEF се използва изчистен шаблон Модел-Изглед-Контролер, който налага ред ограничения и върху използващия платформата редактор.

## *Приложение 3. Списък на възприетите преводи на английски термини*

В тази таблица са събрани всичките възприети в дипломната работа преводи от английски термини на български език.

| Английски термин | Възприет превод на български език |
|---|---|
| Assumption | Предположение |
| auto-completion | автоматично дописване |
| Composite web service | Съставна мрежова услуга |
| Composition of services | Композиция / Композиране на услуги |
| Computer application | Компютърно приложение |
| Concept | Понятие |
| Construction steps | Фази на конструирането |
| Context-sensitive | Контекстно-зависим |
| Design | Разработка |
| Discovery of Web service | Откриване на мрежова услуга |
| Effect | Ефект |
| Envisioning | Проектиране |
| Framework | Среда |
| Goal of a web service | Цел на мрежова услуга |
| Human-readable | Разбираем за човек |
| Implementation | Имплементация |
| Instance | Екземпляр |
| Interface | Интерфейс |
| Invocation of Web service | Извикване на мрежова услуга |
| Keyword | Ключова дума |
| Logical development | Логическо развиване |
| Mediator of a web service | Медиатор на мрежова услуга |
| Nonfunctional properties | Нефункционални свойства |
| Ontology | Онтология |
| ontology-driven | ръководен от онтологии |
| Post-condition | Пост-условие |
| Pre-condition | Пред-условие |
| Publish | Публикуване |
| Refinement | Уточняване |
| Relation | Релация |
| Semantic markup | Семантична анотация |
| Semantic Web | Семантичната глобална мрежа |
| Semantic web service | Семантична мрежова услуга |
| Service capabilities | Възможности на услугата |
| *Service Interface* | Интерфейс на услугата |
| Service orchestration | Оркестрация от услуги |
| Shared variable | Споделена променлива |

| Subconcept | Под-понятие |
| Superconcept | Над-понятие |
| Syntactic/semantic consistence | Синтактична/семантична съгласуваност/правилност |
| Syntax highlighting | Синтактично оцветяване |
| Tool | Средство, инструмент |
| user-friendly | Удобна за потребителя/използване |
| Web resources | Мрежови ресурси |
| Web service | Мрежова услуга |
| Workflow | Работна последователност |

## *Приложение 4. Списък на използваните съкращения*

В тази таблица са събрани всички използвани съкращения на английски език в изложението. Някои от тях са добре утвърдени в IT сферата, други са въведени в настоящата работа с цел улесняване на записа.

| Съкращение | Пълно наименование |
|---|---|
| BPEL4WS | Business Process Execution Language for Web Services |
| CBR | Case-Based Reasoning |
| DC | Dublin Core |
| DST | Design Service Templates |
| PPAE | pre-conditions, post-conditions, assumptions, effects |
| QoS | Quality of Service |
| RDF | Resource Description Framework |
| SOA | Service Oriented Architecture |
| SWS | Semantic web service |
| SWSD | Semantic Web Service Designer |
| SWU | Semantic Web Unit |
| UDDI | Universal Description, Discovery and Integration |
| UML | Unified Modeling Language |
| WS | Web service |
| WSDL | Web Service Description Language |
| WSML | Web Service Modeling Language |
| WSMO | Web Service Modeling Ontology |
| WYSIWYG | What You See Is What You Get |
| XSD | XML Schema Definition |